\documentclass[letterpaper]{article}
\usepackage{reloop_preprint}
\usepackage[hyphens]{url}
\usepackage{colortbl}
\usepackage{graphicx}
\usepackage{natbib}
\usepackage{caption}
\usepackage{booktabs}
\usepackage{amsmath,amssymb}
\usepackage{algorithm}
\usepackage{algorithmic}
\usepackage{array}
\usepackage{placeins}
\usepackage{microtype}
\AtBeginDocument{\microtypesetup{deactivate}}

\title{ReLoop-UME: Recurrent Depth with Learnable Retrieval Registers for Universal Multimodal Embedding}
\author{
Shijie Wang\textsuperscript{\rm 1}\equalcontrib,
Xiangzhao Hao\textsuperscript{\rm 1}\equalcontrib,
Yueti Li\textsuperscript{\rm 2},\\
Guangyu Cao\textsuperscript{\rm 1},
Xinyu Tang\textsuperscript{\rm 3}\corresponding,
Haiyun Guo\textsuperscript{\rm 1}\corresponding
}
\affiliations{
\textsuperscript{\rm 1}Institute of Automation, Chinese Academy of Sciences, Beijing, China\\
\textsuperscript{\rm 2}Tsinghua University, Beijing, China\\
\textsuperscript{\rm 3}AI Platform, evermind\\
wangshijie2026@ia.ac.cn, hxzemail@gmail.com, liyueti0117@163.com,\\
caoguangyu2026@ia.ac.cn, xinyu.tang@evermind.ai, haiyun.guo@nlpr.ia.ac.cn
}

\begin{document}

\maketitle

\begin{abstract}
Universal multimodal embedding (UME) maps heterogeneous multimodal inputs into a shared embedding space. Existing UME models either form embeddings through single forward encoding or add computation through explicit rationale tokens and latent autoregressive states. Although token expansion can improve complex matching, serial generation increases retrieval latency and makes the final embedding depend on generated intermediate states. This raises a different question: can useful computation be expanded along model depth while keeping the token workspace fixed? We analyze positive--negative similarity separation at every layer of independently trained UME models and observe a shared progression: early layers contextualize multimodal inputs, a contiguous middle-to-late stage forms retrieval-discriminative features, and the final layers map them into the embedding space. Based on this finding, we propose ReLoop-UME, which executes the early layers once, recurrently reuses a parameter-shared retrieval-forming block, and applies the final mapping layers after the last loop. Learnable Retrieval Registers provide persistent retrieval-specific states that accumulate and exchange evidence across loops, with the final register serving as the embedding readout. On MMEB-V2 and MRMR, ReLoop-UME consistently improves retrieval across different backbones while running 44.9$\times$ faster than UME-R1 and 1.5$\times$ faster than PLUME.
\end{abstract}

\section{Introduction}
\label{sec:introduction}

\suppressfloats[t]
\begin{figure}[t]
    \centering
    % Replace this placeholder with the final layer-wise discriminability plot.
     \includegraphics[width=\columnwidth]{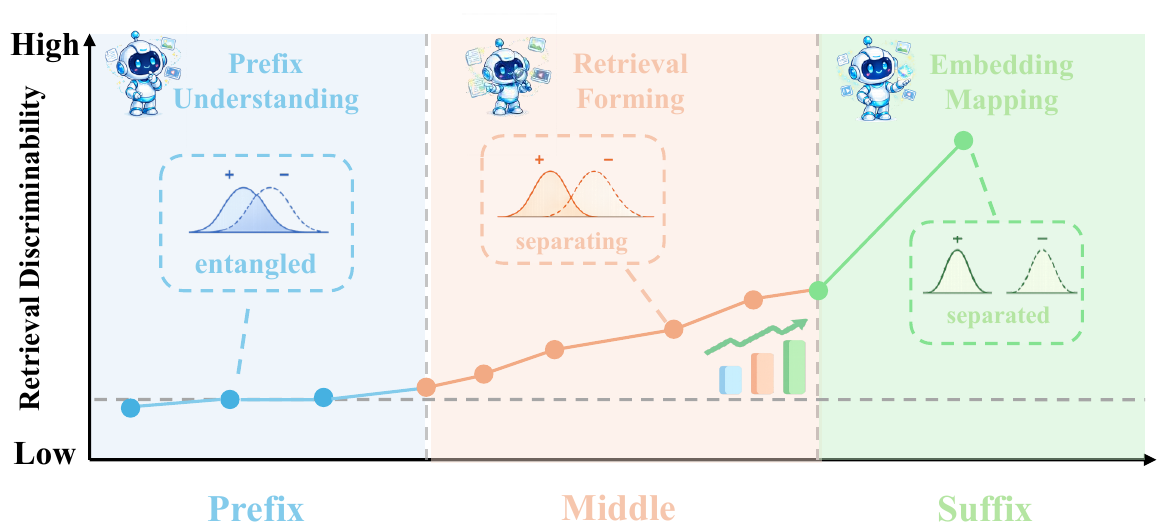}
    \caption{
Abstract layer-wise retrieval-discriminability trend across independently trained UME models. Full curves and model-specific boundaries are provided in the appendix.
}
    \label{fig:layer_discriminability}
\end{figure}

\begin{figure*}[t!]
    \centering
    \includegraphics[width=0.99\textwidth]{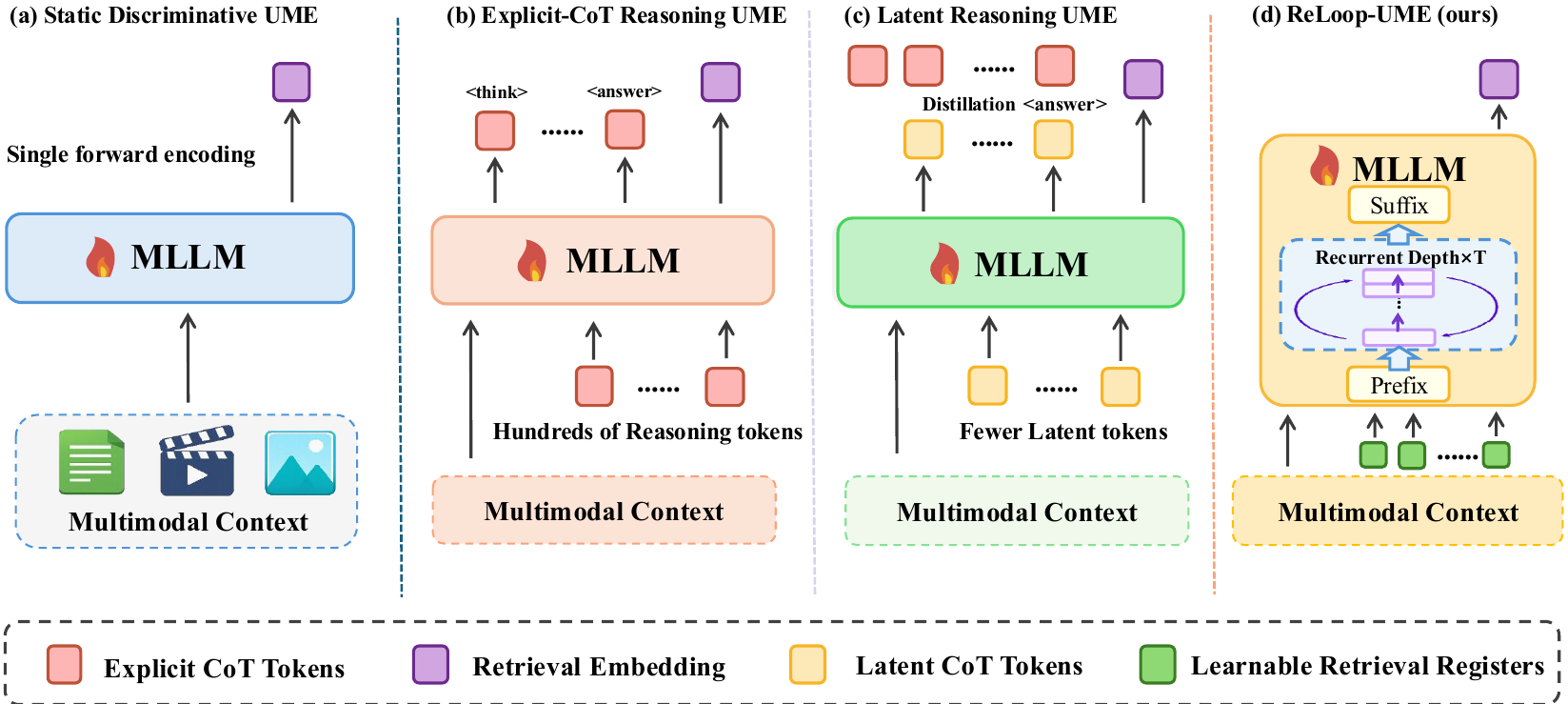}
    \caption{Comparison of UME computation paths. Existing methods allocate
    computation through either single forward encoding or extended explicit/latent
    token trajectories. ReLoop-UME instead executes the prefix once,
    recurrently reuses the retrieval-forming stage, and maps the refined states
    to the final embedding.}
    \label{fig:ume_paradigms}
\end{figure*}

\textbf{Beyond token-expanded reasoning.}
Universal multimodal embedding (UME) encodes heterogeneous queries and candidates, including text, images, videos, and visual documents, into a shared embedding space, where relevance is measured by embedding similarity. Conventional models such as VLM2Vec~\cite{jiang2025vlm2vec}, GME~\cite{zhang2025gme}, LamRA~\cite{liu2025lamra}, VLM2Vec-V2~\cite{meng2026vlm2vecv2}, and BToks~\cite{sun2026btoks} produce each representation through single forward encoding. This paradigm is efficient, but its fixed computation can be insufficient for complex query intent and fine-grained cross-modal matching. Recent methods therefore add computation before embedding readout. UME-R1~\cite{lan2026umer1} and TRACE~\cite{hao2026trace} generate explicit reasoning trajectories, whereas PLUME~\cite{he2026plume} uses latent autoregressive states. Their gains, especially without verbalized rationales, suggest that additional internal computation itself can improve retrieval. However, these methods still expand computation along the token dimension. Serial generation increases retrieval latency, repeatedly invokes the full model, and makes the final embedding depend on generated intermediate states. This motivates a different route: can additional computation be allocated along model depth while keeping the token workspace fixed?

\textbf{Retrieval formation is localized across depth.}
To determine where depth-expanded computation should be placed, we examine how retrieval structure develops inside independently trained UME models using single forward encoding. At every layer, we measure the separation between positive and negative query--candidate similarity distributions. Despite architectural differences, the models exhibit a shared progression. Early layers mainly contextualize multimodal inputs, while positive and negative similarities remain entangled; we call this the \textbf{Prefix Understanding Stage}. A contiguous middle-to-late interval then produces the largest sustained increase in separation, forming the \textbf{Retrieval Formation Stage}. Once the two populations are largely separated, the final layers mainly map the resulting retrieval features into the output embedding space, constituting the \textbf{Embedding Mapping Stage}. Retrieval capability therefore forms non-uniformly across depth. More importantly, this progression identifies a localized stage where additional depth can directly strengthen query--candidate discrimination, rather than recomputing input understanding or final embedding projection.

\textbf{ReLoop-UME couples localized recurrence with persistent retrieval state.}
Motivated by this observation, we propose \textbf{ReLoop-UME} (Fig.~\ref{fig:ume_paradigms}). The understanding layers form a prefix, the retrieval-forming layers form a parameter-shared recurrent block, and the mapping layers form a suffix. Each input is processed by the prefix once, recurrently updated by the same retrieval-forming block for $T$ loops, and converted into an embedding by the suffix after the final loop. This concentrates extra computation on the layers where positive--negative separation is established, without duplicating backbone parameters or extending the token sequence.

Localized recurrence introduces a state-management problem: evidence discovered during one loop must remain accessible to later loops, while the original multimodal tokens must preserve input content and continue being refined. Content-token positions alone provide no retrieval-specific workspace for carrying and consolidating this evidence. We therefore introduce \textbf{Learnable Retrieval Registers}, a small set of trainable states appended to the multimodal sequence. They attend to the complete input, persist through every loop, and provide dedicated positions for accumulating, retaining, and exchanging retrieval evidence. The final register serves as a stable embedding readout. Unlike explicit or latent reasoning tokens, the registers are not generated step by step and do not grow with recurrent depth; they form a fixed-size retrieval workspace that complements the refined content features.

\textbf{Recurrent retrieval refinement improves the accuracy--efficiency tradeoff.}
On MMEB-V2, ReLoop-UME reaches All scores of 63.2 at the 2B scale and 65.9 at the 7B scale, outperforming strong methods based on single forward encoding and token-expanded reasoning. The 7B model also transfers zero-shot to MRMR~\cite{zhang2026mrmr}, and the same recipe yields consistent gains across different backbones. ReLoop-UME runs 44.9$\times$ faster than UME-R1 and 1.5$\times$ faster than PLUME, with only 1.3$\times$ latency overhead over VLM2Vec-V2. Controlled ablations validate stage localization, moderate recurrent depth, and the compact register workspace.

Our contributions are summarized as follows:
\begin{itemize}
    \item \textbf{A stage-wise account of retrieval formation.}
    Layer-wise analyses reveal distinct stages for multimodal understanding, retrieval discrimination, and embedding mapping, identifying the retrieval-forming interval as the target for additional depth.

    \item \textbf{Localized recurrent refinement with dedicated retrieval state.}
    ReLoop-UME repeatedly applies a parameter-shared retrieval-forming block, while Learnable Retrieval Registers provide a fixed-size workspace for cross-loop evidence accumulation and embedding readout.

    \item \textbf{Stronger retrieval on a favorable efficiency frontier.}
    Evaluations on MMEB-V2, zero-shot MRMR transfer, and different backbones show consistent gains with substantially lower cost than token-expanded reasoning methods.
\end{itemize}

\section{Related Work}
\label{sec:related_work}

\paragraph{Universal Multimodal Embedding.} Early dual encoders CLIP, ALIGN, and SigLIP~\cite{radford2021clip,jia2021align,zhai2023siglip} established scalable image--text learning. UniIR and MagicLens~\cite{wei2024uniir,zhang2024magiclens} broadened retrieval to unified tasks and instruction-conditioned search. MLLM-based E5-V, MM-Embed, VLM2Vec, GME, UniME, LamRA, and VLM2Vec-V2~\cite{jiang2024e5v,lin2024mmembed,jiang2025vlm2vec,zhang2025gme,gu2025unime,liu2025lamra,meng2026vlm2vecv2} further improve compositional matching and modality coverage. MegaPairs, ColPali, BToks, and MetaEmbed~\cite{zhou2025megapairs,faysse2025colpali,sun2026btoks,lee2025metaembed} advance data scale, visual-document retrieval, and embedding readout. Most still form each embedding in one forward pass; we instead locate where retrieval geometry emerges and selectively reuse that stage.

\paragraph{Computation Enhanced Embeddings.} Think-Then-Embed (TTE)~\cite{cui2025tte} uses a separate reasoner to generate embedding-centric traces for an embedder; Reasoning Guided Embeddings~\cite{liu2025rge} similarly conditions representations on generated rationales. UME-R1, TRACE, and Embed-RL~\cite{lan2026umer1,hao2026trace,jiang2026embedrl} extend explicit computation through generative objectives, adaptive execution, or reinforcement learning. PLUME and BToks~\cite{he2026plume,sun2026btoks} respectively use autoregressive continuous states and a fixed latent bottleneck. TTE and Embed-RL also rely on separately constructed high-quality reasoning supervision. Because TTE is a two-model reasoner--embedder pipeline rather than a standalone single-model UME, we omit it from direct comparison. These works show extra computation helps, but token expansion ties embedding quality to intermediate trajectories and serial latency. ReLoop-UME instead keeps a fixed token workspace and recurrently refines the full hidden sequence in retrieval-forming layers.

\paragraph{Recurrent Feature Refinement.} The Universal Transformer~\cite{dehghani2019universal} introduced depth-wise parameter sharing, while looped Transformers~\cite{fan2025looped,saunshi2025latentthoughts} showed that repeated application increases effective computation without model growth. Huginn and Ouro~\cite{geiping2025recurrentdepth,zhu2025ouro} apply recurrent depth to pretrained language models; ETD and LoopUS~\cite{koishekenov2025etd,park2026loopus} localize or organize recurrence using layer-wise dynamics. ReLoop-UME transfers this principle to universal multimodal retrieval: positive--negative discriminability selects the recurrent stage, which jointly refines query and candidate features in one contrastive embedding space.

% Add the following packages to the preamble:
% \usepackage{amsmath,amssymb}
% \usepackage{algorithm}
% \usepackage{algorithmic}

\begin{figure*}[t]
    \centering
    \includegraphics[
        width=0.99\textwidth,
        height=0.4\textheight,
        keepaspectratio,
    ]{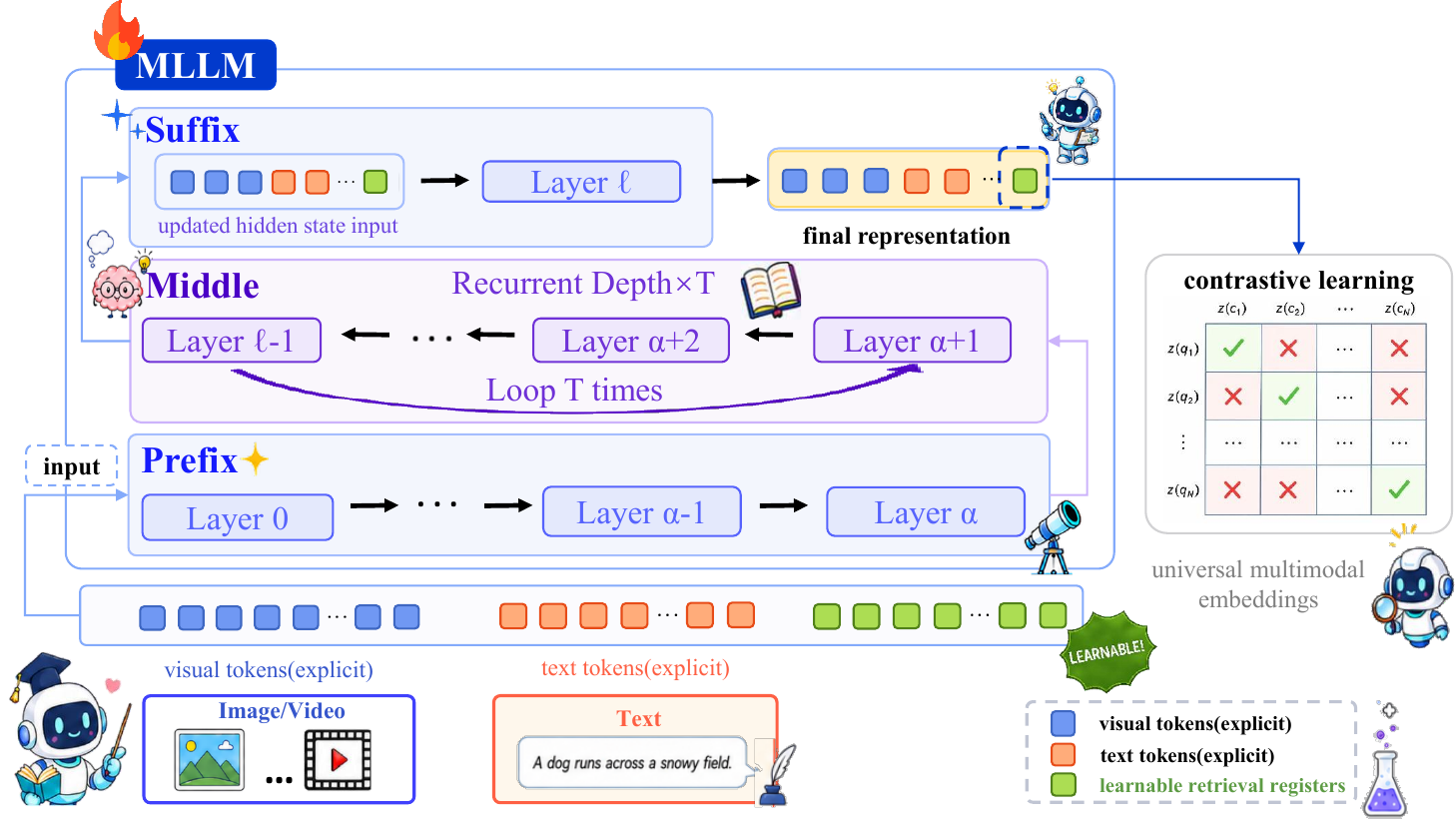}
    \caption{Overview of the recurrent retrieval encoder. Multimodal tokens and
    learnable retrieval registers are processed once by the prefix, iteratively
    refined by a parameter-shared retrieval-forming block for $T$ recurrent
    steps, and mapped by the suffix to a normalized embedding. Query--candidate
    pairs are optimized with terminal contrastive supervision.}
    \label{fig:method-overview}
\end{figure*}

\section{Method}
\label{sec:method}

\subsection{Problem Formulation}

Given positive query--candidate pairs
$\mathcal{D}=\{(q_i,c_i^+)\}_{i=1}^{N}$, a shared UME encoder maps any
supported multimodal input $x$ to a normalized retrieval embedding
\begin{equation}
\begin{array}{@{}l@{}}
\mathbf{z}_{T}(x)=f_{\theta}(x;T)\in\mathbb{S}^{d-1}, \\[2pt]
s_T(q,c)=\mathbf{z}_{T}(q)^{\top}\mathbf{z}_{T}(c),
\end{array}
\label{eq:retrieval-setup}
\end{equation}
where $T$ is the recurrent depth. Queries and candidates share the encoder and
readout rule, and $s_T$ is cosine similarity because the embeddings are
$\ell_2$ normalized.

Single-forward UME forms this embedding in one pass, but our layer-wise
analysis shows a non-uniform progression: the \textbf{Prefix Understanding
Stage} contextualizes the input, the \textbf{Retrieval Formation Stage} forms
retrieval-discriminative features, and the \textbf{Embedding Mapping Stage}
produces the embedding. ReLoop-UME executes the prefix once, recurrently reuses
the parameter-shared retrieval-forming block, and applies the suffix after the
final loop. Learnable Retrieval Registers provide persistent retrieval-specific
states while the token workspace remains fixed (Fig.~\ref{fig:method-overview}).

\subsection{Layer-wise Analysis of UME}

We identify the three stages on the training set using an independently trained
single-forward UME model. Let $\mathbf{H}_{l}(x)$ denote the hidden sequence
after layer $l$, $\rho$ the original retrieval readout position, and
$\widehat{\mathbf{z}}_{l}(x)=\operatorname{norm}_2(\mathbf{H}_{l,\rho}(x))$.
For query $q_i$, let $\mathcal{N}_i$ be its negative candidate pool. We measure
layer-wise retrieval discriminability as
\begin{equation}
\begin{aligned}
&s_{i,l}^{+}
:=
\widehat{\mathbf{z}}_{l}(q_i)^{\top}
\widehat{\mathbf{z}}_{l}(c_i^+), \\[2pt]
&s_{i,l}^{-}
:=
\left\{
\widehat{\mathbf{z}}_{l}(q_i)^{\top}
\widehat{\mathbf{z}}_{l}(c)
\,\middle|\,
c\in\mathcal{N}_i
\right\}, \\[2pt]
&S_l
:=
\frac{1}{N}\sum_{i=1}^{N}
\left[
s_{i,l}^{+}
-
Q_{0.8}\!\left(s_{i,l}^{-}\right)
\right].
\end{aligned}
\label{eq:stage-localization}
\end{equation}
Here $Q_{0.8}$ is the 80th percentile of negative similarities. A positive
margin places the positive above at least 80\% of the negatives, while a larger
$S_l$ indicates stronger retrieval discriminability. Unlike the maximum
negative similarity, it is less dominated by highly similar false negatives.

We locate boundaries from the progression of $S_l$, not its absolute value.
Layers before a sustained rise form the Prefix Understanding Stage; the
contiguous rising interval, ending when separation saturates, forms the
Retrieval Formation Stage; and the remaining terminal layers form the Embedding
Mapping Stage. This places recurrence exactly where retrieval separation is
established, while contextualization and final mapping are computed once. If
this interval is $a{:}b$, layers $0{:}a-1$, $a{:}b$, and $b+1{:}L-1$ define the
prefix, recurrent block, and suffix. The split is fixed before recurrent
training and uses no layer-specific projection head.

This yields layers 0--16, 17--26, and 27 for Qwen2-VL-2B,
Qwen2-VL-7B, and Qwen3-VL-2B, and layers 0--11, 12--22, and 23 for
Qwen3.5-2B. The boundaries are insensitive to the percentile: alternatives to
$0.8$ shift the margin scale but preserve the turning points. Corresponding
visualizations are reported in the appendix.

\subsection{Localized Recurrent Refinement with Learnable Retrieval Registers}

Let $\mathbf{X}(x)\in\mathbb{R}^{n_x\times d}$ be the multimodal token
sequence. We append $M$ Learnable Retrieval Registers
$\mathbf{R}=[\mathbf{r}_1,\ldots,\mathbf{r}_M]\in\mathbb{R}^{M\times d}$,
shared across queries and candidates. Because they follow the input tokens,
each register attends to the complete multimodal input under the causal mask,
and register $m$ also attends to registers $1{:}m-1$. Content-to-register
attention therefore accumulates evidence, residual propagation retains it
across loops, and register-to-register attention exchanges and consolidates it.
Meanwhile, the recurrent block refines the content-token states, so the
registers repeatedly read improved multimodal evidence. The final register
aggregates all preceding registers for stable embedding readout, without
creating new states or increasing sequence length.

Let $F_l$ denote encoder layer $l$, and define
$F_{u:v}=F_v\circ\cdots\circ F_u$. Following the split above, the prefix,
recurrent, and suffix mappings are
$\mathcal{E}=F_{0:a-1}$,
$\mathcal{G}=F_{a:b}$, and
$\mathcal{O}=\operatorname{FinalNorm}\circ F_{b+1:L-1}$, respectively. Let
$\rho_x=n_x+M$ be the position of the final retrieval register. The recurrent
encoder is
\begin{equation}
\begin{aligned}
&\widetilde{\mathbf{X}}(x)
=[\mathbf{X}(x);\mathbf{R}], \\[2pt]
&\mathbf{H}^{(0)}(x)
:=\mathcal{E}\!\left(\widetilde{\mathbf{X}}(x)\right), \\[2pt]
&\mathbf{H}^{(t)}(x)
:=\mathcal{G}\!\left(\mathbf{H}^{(t-1)}(x)\right),
\quad t=1,\ldots,T, \\[2pt]
&\mathbf{z}_{T}(x)
:=\operatorname{norm}_{2}\!\left(
\left[\mathcal{O}\!\left(\mathbf{H}^{(T)}(x)\right)\right]_{\rho_x}
\right).
\end{aligned}
\label{eq:recurrent-encoder}
\end{equation}
The same $\mathcal{G}$ is reused in every loop, each time receiving the complete
preceding hidden sequence with unchanged attention mask, token positions, and
positional indices. The Embedding Mapping Stage usually corresponds only to the
final model layer; accordingly, the suffix in Fig.~\ref{fig:method-overview}
is the final layer in our instantiations.

Only the final register is read out; the preceding $M-1$ registers remain
internal states for accumulating, retaining, and exchanging evidence. They are
not explicit reasoning steps: recurrent depth controls feature refinement while
the workspace stays fixed. Each input undergoes one prefix pass, $T$ shared
retrieval-forming applications, one suffix pass, and terminal normalization.

\subsection{Terminal Contrastive Training}
For a batch $\mathcal{B}=\{(q_i,c_i^+)\}_{i=1}^{B}$, both sides are encoded
with the same recurrent depth. We write
$\mathbf{q}_i=\mathbf{z}_{T}(q_i)$ and
$\mathbf{c}_i=\mathbf{z}_{T}(c_i^+)$, and use the other candidates in the
batch as negatives for $q_i$. With temperature parameter $\tau>0$, the
training objective is
\begin{equation}
\begin{aligned}
    \mathcal{L}_{\mathrm{NCE}}
    = -\frac{1}{B}\sum_{i=1}^{B}
    \log
    \frac{\exp\!\left(\mathbf{q}_i^{\top}\mathbf{c}_i/\tau\right)}
    {\sum_{j=1}^{B}
     \exp\!\left(\mathbf{q}_i^{\top}\mathbf{c}_j/\tau\right)}.
\end{aligned}
\label{eq:terminal-nce}
\end{equation}
The loss is applied only to the embeddings produced after the terminal loop.
Intermediate hidden states still receive gradients through the unrolled
computation, but they are not required to form independently usable retrieval
spaces. This terminal supervision allows the shared recurrent block to optimize
its intermediate transformations for the final query--candidate geometry rather
than treating every loop as a separate exit. The encoder parameters and
retrieval registers are optimized jointly, with $M$, recurrent depth $T$, and
temperature $\tau$ serving as method hyperparameters. 

\subsection{Recurrent Depth and Theoretical Efficiency}
The recurrent depth $T$ is fixed consistently for training and inference. Reusing
the same retrieval-forming parameters deepens feature refinement without changing
model size or token workspace, controlling the accuracy--efficiency trade-off
through the training configuration.

For a general split containing $L_{\mathrm{p}}$ prefix layers,
$L_{\mathrm{r}}$ recurrent layers, and $L_{\mathrm{s}}$ suffix layers, the
decoder-side serial depth is
\begin{equation}
D_{\mathrm{loop}}(T)
:= L_{\mathrm{p}} + T L_{\mathrm{r}} + L_{\mathrm{s}}.
\end{equation}
Relative to single forward encoding, the additional depth is therefore
$(T-1)L_{\mathrm{r}}$, concentrated only in the retrieval-forming block. By
contrast, a token-autoregressive computation that produces $K$ additional
states through the full stack has an approximate serial depth of
$(K+1)(L_{\mathrm{p}}+L_{\mathrm{r}}+L_{\mathrm{s}})$, excluding any
state-transition modules. The recurrent encoder also keeps the sequence length
fixed at $n_x+M$, avoiding token-by-token growth of causal positions and
key--value caches. Its only architecture-specific parameter overhead is the
register matrix, containing $Md$ learnable parameters; all recurrent
applications reuse the same encoder weights and require no separate transition
adapter. Because exact throughput also depends on input length, hardware, and
implementation, we complement this analysis with measured latency and throughput in the experiments.

\section{Experiments}
\label{sec:experiments}

\subsection{Experimental Setup}
\label{sec:experimental-setup}

\paragraph{Datasets and Metrics.}
We evaluate ReLoop-UME on MMEB-V2~\cite{meng2026vlm2vecv2}, which contains 78 tasks spanning 36 image, 18 video, and 24 visual-document (VisDoc) tasks. Following the official protocol, we report Hit@1 for image and video tasks and NDCG@5 for VisDoc tasks. Modality-level \emph{Avg.} scores and the final \emph{All} score are task-level macro averages; all embeddings are $\ell_2$-normalized and ranked by cosine similarity. To further assess generalization in reasoning-intensive scenarios, we directly evaluate the 7B ReLoop-UME trained on MMEB-V2 on MRMR~\cite{zhang2026mrmr}, without MRMR-specific training. Following its standard setting, we use nDCG@10 for all subtasks except \emph{Negation}, for which we report Hit@1.

\paragraph{Baselines and Implementation.}
We compare against UME models based on single forward encoding, including VLM2Vec, GME, VLM2Vec-V2, LamRA, DUME, and BToks~\cite{jiang2025vlm2vec,zhang2025gme,meng2026vlm2vecv2,liu2025lamra,lan2026umer1,sun2026btoks}, together with reasoning-oriented UME-R1 and PLUME~\cite{lan2026umer1,he2026plume}; Table~\ref{tab:mmebv2-main} groups results by model scale. The 2B and 7B ReLoop-UME variants are initialized from the corresponding Qwen2-VL-Instruct backbones~\cite{wang2024qwen2vl}, append five Learnable Retrieval Registers, recurrently reuse layers 17--26, and optimize only the terminal register with InfoNCE at $\tau=0.05$. We train on the official 24 MMEB-V2 training datasets for 5,000 total steps with a global batch size of 256. Videos use eight frames sampled at 2\,FPS, and image/video pixel budgets are restricted to $100{,}352$--$602{,}112$. AdamW uses learning rates of $1\times10^{-5}$ for the language model and multimodal merger and $2\times10^{-6}$ for the vision encoder, $(\beta_1,\beta_2)=(0.9,0.95)$, weight decay $0.1$, gradient clipping at $1.0$, 5\% warm-up, and cosine decay. Training uses BF16, gradient checkpointing, DeepSpeed ZeRO-3, and eight NVIDIA H20 GPUs.

\subsection{Main Results on MMEB-V2}
\label{sec:main-results}

Table~\ref{tab:mmebv2-main} summarizes the MMEB-V2 results at both model scales. At approximately 2B parameters, ReLoop-UME achieves \textbf{63.2} All, outperforming PLUME and UME-R1 by \textbf{1.6} and \textbf{3.1} points, with Image, Video, and VisDoc averages of 68.9, 40.5, and 71.5. At approximately 7B parameters, it reaches \textbf{65.9} All, surpassing UME-R1 by \textbf{1.4} points (65.9 vs.\ 64.5), mainly through higher Image and VisDoc averages (72.0 and 73.9 vs.\ 71.3 and 67.1). Video remains a limitation: ReLoop-UME scores 40.5 at 2B, below PLUME (44.1) and UME-R1 (42.2), and 42.9 at 7B, below UME-R1 (47.5). Recurrent reuse strengthens retrieval-forming features but may underemphasize temporal information across sampled frames, limiting video gains. Overall, recurring over retrieval-forming layers improves the final retrieval representation most clearly on Image and VisDoc tasks, without requiring autoregressive trajectories.

\begin{figure}[t]
    \centering
    \includegraphics[
        width=\columnwidth,
    ]{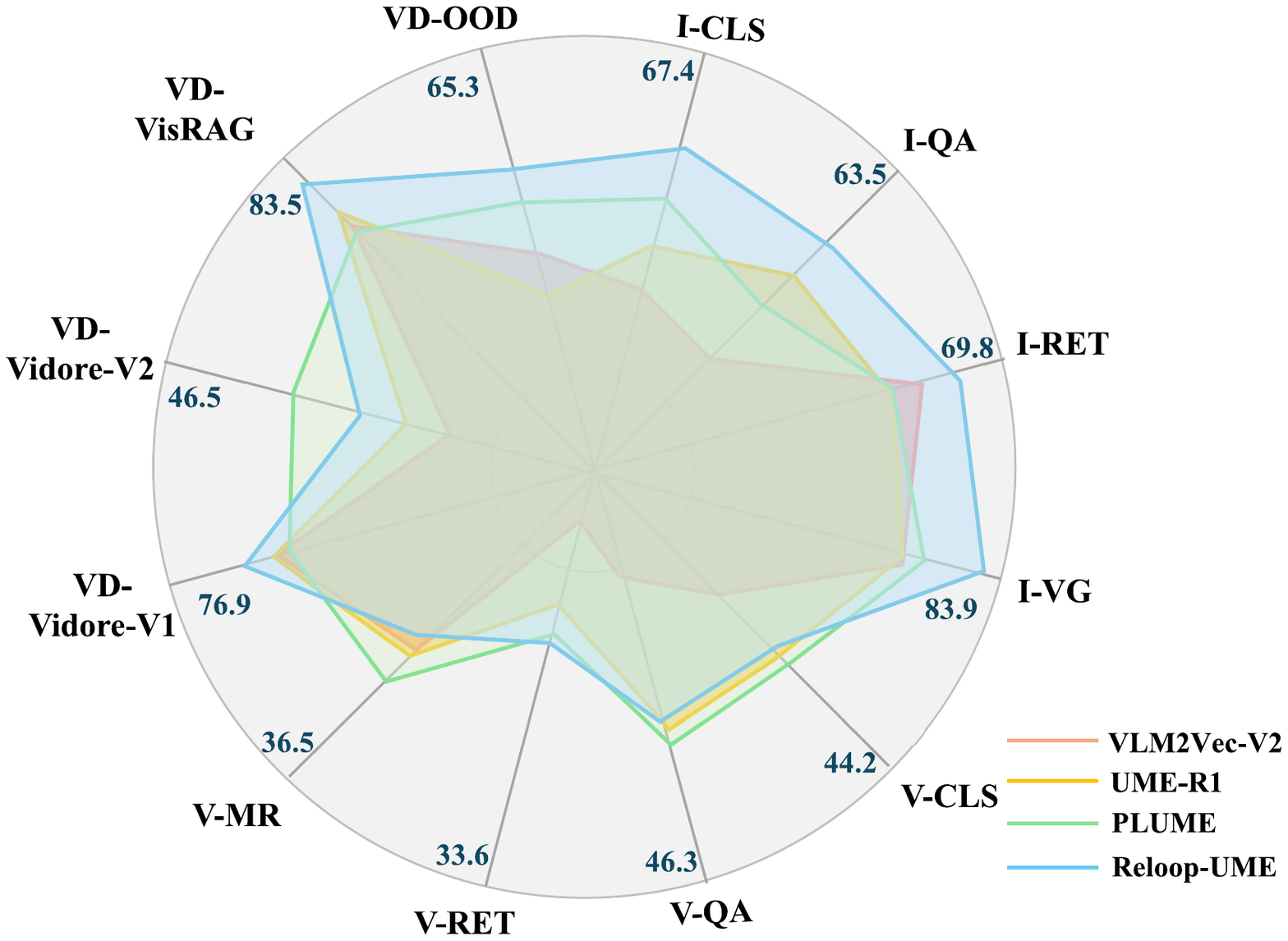}
    \caption{
    Task-group comparison among 2B models on MMEB-V2. ReLoop-UME performs
    strongly on several Image and VisDoc groups, especially visual-document
    retrieval and OOD retrieval, while its Video results are mixed.
    }
    \label{fig:mmebv2-radar}
\end{figure}

\begin{table*}[!t]
\centering
{\small
\setlength{\tabcolsep}{1.2pt}
\renewcommand{\arraystretch}{1.00}
\begin{tabular}{@{}l*{16}{c}@{}}
\toprule
Model
& \multicolumn{5}{c}{Image}
& \multicolumn{5}{c}{Video}
& \multicolumn{5}{c}{VisDoc}
& All \\
\cmidrule(lr){2-6}\cmidrule(lr){7-11}\cmidrule(lr){12-16}
& CLS & QA & RET & GD & Avg.
& CLS & QA & RET & MRET & Avg.
& VDR1 & VDR2 & VR & OOD & Avg.
& \\
\midrule
\# of Datasets
& 10 & 10 & 12 & 4 & 36
& 5 & 5 & 5 & 3 & 18
& 10 & 4 & 6 & 4 & 24
& 78 \\
\midrule
\rowcolor{gray!15}
\multicolumn{17}{c}{\itshape $\sim$ 2B Model Size} \\
LamRA
& 59.2 & 26.5 & \textbf{70.0} & 62.7 & 54.1
& 39.3 & 42.6 & 24.3 & 34.6 & 35.2
& 22.0 & 11.5 & 37.4 & 21.0 & 23.9
& 40.4 \\
VLM2Vec
& 58.7 & 49.3 & 65.0 & 72.9 & 59.7
& 33.4 & 30.5 & 20.6 & 33.0 & 29.0
& 49.8 & 13.5 & 51.8 & 33.5 & 41.6
& 47.0 \\
GME
& 54.4 & 29.9 & 66.9 & 55.5 & 51.9
& 34.9 & 42.0 & 25.6 & 32.4 & 33.9
& \textbf{86.1} & \textbf{54.0} & \underline{82.5} & 43.1 & \textbf{72.7}
& 54.1 \\
VLM2Vec-V2
& 62.9 & 56.3 & 69.5 & 77.3 & 64.9
& 39.3 & 34.3 & 28.8 & 38.5 & 34.9
& 75.5 & 44.9 & 79.4 & 39.4 & 65.4
& 58.0 \\
DUME
& 59.3 & 55.0 & 66.3 & 78.0 & 62.5
& 37.7 & 46.6 & 17.1 & 30.0 & 33.2
& 67.6 & 43.3 & 47.1 & 33.8 & 52.8
& 52.7 \\
BToks
& 64.3 & 59.8 & 68.8 & 77.4 & 66.0
& 43.7 & 47.0 & 33.0 & 33.6 & 39.9
& 71.1 & 38.6 & 81.3 & 38.1 & 62.7
& 59.0 \\
UME-R1
& 64.8 & \underline{62.8} & 67.6 & 77.2 & \underline{66.6}
& \underline{44.3} & \underline{51.2} & 32.9 & \underline{39.7} & \underline{42.2}
& 72.4 & 46.2 & 79.2 & 37.2 & 63.9
& 60.1 \\
PLUME
& \underline{66.5} & 59.2 & 67.6 & \underline{79.7} & 66.3
& \textbf{45.0} & \textbf{52.3} & \underline{33.5} & \textbf{46.7} & \textbf{44.1}
& 72.1 & \underline{49.8} & 78.1 & \underline{57.4} & 67.5
& \underline{61.6} \\
\textbf{ReLoop-UME (Ours)}
& \textbf{67.4} & \textbf{63.5} & \underline{69.8} & \textbf{83.9} & \textbf{68.9}
& 44.2 & 46.3 & \textbf{33.6} & 36.5 & 40.5
& \underline{76.9} & 46.5 & \textbf{83.5} & \textbf{65.3} & \underline{71.5}
& \textbf{63.2} \\
\midrule
\rowcolor{gray!15}
\multicolumn{17}{c}{\itshape $\sim$ 7B Model Size} \\
ColPali-v1.3
& 40.3 & 11.5 & 48.1 & 40.3 & 34.9
& 26.7 & 37.8 & 21.6 & 25.5 & 28.2
& 83.6 & 52.0 & 81.1 & 43.1 & 71.0
& 44.4 \\
LamRA
& 51.7 & 34.1 & 66.9 & 56.7 & 52.4
& 32.9 & 42.6 & 23.2 & 37.6 & 33.7
& 56.3 & 33.3 & 58.2 & 40.1 & 50.2
& 47.4 \\
VLM2Vec
& 62.7 & 56.9 & 69.4 & 82.2 & 65.5
& 39.1 & 30.0 & 29.0 & \textbf{40.6} & 34.0
& 56.9 & 9.4 & 59.1 & 38.1 & 46.4
& 52.3 \\
GME
& 57.7 & 34.7 & 71.2 & 59.3 & 56.0
& 37.4 & 50.4 & 28.4 & 38.2 & 38.6
& \textbf{89.4} & 55.6 & \underline{85.0} & \underline{44.4} & \textbf{75.2}
& 57.8 \\
VLM2Vec-V2
& 65.7 & 61.5 & 70.0 & 85.2 & 68.1
& \underline{45.9} & 33.9 & 27.6 & 39.3 & 36.4
& 78.8 & 52.6 & 82.7 & 42.1 & 69.3
& 61.2 \\
DUME
& 64.2 & 57.0 & 70.8 & 81.8 & 66.4
& 32.9 & 47.4 & 8.6 & 28.0 & 29.4
& 67.1 & 35.2 & 82.6 & 34.9 & 60.3
& 55.9 \\
CAFe
& 63.6 & 61.7 & 69.1 & 87.6 & 67.6
& 35.8 & \underline{58.7} & \underline{34.4} & 39.5 & 42.4
& 70.7 & 49.6 & 79.5 & 38.1 & 63.9
& 60.6 \\
UniME-V2
& 65.6 & \underline{68.7} & \textbf{73.1} & \textbf{90.9} & \underline{71.8}
& 37.2 & 50.6 & 28.9 & \underline{39.6} & 39.0
& 61.8 & 42.0 & 70.5 & 37.9 & 56.7
& 59.6 \\
LCO-Emb
& 56.3 & 16.9 & 52.2 & 57.0 & 44.0
& 39.3 & 57.6 & 24.8 & 26.5 & 38.2
& 80.4 & \underline{56.4} & 79.7 & 41.8 & 69.8
& 50.6 \\
Omni-Embed
& 46.0 & 20.5 & 58.0 & 52.9 & 43.7
& 40.5 & 44.3 & 32.7 & 25.6 & 36.9
& \underline{85.7} & \textbf{61.3} & 84.3 & 43.3 & \underline{74.2}
& 51.5 \\
UME-R1
& \underline{67.1} & \textbf{69.2} & 71.9 & 84.9 & 71.3
& \textbf{48.6} & \textbf{60.7} & \textbf{38.2} & 39.3 & \textbf{47.5}
& 75.7 & 50.5 & 83.7 & 37.6 & 67.1
& \underline{64.5} \\
\textbf{ReLoop-UME (Ours)}
& \textbf{69.5} & 67.5 & \underline{72.5} & \underline{87.8} & \textbf{72.0}
& 44.5 & 52.0 & 34.3 & 39.4 & \underline{42.9}
& 78.9 & 50.6 & \textbf{85.7} & \textbf{67.1} & 73.9
& \textbf{65.9} \\
\bottomrule
\end{tabular}
}
\caption{Results on the full MMEB-V2 benchmark. CLS: classification, QA: question answering, RET: retrieval, GD: grounding, MRET: moment retrieval, VDR: ViDoRe, VR: VisRAG, and OOD: out-of-distribution. \emph{Avg.} and \emph{All} denote modality-level and 78-task macro averages, respectively. Best and second-best results within each model-size group are bolded and underlined.}
\label{tab:mmebv2-main}
\end{table*}

\subsection{Efficiency and Accuracy--Efficiency Tradeoff}
\label{sec:efficiency}

Table~\ref{tab:efficiency} provides a quantitative breakdown of
inference costs. All measurements are conducted on a single NVIDIA
H20 GPU. For each method, we randomly sample 500 inputs per modality
as one evaluation set, preceded by 20 warm-up iterations, and repeat
the experiment with five independently drawn evaluation sets. The
table reports the mean across the five runs; $\pm$ denotes the
standard deviation.

\begin{figure}[!t]
    \centering

    % Replace the placeholder with:
    \includegraphics[width=\linewidth]{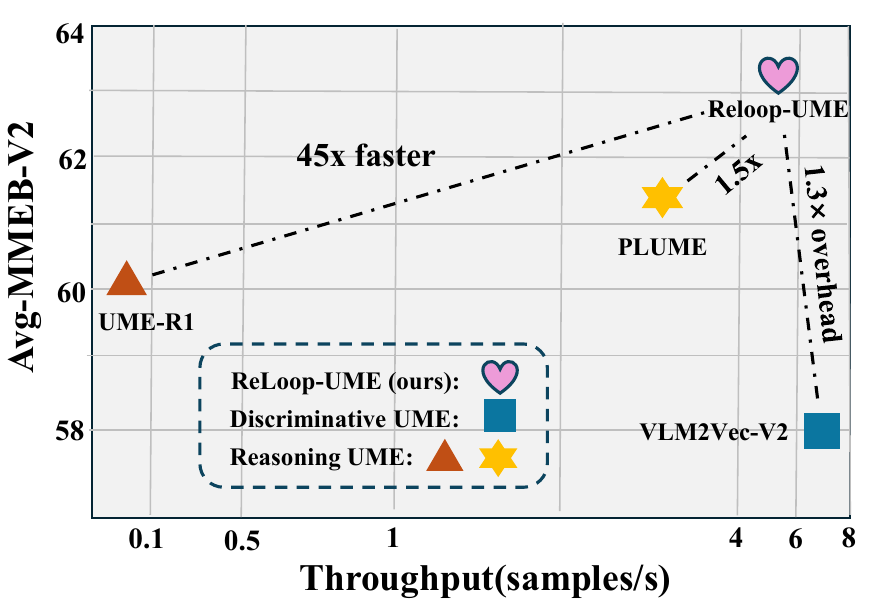}

    \caption{
    ReLoop-UME achieves a favorable accuracy--efficiency tradeoff
    on MMEB-V2. The x-axis shows inference throughput on a single
    H20 GPU, and the y-axis shows the MMEB-V2 All score.
    }
    \label{fig:accuracy_efficiency}
\end{figure}

\begin{table}[!t]
    \centering
    \small
    \setlength{\tabcolsep}{1pt}
    \renewcommand{\arraystretch}{0.96}
    \begin{tabular}{@{}lcccc@{}}
        \toprule
        Metric
        & PLUME
        & UME-R1
        & ReLoop-UME \\
        \midrule
        Steps
        & 8
        & 403
        & 4 \\
        Latency (ms/sample)
        & $298\pm12$
        & $9023\pm187$
        & $201\pm4$ \\
        Throughput (samples/s)
        & $3.3\pm0.1$
        & $0.11\pm0.01$
        & $5.0\pm0.1$ \\
        \midrule
        Speedup
        & $30.3\times$
        & $1.0\times$
        & $44.9\times$ \\
        Overhead
        & $1.9\times$
        & --
        & $1.3\times$ \\
        \bottomrule
    \end{tabular}
    \caption{
    Inference efficiency on a single H20 GPU.
    Steps denotes the number of reasoning tokens for PLUME and UME-R1,
    and the number of recurrent loop iterations for ReLoop-UME.
    Speedup is measured relative to UME-R1, while overhead is measured
    relative to VLM2Vec-V2.
    }
    \label{tab:efficiency}
\end{table}

ReLoop-UME performs additional computation through four recurrent
applications of layers 17--26 rather than autoregressively generating
reasoning states. It reduces per-sample latency from 9023 ms for
UME-R1 to 201 ms, corresponding to a 44.9$\times$ speedup, and is
1.5$\times$ faster than PLUME. Relative to VLM2Vec-V2 with single forward encoding (156 ms), ReLoop-UME introduces only
1.3$\times$ inference overhead while improving the MMEB-V2
All score by 5.2 points (63.2 vs.\ 58.0). These results demonstrate
that recurrent depth adds modest overhead over single forward encoding while
\mbox{avoiding} the severe latency cost of autoregressive reasoning.

Figure~\ref{fig:accuracy_efficiency} visualizes the
accuracy--efficiency tradeoff across recurrent depths. The default
$T=4$ configuration reaches an All score of 63.2 at 5.0 samples/s,
providing the measured reference point used in Table~\ref{tab:efficiency}.

\subsection{Ablation Studies}
\label{sec:ablation}

We ablate the recurrent interval, register count $M$, and recurrent depth $T$
on Qwen2-VL-2B. Unless noted, the default uses layers 17--26, $M=5$,
and $T=4$ under the same training recipe.

\begin{table}[t]
    \centering
    \setlength{\tabcolsep}{3pt}
    \renewcommand{\arraystretch}{0.92}
    \begin{tabular}{@{}lccccc@{}}
        \toprule
        Layers & Decoder apps. & Image & Video & VisDoc & All \\
        \midrule
        0--16 & 79  & 66.0 & 39.4 & 67.0 & 60.2 \\
        0--27 & 112 & 67.1 & 39.8 & 68.8 & 61.3 \\
        17--27 & 61 & 68.1 & 40.2 & 70.2 & 62.3 \\
        27 & 31 & 65.2 & 39.0 & 66.2 & 59.5 \\
        17--26 & 58 & \textbf{68.9} & \textbf{40.5} & \textbf{71.5} & \textbf{63.2} \\
        \bottomrule
    \end{tabular}
    \caption{Ablation on the recurrent layer interval. We report decoder applications and MMEB-V2 performance with the register count and recurrent depth fixed to $M=5$ and $T=4$.}
    \label{tab:ablation_range}
\end{table}

\begin{table}[t]
    \centering
    \setlength{\tabcolsep}{3pt}
    \renewcommand{\arraystretch}{0.92}
    \begin{tabular}{@{}cccccc@{}}
        \toprule
        $M$ & Added params. & Image & Video & VisDoc & All \\
        \midrule
        0  & 0      & 67.5 & 39.9 & 70.4 & 61.8 \\
        5  & 7,680  & \textbf{68.9} & 40.5 & \textbf{71.5} & \textbf{63.2} \\
        8  & 12,288 & 68.5 & 40.6 & 71.2 & 62.9 \\
        10 & 15,360 & 68.0 & \textbf{40.8} & 70.8 & 62.7 \\
        \bottomrule
    \end{tabular}
    \caption{Ablation on the number of retrieval registers $M$. We report the added trainable parameters and MMEB-V2 performance using recurrent layers 17--26 with $T=4$.}
    \label{tab:ablation_registers}
\end{table}

\begin{table}[t]
    \centering
    \setlength{\tabcolsep}{3pt}
    \renewcommand{\arraystretch}{0.92}
    \begin{tabular}{@{}cccccc@{}}
        \toprule
        $T$ & Decoder apps. & Image & Video & VisDoc & All \\
        \midrule
        1 & 28 & 66.5 & 39.7 & 67.5 & 60.6 \\
        2 & 38 & 68.0 & 40.2 & 69.8 & 62.1 \\
        4 & 58 & \textbf{68.9} & \textbf{40.5} & \textbf{71.5} & \textbf{63.2} \\
        8 & 98 & 68.7 & 40.4 & 70.8 & 62.8 \\
        \bottomrule
    \end{tabular}
    \caption{Ablation on recurrent depth $T$. Each configuration uses matching training and inference depth, recurrent layers 17--26, and $M=5$.}
    \label{tab:ablation_depth}
\end{table}

\paragraph{Recurrent interval.}
Table~\ref{tab:ablation_range} tests where recurrence should be placed. Layers
17--26 achieve the best All score of 63.2, surpassing early-layer, full-decoder,
extended-late, and output-only recurrence by 3.0, 1.9, 0.9, and 3.7 points.
More computation alone is insufficient: full-decoder recurrence requires 112
decoder applications, nearly twice the 58 of layers 17--26, yet performs worse;
the final layer alone is also inadequate. This supports the layer-wise pattern
in Fig.~\ref{fig:layer_discriminability}: retrieval geometry forms within a
localized intermediate-to-late stage, making that stage the most effective
target for recurrent reuse.

\paragraph{Retrieval registers.}
Table~\ref{tab:ablation_registers} evaluates the persistent workspace across
loops. Removing the registers lowers All from 63.2 to 61.8, while five registers
recover a 1.4-point gain with only 7,680 additional parameters. Expanding the
workspace to eight or ten registers yields no further improvement. The benefit
therefore comes from preserving a small set of retrieval-specific states, not
from increasing capacity or sequence length. This compact fixed-size workspace
retains and refines multimodal evidence without the latency and memory growth
of token-expansion reasoning.

\paragraph{Recurrent depth.}
Table~\ref{tab:ablation_depth} confirms that iterative refinement across recurrent applications drives the gain. Increasing $T$ from 1 to 2 improves All from
60.6 to 62.1, and $T=4$ reaches 63.2; extending to $T=8$ slightly drops to 62.8
despite substantially more decoder applications. The benefit thus saturates at
a moderate depth instead of growing monotonically with computation. Together,
the ablations close the loop from observation to design: localize the
retrieval-forming stage, recurrently refine it for a moderate number of steps,
and preserve evidence in a compact fixed-size workspace.

\subsection{Generalization Experiments}
\label{sec:generalization}

\paragraph{Cross-dataset generalization.}
We zero-shot evaluate the 7B ReLoop-UME trained on MMEB-V2 on MRMR.
Table~\ref{tab:mrmr-summary} reports group means and the official Avg. against
the four strongest directly comparable baselines; full per-subtask results are deferred
to the appendix. We exclude Qwen3-VL-Embedding~\cite{li2026qwen3vlembedding}
because substantially different training data prevent a controlled zero-shot comparison.

\begin{table}[t]
    \centering
    \small
    \setlength{\tabcolsep}{1.0pt}
    \renewcommand{\arraystretch}{0.96}
    \begin{tabular}{@{}lcccc@{}}
        \toprule
        Model & Knowledge & Theorem & Contra. & Avg. \\
        \midrule
        Ops-MM-Embedding  & 67.4 & 37.4 & 36.6 & 48.1 \\
        NV-Embed-v2              & 62.5 & 36.7 & 32.3 & 44.8 \\
        MM-Embed                 & 61.2 & 31.7 & 21.9 & 39.8 \\
        GME & 51.0 & 28.9 & 36.9 & 39.1 \\
        \midrule
        \textbf{ReLoop-UME (7B)}
        & \textbf{71.6}
        & \textbf{39.1}
        & 30.7
        & \textbf{48.6} \\
        \bottomrule
    \end{tabular}
    \caption{
    Zero-shot transfer from MMEB-V2 to MRMR. Knowledge, Theorem, and
    Contra.\ are task-group means; Contra.\ abbreviates Contradiction,
    and Avg.\ averages all 11 subtasks.
    }
    \label{tab:mrmr-summary}
\end{table}

ReLoop-UME leads Knowledge (71.6), Theorem (39.1), and Avg. (48.6),
0.5 points above Ops-MM-Embedding. \mbox{It does not lead} on Contradiction, however,
and the overall margin is modest. The results therefore indicate only a degree of
transfer and generalization from MMEB-V2 to MRMR, rather than uniform gains
across all task groups.

\begin{center}
    \setlength{\tabcolsep}{2.4pt}
    \renewcommand{\arraystretch}{0.96}
    \begin{tabular}{@{}lccc@{}}
        \toprule
        Backbone & \shortstack{Single forward} & ReLoop-UME & $\Delta$ \\
        \midrule
        Qwen2-VL-2B & 60.6 & 63.2 & 2.6 \\
        Qwen3-VL-2B & 61.7 & 63.9 & 2.2 \\
        Qwen3.5-2B  & 60.2 & 62.6 & 2.4 \\
        \bottomrule
    \end{tabular}
    \captionof{table}{Backbone generalization on MMEB-V2 (All).}
    \label{tab:backbone-generalization}
\end{center}

\paragraph{Backbone generalization.}
Mainstream UME models are commonly built on Qwen-VL backbones. We therefore
select three representative variants~\cite{wang2024qwen2vl,bai2025qwen3vl,qwen2026qwen35} and apply the same stage-localization and recurrent-training recipe (Table~\ref{tab:backbone-generalization}). Single forward uses $T=1$ and $M=0$. The gains support the recipe beyond Qwen2-VL. This suggests that the recipe can transfer across different Qwen-VL backbones and model scales. 

\section{Conclusion}
\label{sec:conclusion}

We introduced ReLoop-UME, which recurrently reuses retrieval-forming layers
with shared parameters while keeping the token workspace fixed. Learnable
Retrieval Registers preserve multimodal evidence across loops, directing extra
computation to the stage where retrieval geometry forms instead of autoregressive
token expansion. Across MMEB-V2, zero-shot MRMR transfer, and different backbones, ReLoop-UME
is 44.9$\times$ faster than UME-R1 and 1.5$\times$ faster than PLUME, with
1.3$\times$ latency overhead over VLM2Vec-V2. Ablations confirm stage localization,
moderate recurrent depth, and a compact register workspace, supporting recurrent
feature refinement as an efficient route to universal multimodal embeddings.

% -----------------------------------------------------------------------------
% Supplementary material (merged into the same arXiv source)
% -----------------------------------------------------------------------------
\clearpage
\microtypesetup{reactivate}
\setlength{\textfloatsep}{5pt plus 1pt minus 1pt}
\setlength{\dbltextfloatsep}{5pt plus 1pt minus 1pt}
\setlength{\floatsep}{4pt plus 1pt minus 1pt}
\setlength{\intextsep}{4pt plus 1pt minus 1pt}
\renewcommand{\topfraction}{0.97}
\renewcommand{\bottomfraction}{0.92}
\renewcommand{\textfraction}{0.02}
\renewcommand{\floatpagefraction}{0.90}
\renewcommand{\dbltopfraction}{0.97}
\renewcommand{\dblfloatpagefraction}{0.90}
\makeatletter
\setlength{\@fptop}{0pt}
\setlength{\@dblfptop}{0pt}
\makeatother

\newcommand{\method}{ReLoop-UME}
\newcommand{\qwen}{Qwen}

\begin{center}
\vspace*{0.38in}
{\LARGE\bf Supplementary Material for\\
ReLoop-UME: Recurrent Depth with Learnable Retrieval Registers for Universal Multimodal Embedding\par}
\vspace{0.24in}
\end{center}

\appendix
\setcounter{figure}{0}
\setcounter{table}{0}
\setcounter{equation}{0}

\section{Supplementary Overview}
\label{app:overview}

This supplement provides benchmark details, complete evaluation tables,
reproducibility settings, layer-wise analyses, and controlled ablations that
support the main paper.  The organization follows the main claim: retrieval
structure is formed non-uniformly across depth; the corresponding middle-to-late
stage can be localized reproducibly; recurrent reuse of that stage improves the
terminal embedding; and Learnable Retrieval Registers are most useful when they
can preserve evidence across multiple recurrent applications.

\begin{center}
\footnotesize
\setlength{\tabcolsep}{3.2pt}
\renewcommand{\arraystretch}{1.02}
\begin{tabular}{@{}lp{0.73\columnwidth}@{}}
\toprule
\textbf{Section} & \textbf{Supplementary evidence} \\
\midrule
B & MMEB-V2 task inventory, MRMR statistics, and training sources. \\
C & Deterministic stage localization, recurrent encoding, optimization, and software configuration. \\
D & Complete 78-task MMEB-V2 results for four ReLoop-UME backbones. \\
E & Twelve layer-wise curves covering four quantiles and three Qwen-VL backbones. \\
F & One complete two-column MRMR table using the official MRMR baseline results. \\
G & Backbone architectures and matched cross-backbone interpretation. \\
H--J & Controlled ablations, eleven qualitative cases, remaining limitations, and targeted future directions. \\
\bottomrule
\end{tabular}
\captionof{table}{Guide to the supplementary evidence.}
\label{tab:supp-map}
\end{center}

\section{Datasets and Evaluation Protocols}
\label{app:datasets}

The primary experiments use MMEB-V2~\cite{supp_meng2026vlm2vecv2}, which contains 78
tasks: 36 image, 18 video, and 24 visual-document tasks.  MRMR~\cite{supp_zhang2026mrmr}
is used only for zero-shot transfer of the 7B checkpoint trained on the MMEB-V2
mixture.  All embeddings are $\ell_2$-normalized and ranked with cosine
similarity.  MMEB-V2 reports Hit@1 for image and video tasks and nDCG@5 for
visual-document tasks.  MRMR reports nDCG@10 except for Negation, which uses
Hit@1.

\subsection{MMEB-V2 Task Inventory}

\begin{center}
\footnotesize
\setlength{\tabcolsep}{3pt}
\renewcommand{\arraystretch}{1.01}
\begin{tabular}{@{}lcp{0.63\columnwidth}@{}}
\toprule
\textbf{Image family} & \textbf{\#} & \textbf{Datasets / configurations} \\
\midrule
I-CLS & 10 & ImageNet-1K, N24News, HatefulMemes, VOC2007, SUN397, Places365, ImageNet-A, ImageNet-R, ObjectNet, Country211. \\
I-QA & 10 & OK-VQA, A-OKVQA, DocVQA, InfographicsVQA, ChartQA, Visual7W, ScienceQA, VizWiz, GQA, TextVQA. \\
I-RET & 12 & VisDial, CIRR, VisualNews (t2i/i2t), MSCOCO (t2i/i2t), NIGHTS, WebQA, FashionIQ, Wiki-SS-NQ, OVEN, EDIS. \\
I-GD & 4 & MSCOCO grounding, RefCOCO, RefCOCO-Matching, Visual7W-Pointing. \\
\bottomrule
\end{tabular}
\captionof{table}{The 36 image tasks in MMEB-V2.  Directional retrieval configurations are counted separately.}
\label{tab:mmeb-image}
\end{center}

\begin{center}
\footnotesize
\setlength{\tabcolsep}{3pt}
\renewcommand{\arraystretch}{0.93}
\begin{tabular}{@{}lcp{0.63\columnwidth}@{}}
\toprule
\textbf{Family} & \textbf{\#} & \textbf{Datasets / configurations} \\
\midrule
V-CLS & 5 & Kinetics-700, Something-Something V2, HMDB51, UCF101, Breakfast. \\
V-QA & 5 & MVBench, Video-MME, NExTQA, EgoSchema, ActivityNetQA. \\
V-RET & 5 & DiDeMo, MSR-VTT, MSVD, VATEX, YouCook2. \\
V-MRET & 3 & QVHighlight, Charades-STA, MomentSeeker. \\
\midrule
VD-VDR1 & 10 & ArxivQA, DocVQA, InfoVQA, TabFQuAD, TAT-DQA, Shift Project, AI, Energy, Government, and Healthcare reports. \\
VD-VDR2 & 4 & CSG reports (human-labeled V2), Biomedical Lectures V2, Economics Reports V2, CSG Reports V2 multilingual. \\
VD-VR & 6 & ArxivQA, ChartQA, MP-DocVQA, SlideVQA, InfoVQA, PlotQA. \\
VD-OOD & 4 & ViDoSeek-page/document and MMLongBench-page/document. \\
\bottomrule
\end{tabular}
\captionof{table}{The 18 video and 24 visual-document tasks in MMEB-V2.}
\label{tab:mmeb-video-doc}
\end{center}

\subsection{MRMR Taxonomy and Data Statistics}

MRMR contains interleaved image--text queries and mixed-modality corpus
documents.  Knowledge retrieval tests expert evidence matching; Theorem retrieval
requires inferring an abstract principle from a concrete multimodal problem; and
Contradiction retrieval requires identifying a conflicting statement or violated
rule.  Table~\ref{tab:mrmr-statistics} summarizes the official task sizes.

\begin{table*}[!t]
\centering
\footnotesize
\setlength{\tabcolsep}{4.3pt}
\renewcommand{\arraystretch}{0.96}
\begin{tabular}{@{}llrrrrp{0.34\textwidth}@{}}
\toprule
\textbf{Group} & \textbf{Subtask} & \textbf{Queries} & \textbf{Docs} & \textbf{Avg. $D^+$} & \textbf{Metric} & \textbf{Retrieval requirement} \\
\midrule
Knowledge & Art & 157 & 26,223 & 1.8 & nDCG@10 & Expert-domain evidence for visual and interleaved questions. \\
& Medicine & 167 & 26,223 & 2.2 & nDCG@10 & Pathological, diagnostic, and biomedical evidence. \\
& Science & 137 & 26,223 & 1.8 & nDCG@10 & Scientific concepts grounded in expert images. \\
& Humanities & 94 & 26,223 & 1.9 & nDCG@10 & Historical, cultural, and social-science evidence. \\
\midrule
Theorem & Math & 72 & 14,257 & 2.6 & nDCG@10 & Theorem or solved-problem retrieval for calculation questions. \\
& Physics & 107 & 14,257 & 2.1 & nDCG@10 & Infer the governing physical principle from the problem. \\
& Engineering & 236 & 14,257 & 2.1 & nDCG@10 & Mechanical, electronic, and computer-science principles. \\
& Business & 164 & 14,257 & 2.2 & nDCG@10 & Finance, economics, marketing, and related principles. \\
\midrule
Contradiction & Negation & 200 & 4 & 1.0 & Hit@1 & Select the unique description contradicting the image. \\
& Design & 88 & 700 & 1.0 & nDCG@10 & Retrieve the requirement violated by a vehicle design. \\
& Traffic & 80 & 796 & 1.8 & nDCG@10 & Retrieve the driving rule contradicted by the depicted case. \\
\bottomrule
\end{tabular}
\caption{MRMR task statistics and retrieval requirements, adapted from the official benchmark.  Knowledge subtasks share one corpus, and Theorem subtasks share another.}
\label{tab:mrmr-statistics}
\end{table*}

\subsection{Training-Source Taxonomy}

The model is trained on the official 24 MMEB-V2 training datasets.  A source
dataset and a retrieval configuration are not always identical: MSCOCO and
VisualNews contribute both text-to-image and image-to-text directions, while
video sources may contribute caption retrieval, question answering, and
text-to-video retrieval.  The inventory below follows the task organization used
by the supplied UME LaTeX sources.

\begin{center}
\footnotesize
\setlength{\tabcolsep}{3pt}
\renewcommand{\arraystretch}{1.02}
\begin{tabular}{@{}p{0.22\columnwidth}p{0.68\columnwidth}@{}}
\toprule
\textbf{Family} & \textbf{Source / configuration inventory} \\
\midrule
Image-centered & A-OKVQA, CIRR, ChartQA, DocVQA, HatefulMemes, ImageNet-1K, InfographicsVQA, MSCOCO matching and i2t/t2i, N24News, NIGHTS, OK-VQA, SUN397, VOC2007, Visual7W, VisDial, VisualNews i2t/t2i, WebQA. \\
Video-centered & LLaVA-Hound caption retrieval, video QA, and text-to-video retrieval. \\
Visual documents & ViDoRe and VisRAG training pairs. \\
\bottomrule
\end{tabular}
\captionof{table}{Training-source taxonomy.  Training always uses the official 24-dataset MMEB-V2 mixture rather than MRMR examples.}
\label{tab:training-sources}
\end{center}

\section{Experimental and Reproducibility Details}
\label{app:reproducibility}

\subsection{Problem Setting}

Given positive query--candidate pairs $\mathcal{D}=\{(q_i,c_i^+)\}_{i=1}^{N}$,
a shared multimodal encoder maps every supported input $x$ to a normalized
embedding
\begin{equation}
\mathbf{z}_{T}(x)=f_{\theta}(x;T)\in\mathbb{S}^{d-1},
\qquad s_T(q,c)=\mathbf{z}_{T}(q)^{\top}\mathbf{z}_{T}(c),
\label{eq:app-retrieval}
\end{equation}
where $T$ is recurrent depth.  Queries and candidates use the same encoder,
registers, recurrent interval, and terminal readout.

\subsection{Deterministic Localization of Stage Boundaries}
\label{app:localization}

Stage localization is performed on an independently trained single-forward UME
checkpoint before recurrent training.  Let $\widehat{\mathbf z}_{l}(x)$ be the
$\ell_2$-normalized readout at physical decoder layer $l$.  For negative quantile
$q$, the layer-wise separation score is
\begin{equation}
S_l^{(q)}=\frac{1}{N}\sum_{i=1}^{N}
\left[s_{i,l}^{+}-Q_q\!\left(s_{i,l}^{-}\right)\right].
\label{eq:quantile-score}
\end{equation}
The main paper uses $q=0.8$ because it emphasizes hard distractors without the
instability of a single maximum negative.  To convert the qualitative notions of
``sustained rise'' and ``saturation'' into a reproducible boundary rule, we use
the following fixed procedure.

First, compute $S_l^{(q)}$ for $q\in\{0.5,0.6,0.7,0.8\}$ on the same probe set and
normalize each curve to $[0,1]$ over physical layers.  Second, enumerate every
ordered pair of breakpoints $(a,b)$ satisfying a minimum prefix length of four
layers, a minimum formation interval of four layers, and at least one terminal
mapping layer.  For each pair, fit independent least-squares lines to the three
contiguous segments $[0,a-1]$, $[a,b]$, and $[b+1,L-1]$ for every quantile and sum
the residual errors over quantiles.  We retain only candidates whose middle
segment has positive slope and a larger slope than both the prefix and terminal
segments.  The selected $(a,b)$ minimizes the pooled residual; ties are resolved
by the smallest worst-quantile residual and then the shorter middle interval.
This is a three-segment change-point fit with fixed constraints, not a manually
chosen score threshold.

The resulting middle segment is the Retrieval Formation Stage.  Layers before
$a$ are the Prefix Understanding Stage, and layers after $b$ are the Embedding
Mapping Stage.  Applying the same rule across backbones and quantiles gives the
boundaries in Table~\ref{tab:quantile-boundaries}.  The value $q=0.8$ therefore
affects the vertical margin scale but not the operational interval.

\begin{center}
\footnotesize
\setlength{\tabcolsep}{2.8pt}
\renewcommand{\arraystretch}{1.02}
\begin{tabular}{@{}lccccp{0.19\columnwidth}@{}}
\toprule
\textbf{Backbone} & $q=.5$ & $q=.6$ & $q=.7$ & $q=.8$ & \textbf{Stages} \\
\midrule
Qwen2-VL-2B & 17--26 & 17--26 & 17--26 & 17--26 & 0--16 / 17--26 / 27 \\
Qwen3-VL-2B & 17--26 & 17--26 & 17--26 & 17--26 & 0--16 / 17--26 / 27 \\
Qwen3.5-2B & 12--22 & 12--22 & 12--22 & 12--22 & 0--11 / 12--22 / 23 \\
\bottomrule
\end{tabular}
\captionof{table}{Quantile sensitivity of the selected interval.  The final column lists Prefix / Retrieval Formation / Embedding Mapping layers.  Qwen2-VL-7B uses the same 0--16 / 17--26 / 27 partition reported in the main paper.}
\label{tab:quantile-boundaries}
\end{center}

\subsection{Recurrent Retrieval Registers}

Let $\mathbf{X}(x)\in\mathbb{R}^{n_x\times d}$ be the multimodal token sequence
and $\mathbf{R}=[\mathbf r_1,\ldots,\mathbf r_M]\in\mathbb{R}^{M\times d}$ be
$M$ learnable retrieval registers.  If $\mathcal E$, $\mathcal G$, and
$\mathcal O$ denote the prefix, retrieval-forming block, and terminal mapping,
respectively,
\begin{equation}
\begin{aligned}
\widetilde{\mathbf X}(x)&=[\mathbf X(x);\mathbf R],\\
\mathbf H^{(0)}(x)&=\mathcal E(\widetilde{\mathbf X}(x)),\\
\mathbf H^{(t)}(x)&=\mathcal G(\mathbf H^{(t-1)}(x)),\quad t=1,\ldots,T,\\
\mathbf z_T(x)&=\operatorname{norm}_2\!\left(
[\mathcal O(\mathbf H^{(T)}(x))]_{\rho_x}\right).
\end{aligned}
\label{eq:app-recurrent}
\end{equation}
The parameters of $\mathcal G$ are shared across loops.  The complete hidden
sequence is propagated with unchanged masks and positional indices.  Only the
terminal register is used as the embedding readout; the earlier registers serve
as persistent retrieval-specific workspace.

\subsection{Objective and Training Configuration}

For a batch $\mathcal B=\{(q_i,c_i^+)\}_{i=1}^{B}$, terminal embeddings are
optimized with in-batch InfoNCE:
\begin{equation}
\mathcal L_{\mathrm{NCE}}=-\frac{1}{B}\sum_{i=1}^{B}
\log\frac{\exp(\mathbf q_i^{\top}\mathbf c_i/\tau)}
{\sum_{j=1}^{B}\exp(\mathbf q_i^{\top}\mathbf c_j/\tau)}.
\label{eq:app-nce}
\end{equation}
The loss is applied only after the final loop.  Intermediate states receive
gradients through the unrolled recurrent graph but are not forced to form
separate embedding spaces.  No language-modeling loss, rationale supervision,
reinforcement-learning objective, or per-loop contrastive head is used.

\begin{center}
\footnotesize
\setlength{\tabcolsep}{3pt}
\renewcommand{\arraystretch}{1.02}
\begin{tabular}{@{}p{0.43\columnwidth}p{0.46\columnwidth}@{}}
\toprule
\textbf{Item} & \textbf{Setting} \\
\midrule
Training data & Official 24 MMEB-V2 training datasets \\
Total optimization steps / seed & 5,000 / 42 \\
Global batch size & 256 query--candidate pairs \\
Registers / recurrent depth & $M=5$ / $T=4$ \\
Default Qwen2-VL recurrent block & Layers 17--26 \\
Video input & 8 frames sampled at 2 FPS \\
Pixel budget & 100,352--602,112 pixels \\
Similarity / temperature & Cosine / $\tau=0.05$ \\
Optimizer & AdamW \\
LM and merger learning rate & $1\times10^{-5}$ \\
Vision-encoder learning rate & $2\times10^{-6}$ \\
Betas / weight decay & $(0.9,0.95)$ / 0.1 \\
Gradient clipping / warm-up & 1.0 / 5\% \\
Schedule & Cosine decay \\
Precision / memory strategy & BF16 / gradient checkpointing + ZeRO-3 \\
Hardware & 8$\times$ NVIDIA H20 \\
PyTorch & 2.5.1+cu121 \\
torchvision / torchaudio & 0.20.1+cu121 / 2.5.1+cu121 \\
Transformers & 5.3.0 \\
\bottomrule
\end{tabular}
\captionof{table}{Complete training, software, and input-processing configuration.}
\label{tab:training-config}
\end{center}

\paragraph{Evaluation aggregation and checkpoint selection.}
Every MMEB-V2 task is evaluated with the benchmark-provided candidate pool and
its official metric.  We first compute one score for each of the 78 tasks and
then macro-average tasks within Image, Video, and VisDoc; the reported All score
is the macro-average over all tasks rather than an average of the three modality
means.  Hyperparameters and the recurrent block are fixed before test evaluation,
and the checkpoint at step 5,000 is used for every backbone and every reported
ablation.

\paragraph{Matched preprocessing.}
All compared recurrent variants reuse the same task instructions, image and
video preprocessing, candidate representations, contrastive temperature, batch
construction, and optimization schedule.  Consequently, changing $T$, $M$, or
the recurrent interval changes only the computation studied by that ablation;
it does not change the candidate pool, task sampling distribution, or evaluation
script.

\onecolumn
\section{Detailed Scores on MMEB-V2}
\label{app:mmeb-detailed}
\noindent
Table~\ref{tab:mmebv2-detailed} reports the complete 78-task results for the four
ReLoop-UME checkpoints evaluated in the main paper.  The columns correspond to
Qwen2-VL-2B, Qwen2-VL-7B, Qwen3-VL-2B, and Qwen3.5-2B; all numbers are read
directly from the corresponding evaluation summaries.  Parenthesized counts
show the number of tasks in each aggregate, and the best result among the four
checkpoints is bolded in every row.

\begin{center}
\captionof{table}{Complete task-level MMEB-V2 results for ReLoop-UME across four backbones. Image and video tasks report Hit@1; visual-document tasks report nDCG@5. Modality and overall scores are task-level macro averages.}
\label{tab:mmebv2-detailed}
\vspace{-3pt}
\fontsize{5.05}{5.33}\selectfont
\setlength{\tabcolsep}{5.0pt}
\renewcommand{\arraystretch}{0.83}
\begin{tabular*}{\textwidth}{@{\extracolsep{\fill}}p{0.52\textwidth}|rrrr@{}}

\toprule
\textbf{Task / aggregate} & \shortstack{\textbf{Qwen2-VL}\\\textbf{2B}} & \shortstack{\textbf{Qwen2-VL}\\\textbf{7B}} & \shortstack{\textbf{Qwen3-VL}\\\textbf{2B}} & \shortstack{\textbf{Qwen3.5}\\\textbf{2B}} \\
\midrule
\rowcolor{gray!16}\textbf{Avg. - All (78 tasks)} & 63.2 & \textbf{65.9} & 63.9 & 62.6 \\
\rowcolor{blue!10}\textbf{Avg. - Image (36 tasks, Hit@1)} & 68.9 & \textbf{72.0} & 68.9 & 68.0 \\
\rowcolor{orange!12}\textbf{Avg. - Video (18 tasks, Hit@1)} & 40.5 & \textbf{42.9} & 42.9 & 40.6 \\
\rowcolor{violet!10}\textbf{Avg. - VisDoc (24 tasks, nDCG@5)} & 71.5 & \textbf{73.9} & 72.3 & 70.9 \\
\rowcolor{blue!6}\textbf{I-CLS (10)} & 67.4 & \textbf{69.5} & 66.6 & 65.0 \\
\rowcolor{blue!6}\textbf{I-QA (10)} & 63.5 & \textbf{67.5} & 64.2 & 64.3 \\
\rowcolor{blue!6}\textbf{I-RET (12)} & 69.8 & \textbf{72.5} & 68.6 & 67.5 \\
\rowcolor{blue!6}\textbf{I-VG (4)} & 83.9 & \textbf{87.8} & 87.3 & 86.4 \\
\rowcolor{orange!7}\textbf{V-CLS (5)} & 44.2 & 44.5 & \textbf{45.3} & 39.7 \\
\rowcolor{orange!7}\textbf{V-QA (5)} & 46.3 & \textbf{52.0} & 49.8 & 48.0 \\
\rowcolor{orange!7}\textbf{V-RET (5)} & 33.6 & 34.3 & \textbf{35.2} & 33.8 \\
\rowcolor{orange!7}\textbf{V-MR (3)} & 36.5 & 39.4 & 39.9 & \textbf{41.1} \\
\rowcolor{violet!6}\textbf{VD-ViDoRe-V1 (10)} & 76.9 & \textbf{78.9} & 77.4 & 77.0 \\
\rowcolor{violet!6}\textbf{VD-ViDoRe-V2 (4)} & 46.5 & \textbf{50.6} & 50.5 & 42.7 \\
\rowcolor{violet!6}\textbf{VD-VisRAG (6)} & 83.5 & \textbf{85.7} & 83.1 & 84.7 \\
\rowcolor{violet!6}\textbf{VD-OOD (4)} & 65.3 & \textbf{67.1} & 64.9 & 63.2 \\
\midrule
VOC2007 & 91.4 & \textbf{93.4} & 93.1 & 93.3 \\
N24News & 80.4 & \textbf{82.5} & 78.2 & 78.6 \\
SUN397 & 81.4 & \textbf{82.4} & 81.6 & 80.1 \\
ObjectNet & \textbf{63.6} & 63.0 & 62.4 & 55.7 \\
Country211 & 24.7 & \textbf{26.0} & 15.8 & 14.0 \\
Place365 & 46.7 & \textbf{48.2} & 45.7 & 45.4 \\
ImageNet-1K & 82.8 & \textbf{83.0} & 80.2 & 79.3 \\
HatefulMemes & 63.8 & \textbf{70.8} & 64.1 & 65.5 \\
ImageNet-A & 49.1 & \textbf{54.5} & 52.9 & 50.6 \\
ImageNet-R & 89.8 & 90.7 & \textbf{91.6} & 87.5 \\
\addlinespace[0.35pt]
MSCOCO i2t & 72.0 & \textbf{74.6} & 73.6 & 73.0 \\
MSCOCO t2i & 72.6 & 77.4 & \textbf{77.5} & 75.8 \\
VisDial & 75.6 & 81.0 & \textbf{81.2} & 79.6 \\
CIRR & 58.5 & 60.7 & \textbf{61.7} & 60.3 \\
VisualNews i2t & 79.0 & \textbf{82.7} & 73.6 & 71.1 \\
VisualNews t2i & 75.3 & \textbf{80.1} & 71.2 & 68.5 \\
NIGHTS & 68.7 & 67.5 & 69.0 & \textbf{69.1} \\
WebQA & 89.7 & \textbf{91.5} & 88.9 & 90.2 \\
EDIS & \textbf{88.7} & 81.3 & 77.3 & 73.1 \\
OVEN & 67.7 & \textbf{72.7} & 59.3 & 57.2 \\
Wiki-SS-NQ & 67.3 & \textbf{72.6} & 66.4 & 66.3 \\
FashionIQ & 22.1 & \textbf{27.9} & 23.0 & 25.6 \\
\addlinespace[0.35pt]
OK-VQA & 67.3 & \textbf{73.1} & 67.3 & 65.6 \\
A-OKVQA & 57.7 & \textbf{62.1} & 56.7 & 56.4 \\
DocVQA & 92.5 & \textbf{93.9} & 92.6 & 92.6 \\
InfographicsVQA & 64.4 & \textbf{72.6} & 68.4 & 70.4 \\
ChartQA & 57.0 & 63.1 & 61.8 & \textbf{64.1} \\
Visual7W & 53.6 & \textbf{55.7} & 54.3 & 54.0 \\
ScienceQA & 44.1 & \textbf{52.4} & 45.2 & 46.2 \\
GQA & \textbf{70.3} & 69.0 & 63.5 & 65.7 \\
TextVQA & 79.1 & 81.4 & \textbf{83.8} & 81.2 \\
VizWiz & 49.3 & \textbf{51.9} & 48.6 & 47.1 \\
\addlinespace[0.35pt]
MSCOCO grounding & 78.0 & 83.2 & \textbf{84.4} & 84.2 \\
RefCOCO & 89.8 & 92.0 & \textbf{93.1} & 92.5 \\
RefCOCO-Matching & 88.1 & \textbf{90.4} & 87.4 & 85.4 \\
Visual7W-Pointing & 79.5 & \textbf{85.7} & 84.4 & 83.5 \\
\specialrule{0.55pt}{1.0pt}{1.0pt}
UCF101 & \textbf{63.8} & 60.4 & 57.7 & 51.5 \\
HMDB51 & \textbf{50.3} & 49.9 & 49.0 & 44.9 \\
K700 & 47.2 & \textbf{51.6} & 46.6 & 44.2 \\
Breakfast & 15.0 & 18.0 & \textbf{27.7} & 17.6 \\
SmthSmthV2 & 44.8 & 42.5 & \textbf{45.6} & 40.4 \\
\addlinespace[0.35pt]
Video-MME & 40.4 & \textbf{46.3} & 46.0 & 40.6 \\
MVBench & 43.0 & \textbf{50.8} & 46.3 & 43.4 \\
NExTQA & 51.2 & \textbf{57.3} & 56.2 & 51.1 \\
EgoSchema & 48.2 & \textbf{49.6} & 48.0 & 47.6 \\
ActivityNetQA & 48.5 & 56.2 & 52.7 & \textbf{57.4} \\
\addlinespace[0.35pt]
MSR-VTT & 36.7 & \textbf{39.2} & \textbf{39.2} & 37.6 \\
MSVD & 55.8 & 56.1 & \textbf{57.5} & 56.3 \\
DiDeMo & 34.9 & 34.2 & \textbf{35.8} & 34.1 \\
VATEX & 27.2 & 28.4 & \textbf{29.9} & 29.0 \\
YouCook2 & 13.2 & 13.6 & \textbf{13.8} & 12.2 \\
\addlinespace[0.35pt]
QVHighlight & 48.0 & 52.8 & 52.6 & \textbf{57.2} \\
Charades-STA & 17.5 & 19.8 & 20.2 & \textbf{20.9} \\
MomentSeeker & 44.1 & 45.6 & \textbf{46.8} & 45.2 \\
\specialrule{0.55pt}{1.0pt}{1.0pt}
ViDoRe-arxivqa & 84.4 & \textbf{87.9} & 85.3 & 86.5 \\
ViDoRe-docvqa & 44.0 & \textbf{46.0} & 45.1 & 44.9 \\
ViDoRe-infovqa & 86.1 & \textbf{88.7} & 85.9 & 87.1 \\
ViDoRe-shiftproject & \textbf{67.9} & 67.5 & 67.6 & 61.8 \\
ViDoRe-artificial intelligence & 86.0 & 87.1 & \textbf{87.4} & 86.9 \\
ViDoRe-energy & 83.2 & 85.0 & 84.6 & \textbf{87.0} \\
ViDoRe-government reports & 84.6 & 87.1 & \textbf{87.1} & 84.7 \\
ViDoRe-healthcare industry & 90.3 & \textbf{93.8} & 89.1 & 89.8 \\
ViDoRe-tabfquad & 92.4 & \textbf{93.1} & 91.7 & 91.4 \\
ViDoRe-tatdqa & 49.8 & \textbf{53.1} & 50.2 & 49.8 \\
\addlinespace[0.35pt]
ViDoRe-biomedical lectures-v2 multilingual & 46.8 & \textbf{51.4} & 51.2 & 49.3 \\
ViDoRe-economics reports-v2 multilingual & 47.0 & 49.6 & \textbf{51.6} & 49.5 \\
ViDoRe-esg reports human labeled-v2 & 50.5 & \textbf{58.1} & 50.8 & 42.7 \\
ViDoRe-esg reports-v2 multilingual & 41.5 & 43.2 & \textbf{48.5} & 29.3 \\
\addlinespace[0.35pt]
VisRAG ArxivQA & 80.8 & \textbf{84.0} & 81.7 & 82.4 \\
VisRAG ChartQA & \textbf{87.8} & 87.2 & 84.8 & 86.2 \\
VisRAG InfoVQA & 87.7 & \textbf{89.9} & 87.5 & 89.2 \\
VisRAG MP-DocVQA & 80.7 & \textbf{86.5} & 80.9 & 83.3 \\
VisRAG PlotQA & 72.6 & 74.1 & 72.1 & \textbf{75.3} \\
VisRAG SlideVQA & 91.0 & \textbf{92.6} & 91.3 & 92.1 \\
\addlinespace[0.35pt]
ViDoSeek-document & 80.4 & \textbf{82.4} & 80.8 & 80.0 \\
ViDoSeek-page & 80.5 & \textbf{82.3} & 79.1 & 76.5 \\
\addlinespace[0.35pt]
MMLongBench-document & 48.9 & \textbf{51.7} & 50.0 & 47.8 \\
MMLongBench-page & 51.2 & \textbf{52.1} & 49.8 & 48.5 \\
\bottomrule
\end{tabular*}
\end{center}

\section{Extended Layer-Wise Retrieval Analysis}
\label{app:layerwise}
Figures~\ref{fig:qwen23-quantiles} and~\ref{fig:qwen35-quantiles} include all
twelve supplied layer-wise curves.  Each colored trajectory comes from an
independently trained retrieval model, while the thick black trajectory is the
cross-run mean.  The four negative quantiles alter the hardness and vertical
scale of the positive--negative margin, but not the physical interval in which
retrieval separation is formed.

\begin{center}
\setlength{\tabcolsep}{1.4pt}
\renewcommand{\arraystretch}{0.84}
\begin{tabular}{@{}cc@{}}
\includegraphics[width=.492\textwidth]{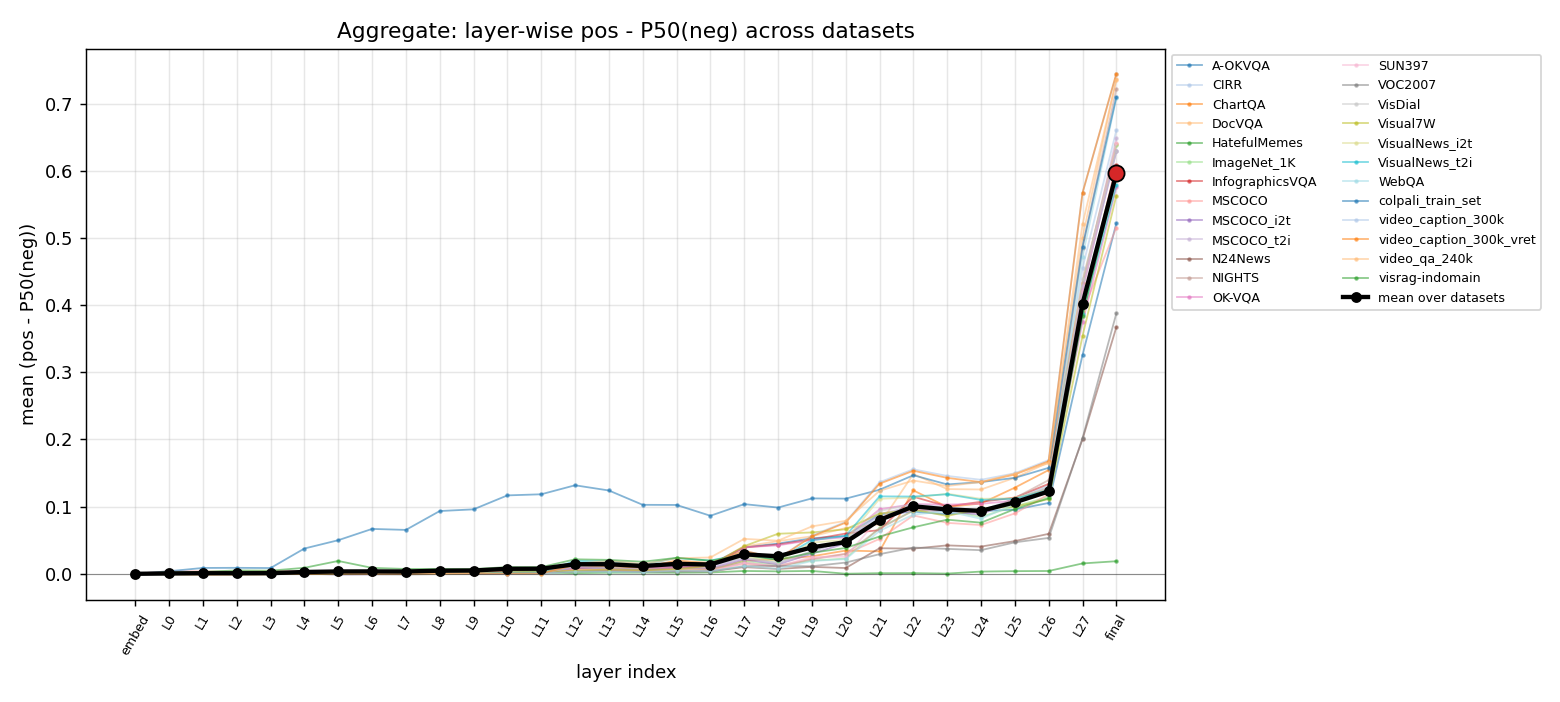} &
\includegraphics[width=.492\textwidth]{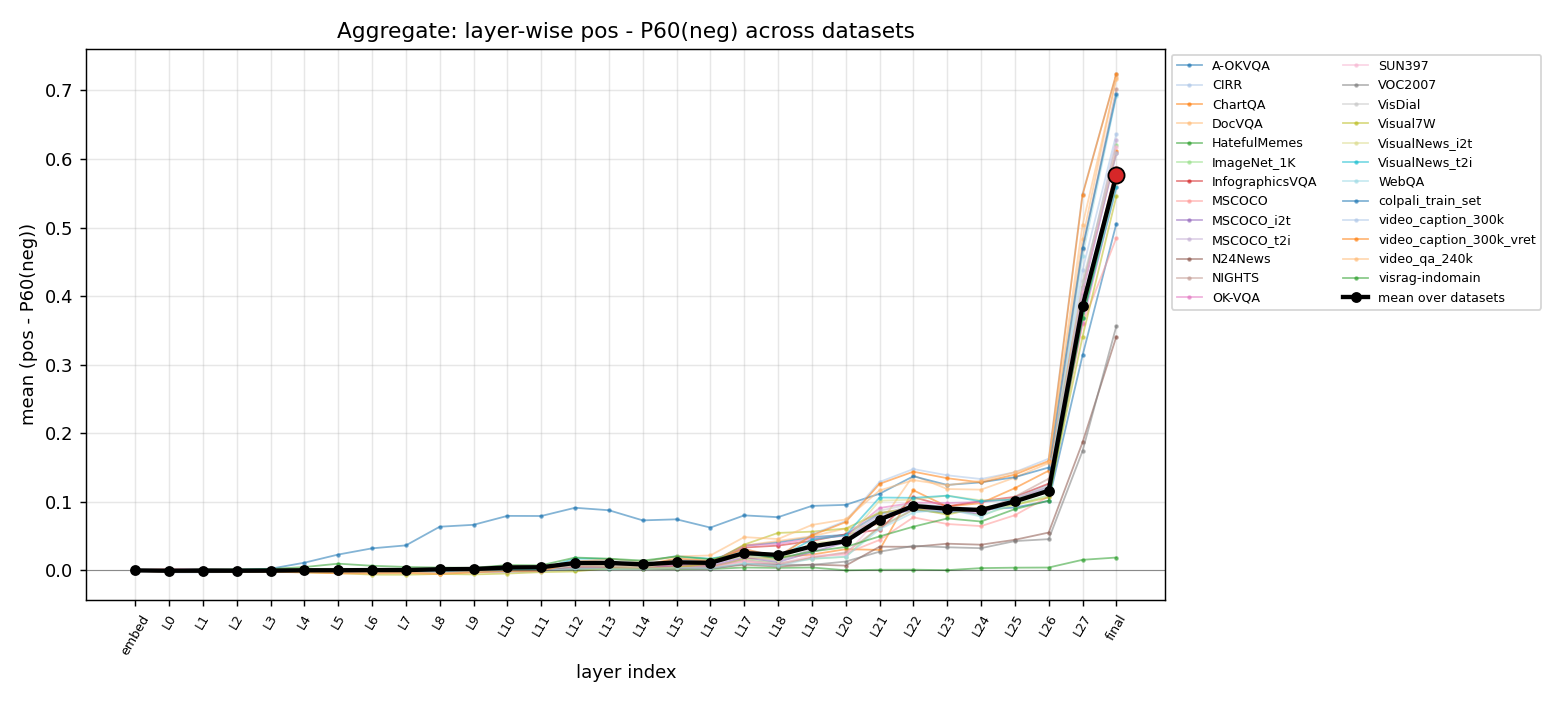} \\
[-6pt]\scriptsize (a) Qwen2-VL, $q=0.5$ & \scriptsize (b) Qwen2-VL, $q=0.6$ \\
\includegraphics[width=.492\textwidth]{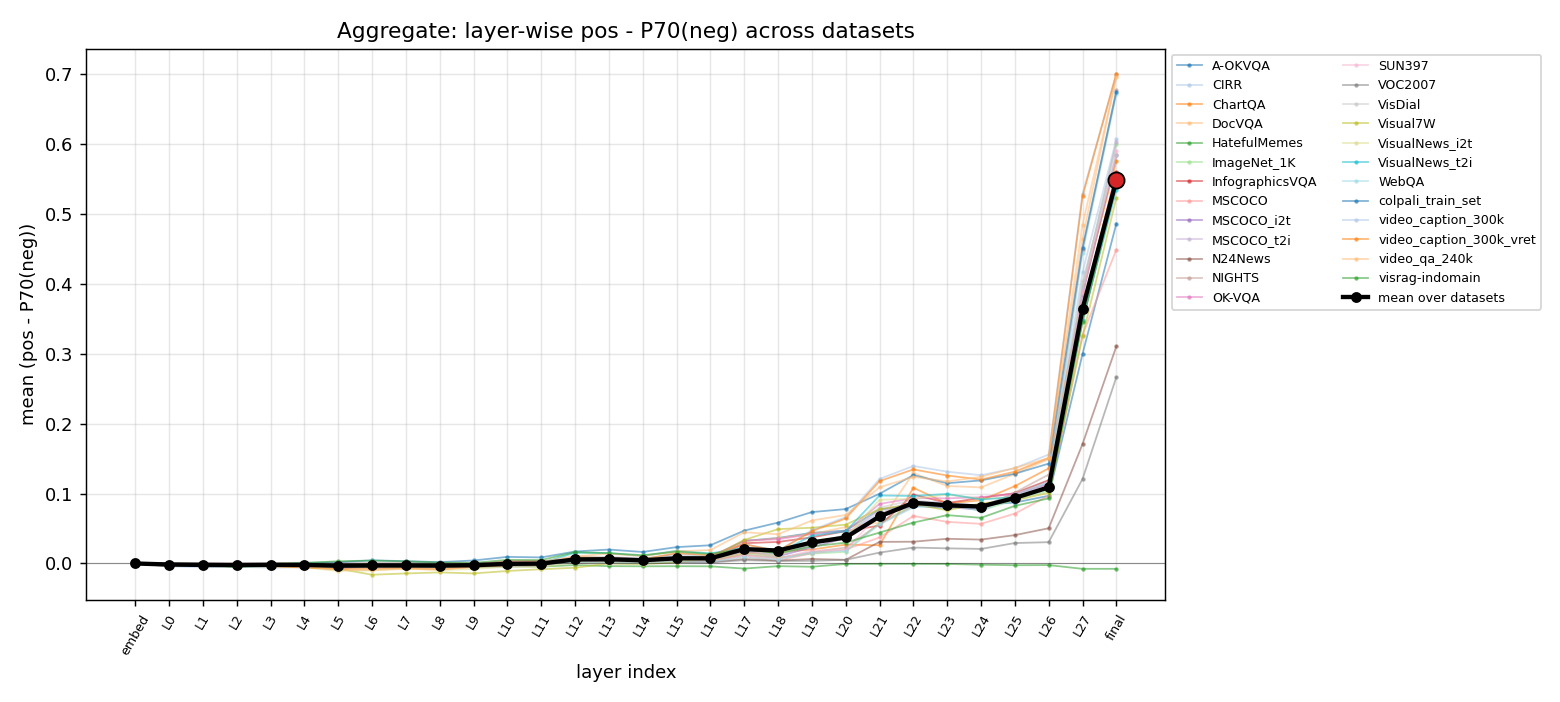} &
\includegraphics[width=.492\textwidth]{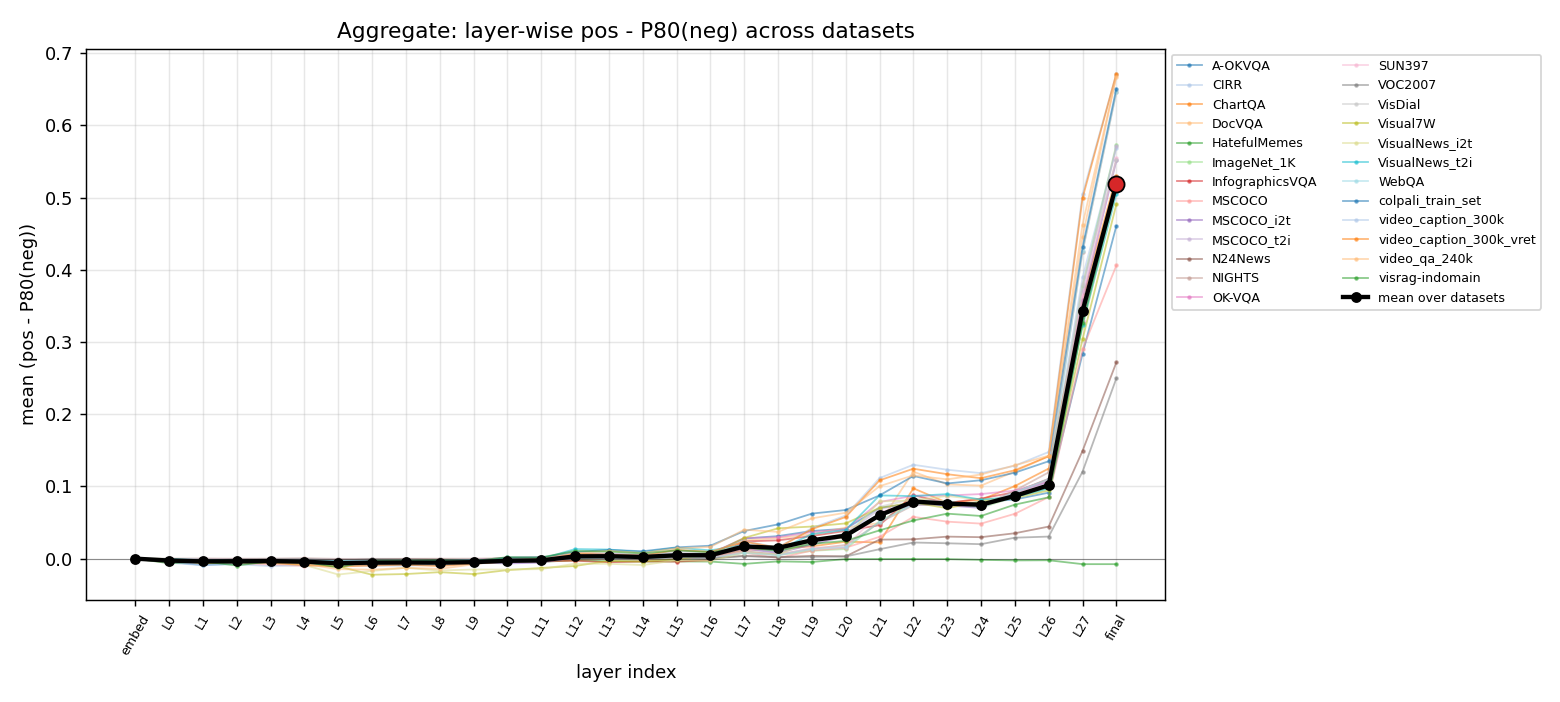} \\
[-6pt]\scriptsize (c) Qwen2-VL, $q=0.7$ & \scriptsize (d) Qwen2-VL, $q=0.8$ \\
\includegraphics[width=.492\textwidth]{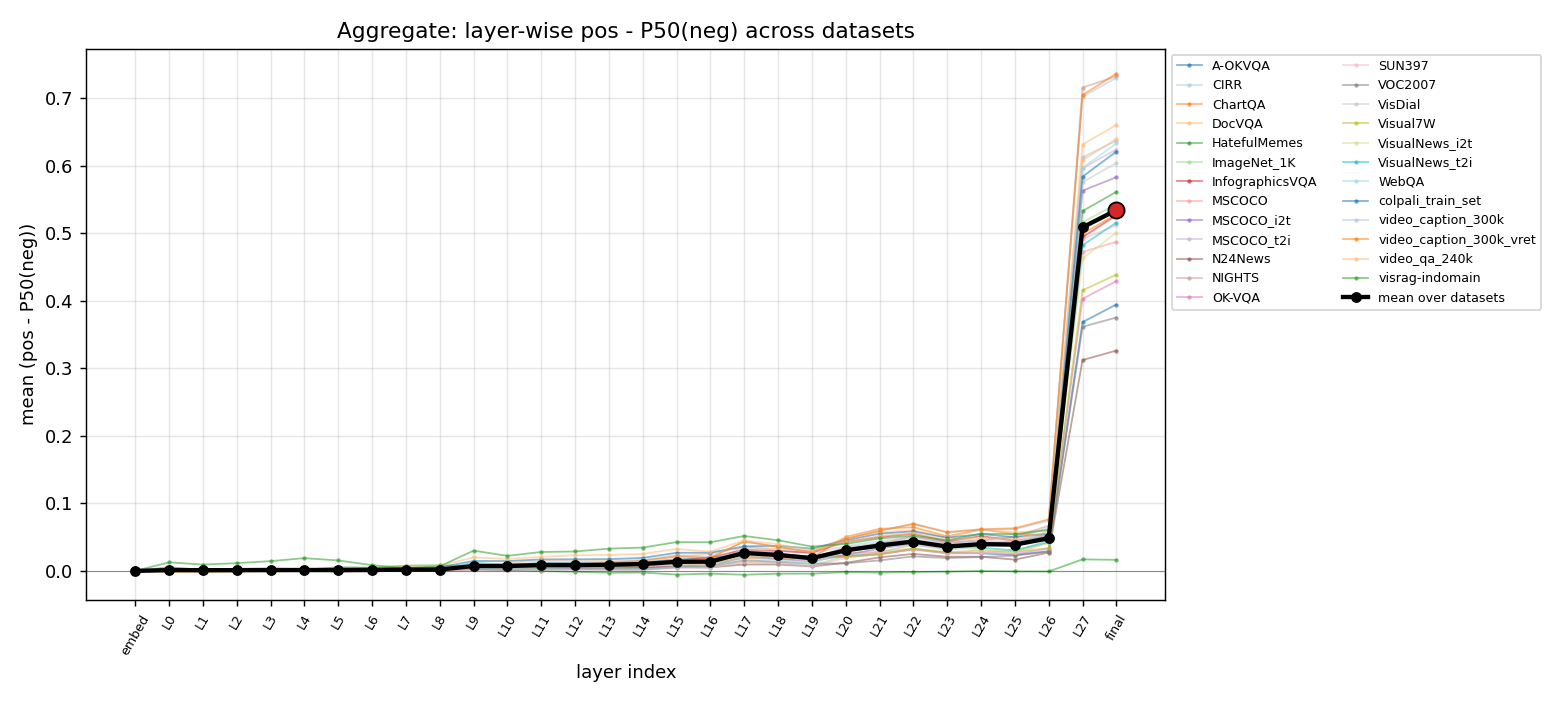} &
\includegraphics[width=.492\textwidth]{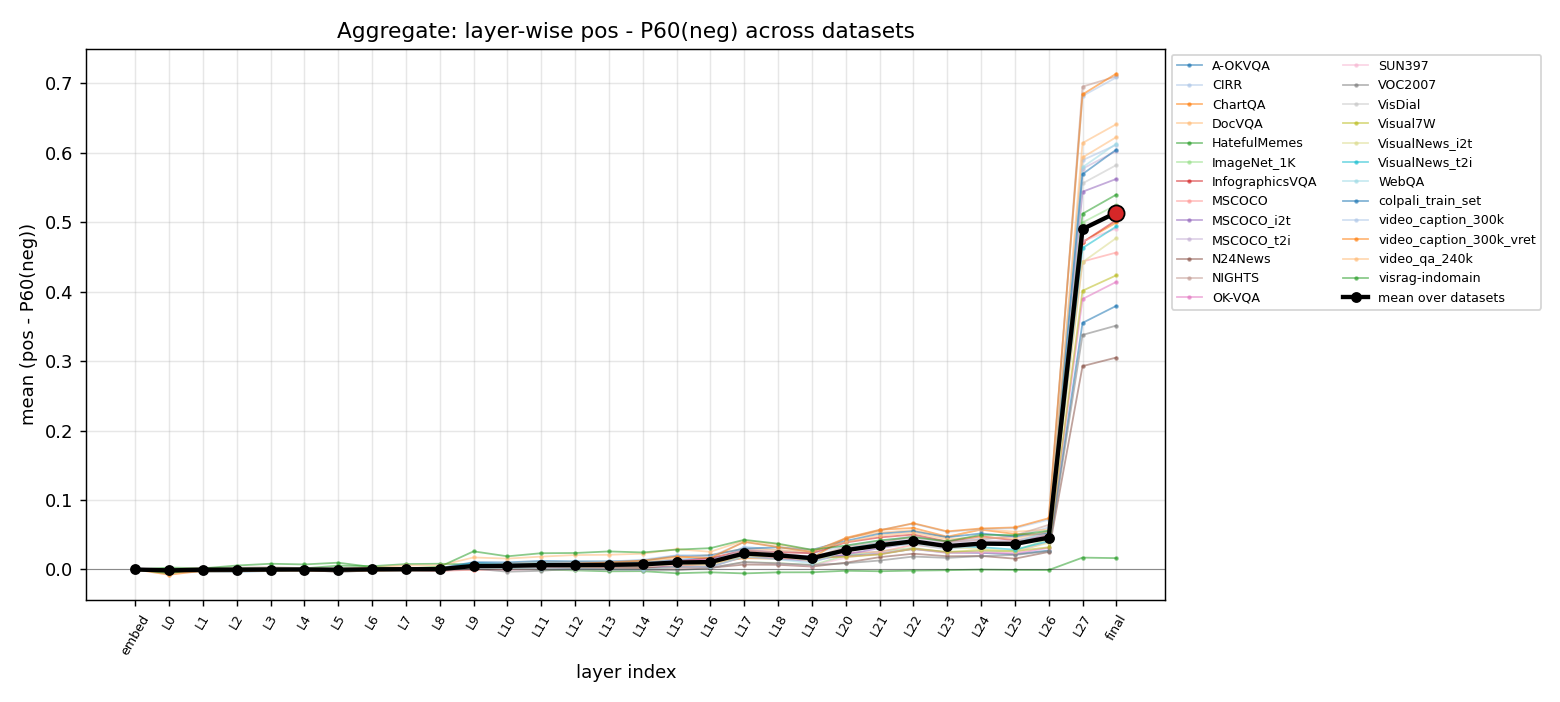} \\
[-6pt]\scriptsize (e) Qwen3-VL, $q=0.5$ & \scriptsize (f) Qwen3-VL, $q=0.6$ \\
\includegraphics[width=.492\textwidth]{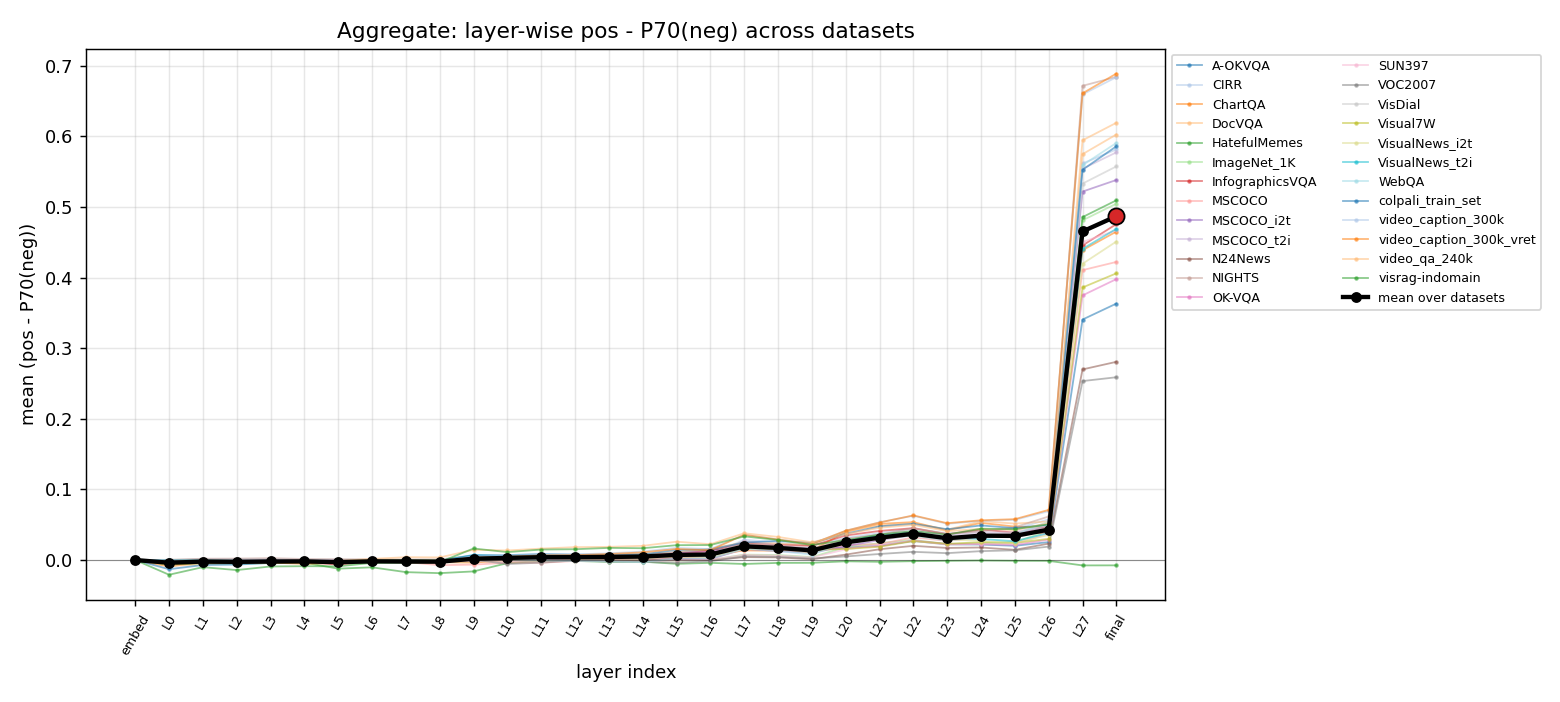} &
\includegraphics[width=.492\textwidth]{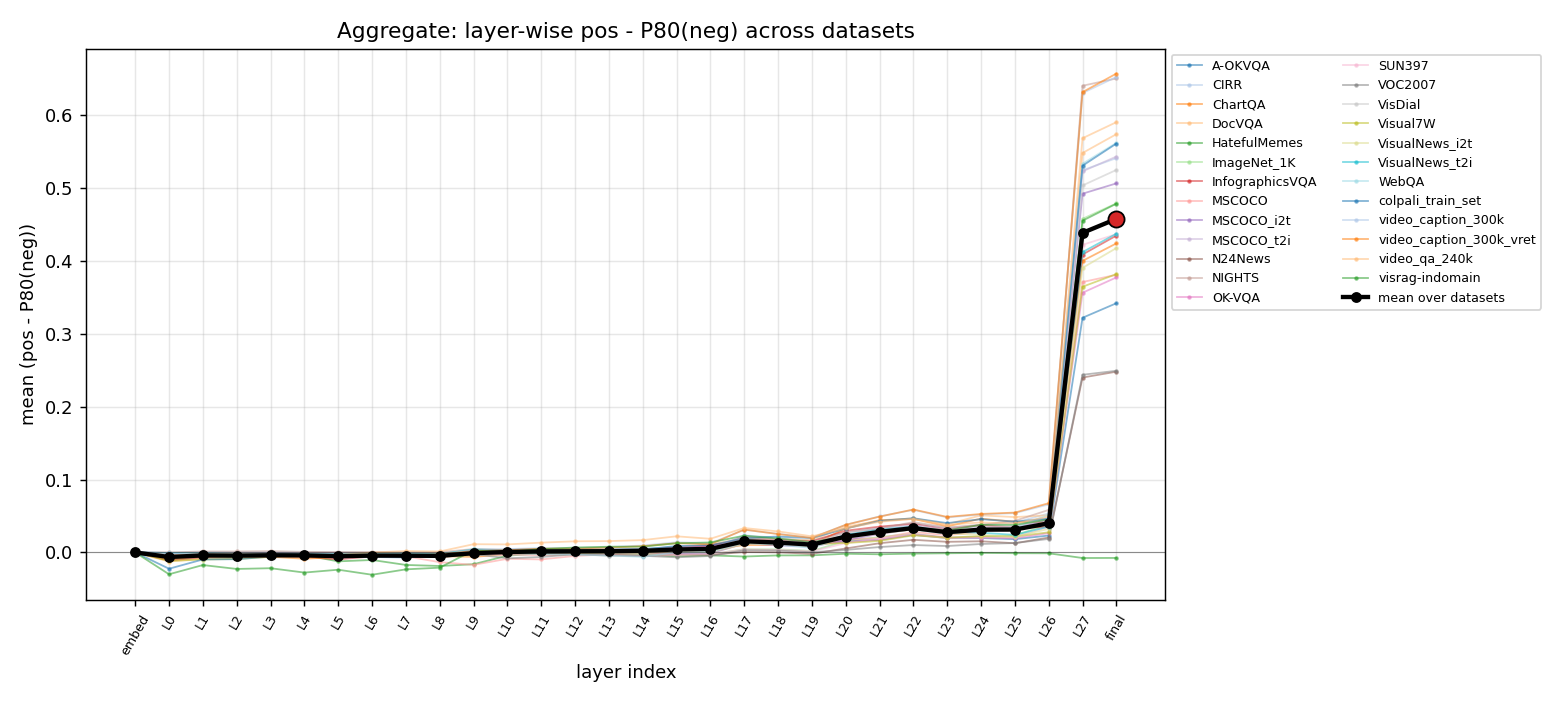} \\
[-6pt]\scriptsize (g) Qwen3-VL, $q=0.7$ & \scriptsize (h) Qwen3-VL, $q=0.8$ \\
\end{tabular}
\captionof{figure}{Layer-wise positive--negative separation for Qwen2-VL-2B and Qwen3-VL-2B.  Both 28-layer backbones exhibit a low-slope prefix, a sustained retrieval-formation rise over layers 17--26, and a sharp terminal mapping transition at layer 27.  Every tested quantile selects the same recurrent block.}
\label{fig:qwen23-quantiles}
\end{center}
\vspace{-5pt}

\paragraph{Deterministic pooled change-point fitting.}
Stage boundaries are selected before recurrent training and are never tuned on
final retrieval accuracy.  For independently trained run $r$ and negative
quantile $q$, let $\widetilde S_{r,\ell}^{(q)}$ be the layer-wise separation curve
normalized to $[0,1]$.  For every ordered pair $(a,b)$ satisfying a minimum
four-layer prefix, a minimum four-layer middle interval, and at least one
terminal layer, we fit separate least-squares lines to the three contiguous
segments.  The selected boundaries minimize
\[
J(a,b)=\sum_{r,q}\sum_{s=1}^{3}\sum_{\ell\in I_s(a,b)}
\left(\widetilde S_{r,\ell}^{(q)}-\alpha_{rqs}\ell-\beta_{rqs}\right)^2,
\]
subject to the middle slope being positive and larger than the prefix and
terminal slopes.  Ties are resolved by the smallest worst-run residual and then
the shorter middle interval.  Thus, a single favorable run or visually unusual
curve cannot determine the block.

\newpage
\begin{center}
\setlength{\tabcolsep}{1.8pt}
\renewcommand{\arraystretch}{0.86}
\begin{tabular}{@{}cc@{}}
\includegraphics[width=.486\textwidth]{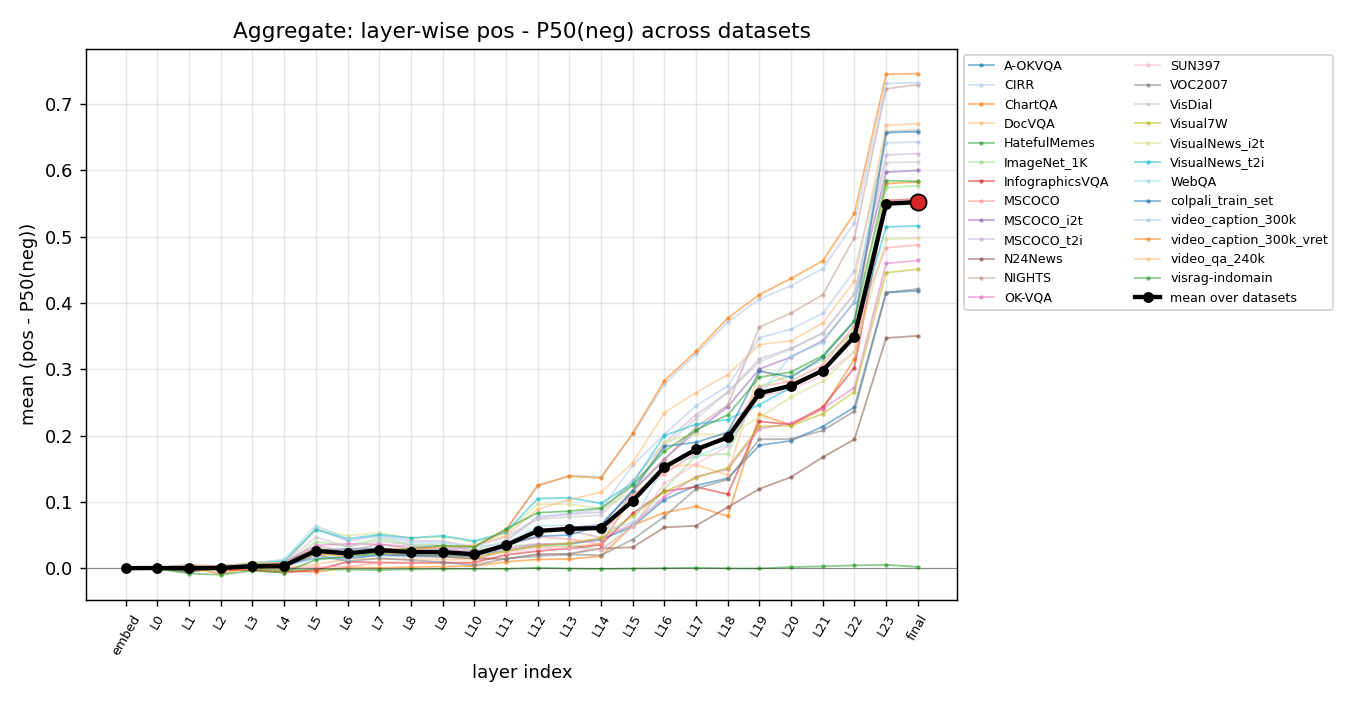} &
\includegraphics[width=.486\textwidth]{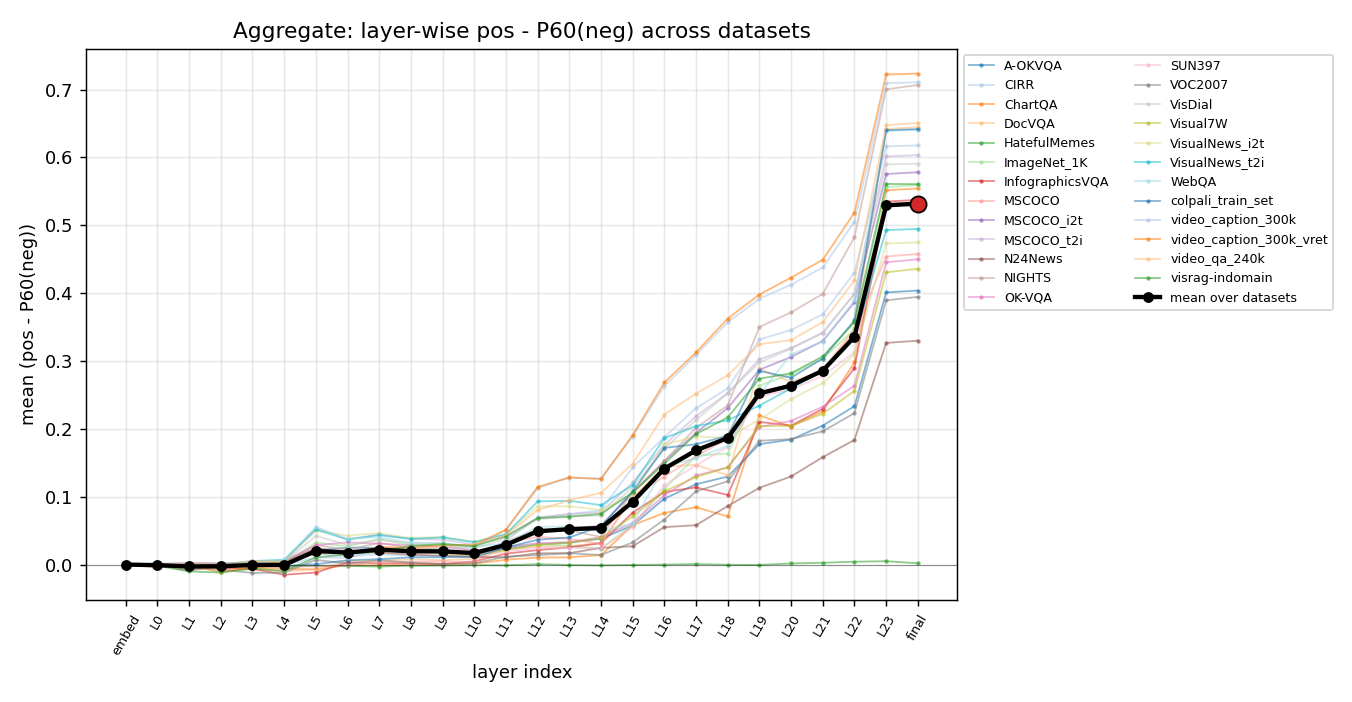} \\
[-5pt]\scriptsize (a) Qwen3.5, $q=0.5$ & \scriptsize (b) Qwen3.5, $q=0.6$ \\
\includegraphics[width=.486\textwidth]{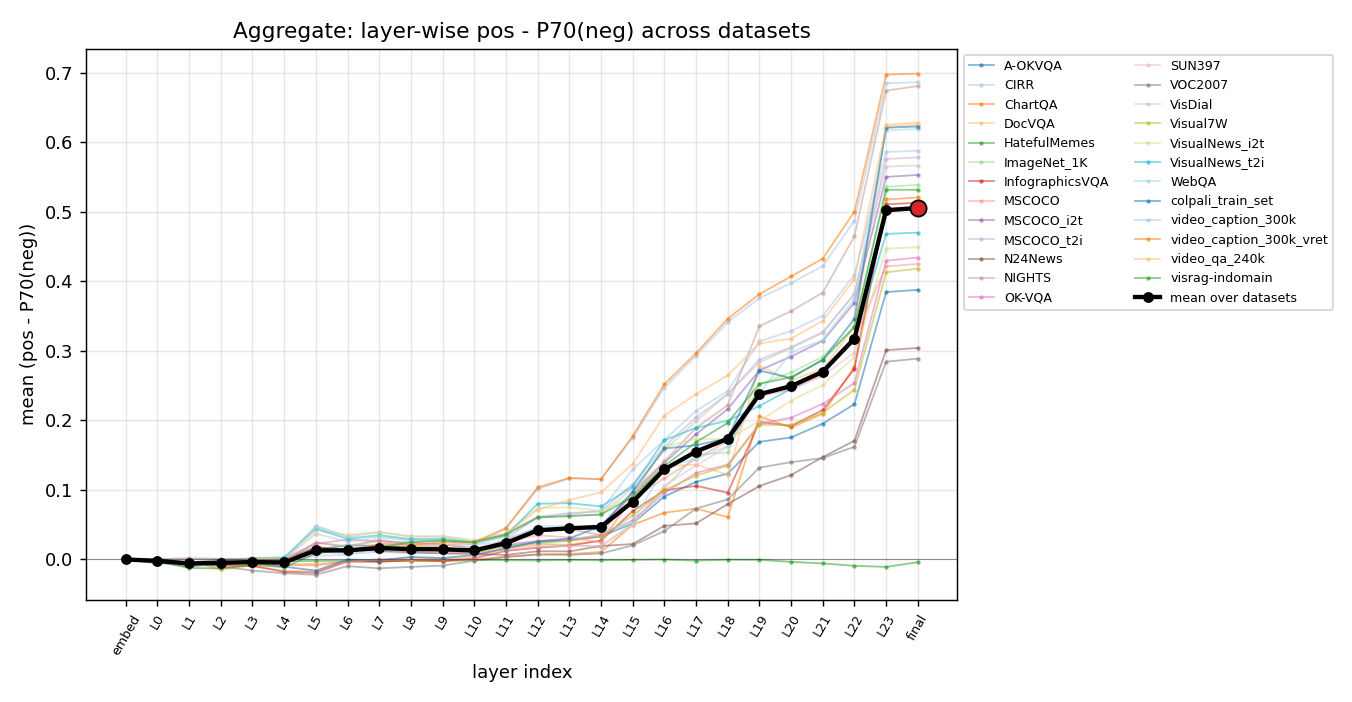} &
\includegraphics[width=.486\textwidth]{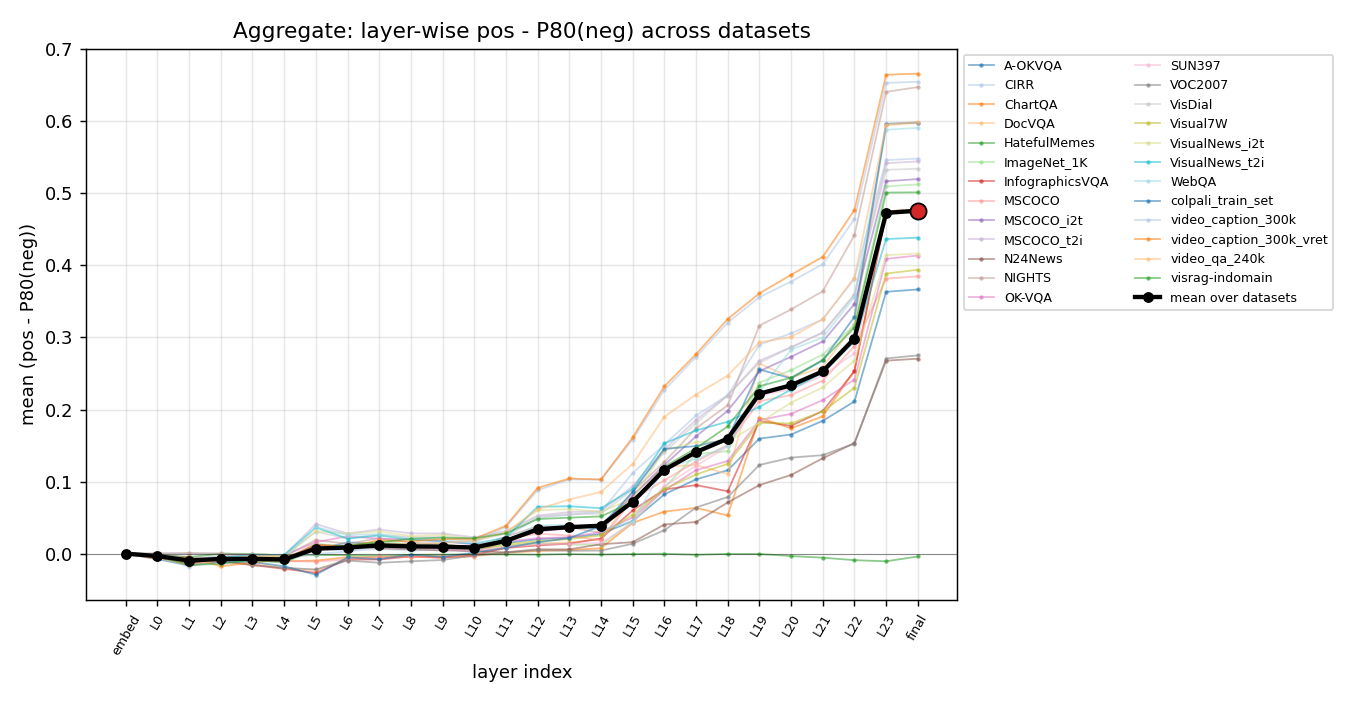} \\
[-5pt]\scriptsize (c) Qwen3.5, $q=0.7$ & \scriptsize (d) Qwen3.5, $q=0.8$ \\
\end{tabular}
\captionof{figure}{Layer-wise separation for the 24-layer Qwen3.5-2B hybrid stack.  All four quantiles select prefix layers 0--11, retrieval formation over layers 12--22, and terminal mapping at layer 23.}
\label{fig:qwen35-quantiles}
\end{center}
\vspace{-4pt}

\paragraph{Quantile and run stability.}
For Qwen2-VL and Qwen3-VL, every quantile and independent run places the onset
within layers 16--17 and the saturation transition within 26--27, so the fixed
tie rule always returns 17--26.  Lower quantiles compress the vertical gap by
using easier negatives, while $q=0.8$ makes the hard-negative transition most
legible.  It is therefore a visualization choice rather than a sensitive
boundary hyperparameter.  Qwen3.5 shows the same invariance at 12--22.

\paragraph{Depth-aware rather than fraction-based localization.}
Qwen3.5 contains 24 decoder layers and mixes linear-attention with full-attention
blocks.  Its sustained rise begins earlier in absolute index, at layer 12, while
the terminal mapping remains concentrated in the final layer.  Applying exactly
the same constrained objective returns 12--22 for every quantile.  A fixed layer
index or a fixed depth fraction would not simultaneously recover 17--26 for the
28-layer models and 12--22 for Qwen3.5; the observed retrieval dynamics do.

\begin{center}
\small
\setlength{\tabcolsep}{10pt}
\renewcommand{\arraystretch}{1.10}
\begin{tabular}{@{}lcccc@{}}
\toprule
\textbf{Backbone} & \textbf{Depth} & \textbf{Prefix} & \textbf{Formation} & \textbf{Mapping} \\
\midrule
Qwen2-VL-2B & 28 & 0--16 & 17--26 & 27 \\
Qwen3-VL-2B & 28 & 0--16 & 17--26 & 27 \\
Qwen3.5-2B & 24 & 0--11 & 12--22 & 23 \\
\bottomrule
\end{tabular}
\hspace{14pt}
\begin{tabular}{@{}lcccc@{}}
\toprule
\textbf{Backbone} & $q=.5$ & $q=.6$ & $q=.7$ & $q=.8$ \\
\midrule
Qwen2-VL-2B & 17--26 & 17--26 & 17--26 & 17--26 \\
Qwen3-VL-2B & 17--26 & 17--26 & 17--26 & 17--26 \\
Qwen3.5-2B & 12--22 & 12--22 & 12--22 & 12--22 \\
\bottomrule
\end{tabular}
\captionof{table}{Backbone-specific stage partitions and quantile sensitivity.}
\label{tab:all-boundaries}
\end{center}

\paragraph{Why the terminal mapping layer is not looped.}
The final-layer jump is consistent with mapping an already discriminative hidden
state into the output embedding space.  Repeating it would reapply a readout
transformation rather than extend the interval over which evidence is gradually
separated.  ReLoop-UME therefore repeats only the rising middle block and applies
the terminal layer once after the final recurrent pass.  This interpretation is
also consistent with the range ablation: repeating 17--26 outperforms looping the
full stack, the suffix, the prefix, or only the final layer despite using fewer
decoder applications than full-stack recurrence.

\paragraph{Cross-backbone conclusion.}
The three architectures differ in depth, attention composition, and visual
injection, yet all show the same functional progression: prefix understanding,
retrieval formation, and terminal embedding mapping.  The exact physical index
is backbone-specific, but it is stable across independently trained runs and
negative quantiles.  The matched results in Table~\ref{tab:backbone-results}
close the empirical loop: recurrence over the independently localized interval
improves all three backbones, while Qwen3-VL and Qwen3.5 show slightly smaller
gains because their single-pass representations and internal visual pathways
differ from the homogeneous Qwen2-VL stack.

\newpage
% Portability guard: small font-metric differences across TeX Live releases can
% otherwise strand the final two lines on a nearly empty page before \twocolumn.
\enlargethispage{3\baselineskip}
\section{Complete MRMR Evaluation}
\label{app:mrmr}

The 7B ReLoop-UME checkpoint is trained only on MMEB-V2 and evaluated on MRMR
without benchmark-specific optimization.  Table~\ref{tab:mrmr-full} reproduces
the complete official MRMR baseline table and appends the ReLoop-UME result from
the main paper.  No LaME, UME-R1, or RIME values are imported into this table.
The ReLoop-UME row is therefore directly traceable to the reported 11-subtask
result, while every baseline row follows the official MRMR evaluation setup.

\begin{center}
\captionof{table}{Complete MRMR evaluation. Official baseline values are reproduced from MRMR Table 3. All subtasks use nDCG@10 except Negation, which uses Hit@1. ReLoop-UME is transferred zero-shot from MMEB-V2; its 11 values and 48.64 macro average match the main paper.}
\label{tab:mrmr-full}
\vspace{-3pt}
\fontsize{7.25}{7.85}\selectfont
\setlength{\tabcolsep}{3.7pt}
\renewcommand{\arraystretch}{1.02}
\begin{tabular*}{\textwidth}{@{\extracolsep{\fill}}lrrrrrrrrrrrr@{}}
\toprule
& \multicolumn{4}{c}{\textbf{Knowledge}} & \multicolumn{4}{c}{\textbf{Theorem}} & \multicolumn{3}{c}{\textbf{Contradiction}} & \\
\cmidrule(lr){2-5}\cmidrule(lr){6-9}\cmidrule(lr){10-12}
\textbf{Model} & \textbf{Art} & \textbf{Med.} & \textbf{Sci.} & \textbf{Hum.} & \textbf{Math} & \textbf{Phy.} & \textbf{Eng.} & \textbf{Bus.} & \textbf{Neg.} & \textbf{Des.} & \textbf{Tra.} & \textbf{Avg.} \\
\midrule
\rowcolor{gray!12}\multicolumn{13}{c}{\textit{Text Models with Image Caption (T2T)}} \\
BGE-M3 & 48.6 & 30.0 & 42.4 & 45.6 & 13.5 & 15.7 & 18.3 & 26.6 & 16.0 & 25.9 & 17.4 & 27.3 \\
NV-Embed-v2 & 70.7 & 46.8 & 65.7 & 66.6 & 26.2 & 27.3 & 29.0 & 36.9 & 12.5 & 42.1 & 42.2 & 42.4 \\
Qwen3-Embedding & 71.9 & 53.2 & 72.5 & 74.4 & 35.9 & 48.1 & 39.6 & 43.7 & 12.0 & 67.8 & 54.2 & 52.1 \\
\rowcolor{gray!12}\multicolumn{13}{c}{\textit{Text--Image Two-Stream with Vector Fusion (IT2IT)}} \\
EVA-CLIP & 10.2 & 13.5 & 26.1 & 12.9 & 6.2 & 10.5 & 9.3 & 11.7 & 8.5 & 4.4 & 5.4 & 10.8 \\
SigLIP & 26.7 & 14.7 & 26.7 & 12.3 & 6.2 & 5.5 & 4.1 & 7.5 & 13.5 & 4.9 & 9.6 & 12.0 \\
OpenCLIP & 56.0 & 17.9 & 33.2 & 22.0 & 5.7 & 5.0 & 7.0 & 9.7 & 13.0 & 8.1 & 12.4 & 17.3 \\
JinaCLIP & 21.4 & 16.8 & 27.1 & 10.7 & 8.3 & 5.9 & 8.4 & 10.4 & 10.5 & 16.5 & 9.7 & 13.2 \\
\rowcolor{gray!12}\multicolumn{13}{c}{\textit{Multimodal Models with Merged Image (IT2IT)}} \\
VISTA & 21.3 & 27.8 & 32.6 & 17.0 & 14.2 & 14.3 & 19.5 & 14.2 & 20.0 & 20.2 & 9.4 & 19.1 \\
E5-V & 25.1 & 11.7 & 16.6 & 10.8 & 1.1 & 1.5 & 4.1 & 2.0 & 11.5 & 3.7 & 2.1 & 8.2 \\
MM-Embed & 65.6 & 53.0 & 63.5 & 62.8 & 21.6 & 26.3 & 24.4 & 31.7 & 7.0 & 23.8 & 34.9 & 37.7 \\
VLM2Vec & 53.5 & 22.4 & 36.7 & 24.0 & 1.1 & 1.3 & 2.4 & 2.5 & 11.5 & 5.6 & 18.3 & 16.3 \\
GME-Qwen2-VL & 54.3 & 40.1 & 46.8 & 45.6 & 3.0 & 3.6 & 9.3 & 4.6 & 15.0 & 26.3 & 29.6 & 25.3 \\
Ops-MM-Embedding & 79.3 & 52.5 & 70.0 & 67.8 & 23.7 & 34.2 & 27.0 & 35.3 & 8.0 & 55.9 & 45.8 & 45.4 \\
\rowcolor{gray!12}\multicolumn{13}{c}{\textit{Multimodal Models with Document as Image (T2I)}} \\
GME-Qwen2-VL & 54.0 & 40.7 & 59.0 & 50.1 & 15.7 & 22.7 & 20.5 & 32.5 & 14.5 & 56.1 & 40.1 & 36.9 \\
Ops-MM-Embedding & 67.7 & 48.8 & 67.7 & 63.9 & 24.4 & 29.3 & 25.7 & 33.7 & 10.5 & 59.8 & 46.3 & 43.4 \\
ColPali & 36.1 & 29.9 & 42.7 & 29.2 & 5.7 & 14.8 & 12.0 & 24.6 & 28.5 & 19.4 & 18.2 & 23.7 \\
\midrule
\rowcolor{gray!12}\multicolumn{13}{c}{\textit{Native Interleaved UME, Zero-Shot from MMEB-V2}} \\
\rowcolor{orange!12}\textbf{ReLoop-UME (7B)} & \textbf{78.12} & \textbf{59.71} & \textbf{75.17} & \textbf{73.57} & \textbf{27.35} & \textbf{44.97} & \textbf{34.67} & \textbf{49.48} & \textbf{9.50} & \textbf{43.29} & \textbf{39.19} & \textbf{48.64} \\
\bottomrule
\end{tabular*}
\end{center}

\paragraph{Evaluation-snapshot alignment.}
Table~\ref{tab:mrmr-full} reproduces the original official benchmark table without
mixing cells from later third-party papers.  The compact native-UME table in the
main paper follows the evaluation snapshot used by those matched UME experiments;
some public baselines were subsequently re-evaluated under updated model or input
pipelines, so their compact group aggregates need not be obtained by averaging the
original Table~3 cells shown here.  The ReLoop-UME values are identical in both
places.  Keeping the snapshots explicit prevents a per-task table assembled from
incompatible evaluation versions.

ReLoop-UME is strongest on Knowledge, reaching 71.64 on the four expert-domain
subtasks, and remains competitive on Theorem at 39.12.  Its largest weakness is
Negation (9.50), so the 30.66 Contradiction average should not be interpreted as
solving polarity-sensitive retrieval.  The result instead shows that recurrent
feature refinement transfers well to expert semantic matching and part of the
abstract theorem-matching problem without MRMR-specific training.

\begin{center}
\small
\setlength{\tabcolsep}{10pt}
\renewcommand{\arraystretch}{1.08}
\begin{tabular}{@{}lcccc@{}}
\toprule
\textbf{Group} & \textbf{Tasks} & \textbf{Average} & \textbf{Best subtask} & \textbf{Weakest subtask} \\
\midrule
Knowledge & 4 & 71.64 & Art (78.12) & Medicine (59.71) \\
Theorem & 4 & 39.12 & Business (49.48) & Math (27.35) \\
Contradiction & 3 & 30.66 & Design (43.29) & Negation (9.50) \\
\midrule
All & 11 & 48.64 & Art (78.12) & Negation (9.50) \\
\bottomrule
\end{tabular}
\captionof{table}{MRMR group-level summary for ReLoop-UME-7B.}
\label{tab:mrmr-groups}
\end{center}

The gap between Knowledge and the other groups clarifies what transfers from
MMEB-V2.  Recurrent refinement is effective when the query and candidate expose
rich semantic evidence that can be consolidated into one vector.  It is less
reliable when success depends on formal symbolic structure or on identifying one
small polarity reversal.  These failures are consistent with the method's design:
ReLoop-UME adds retrieval-forming depth but does not introduce a theorem solver,
a dedicated negation objective, or benchmark-specific supervision.

\paragraph{Why Qwen3-Embedding is not the primary controlled comparison.}
The official Qwen3-Embedding row uses a two-model system: an external MLLM first
converts images into captions, after which a specialized text embedding model
performs retrieval~\cite{supp_zhang2026mrmr}.  Its score therefore combines captioner
quality, textualization, and text retrieval, whereas ReLoop-UME directly encodes
the original interleaved multimodal sequence.  In addition, the official
Qwen3-Embedding training recipe uses large-scale weak supervision, high-quality
supervised data, synthetic task expansion, and model merging
\cite{supp_zhang2025qwen3embedding,supp_qwen2025embeddingblog}.  This is a valuable system-level reference, but it is not an
architecture-matched test of recurrent depth; the controlled comparison therefore
uses native UME systems with comparable backbone scale and training data.

\twocolumn
\section{Backbone Architectures and Cross-Backbone Results}
\label{app:backbones}

\subsection{Released Architecture Configurations}

Qwen2-VL and Qwen3-VL use 28 text-decoder layers, whereas Qwen3.5 uses 24.
Qwen3.5 is a hybrid stack with 18 linear-attention layers and six full-attention
layers.  Qwen3-VL additionally injects visual features into selected decoder
depths through DeepStack.  These differences motivate backbone-specific stage
localization rather than assuming that a fixed fraction of depth always serves
the same function~\cite{supp_qwen2vl2bconfig,supp_qwen3vl2bconfig,supp_qwen35_2bconfig}.

\begin{center}
\footnotesize
\setlength{\tabcolsep}{2.7pt}
\renewcommand{\arraystretch}{1.02}
\begin{tabular}{@{}lrrrr@{}}
\toprule
\textbf{Backbone} & \textbf{Text layers} & \textbf{Width} & \textbf{Heads/KV} & \textbf{Context} \\
\midrule
Qwen2-VL-2B & 28 & 1536 & 12/2 & 32K \\
Qwen3-VL-2B & 28 & 2048 & 16/8 & 256K \\
Qwen3.5-2B & 24 & 2048 & 8/2 & 256K \\
\bottomrule
\end{tabular}
\captionof{table}{Text-stack configurations of the three 2B backbones.}
\label{tab:backbone-text}
\end{center}

\begin{center}
\footnotesize
\setlength{\tabcolsep}{2.1pt}
\renewcommand{\arraystretch}{1.02}
\begin{tabular}{@{}p{0.23\columnwidth}ccp{0.40\columnwidth}@{}}
\toprule
\textbf{Backbone} & \textbf{Vision layers} & \textbf{Patch} & \textbf{Relevant structure} \\
\midrule
Qwen2-VL-2B & 32 & 14 & Homogeneous decoder with M-RoPE and no DeepStack index list. \\
Qwen3-VL-2B & 24 & 16 & Interleaved M-RoPE; visual injection at decoder indices 5, 11, and 17. \\
Qwen3.5-2B & 24 & 16 & 18 linear-attention and six full-attention layers; full attention every fourth layer. \\
\bottomrule
\end{tabular}
\captionof{table}{Vision and fusion structures relevant to recurrent reuse.}
\label{tab:backbone-vision}
\end{center}

\subsection{Matched Retrieval Results}

\begin{center}
\small
\setlength{\tabcolsep}{4pt}
\renewcommand{\arraystretch}{1.01}
\begin{tabular}{@{}lrrr@{}}
\toprule
\textbf{Backbone} & \textbf{Single forward} & \textbf{ReLoop-UME} & $\boldsymbol{\Delta}$ \\
\midrule
Qwen2-VL-2B & 60.6 & 63.2 & +2.6 \\
Qwen3-VL-2B & 61.7 & \textbf{63.9} & +2.2 \\
Qwen3.5-2B & 60.2 & 62.6 & +2.4 \\
\bottomrule
\end{tabular}
\captionof{table}{Backbone generalization on MMEB-V2 All using the same training recipe.}
\label{tab:backbone-results}
\end{center}

Qwen3-VL has the strongest single-forward and final result, while Qwen2-VL shows
the largest absolute gain.  The smaller Qwen3-VL gain is consistent with less
headroom and with stage-specific DeepStack injection: later recurrent
applications refine the fused sequence but do not introduce new visual features
at every loop.  Qwen3.5 has the lowest single-forward result under the shared
hyperparameters.  Its 24-layer hybrid stack provides fewer physical layers and
only six full-attention layers, so a recipe initially developed around a
homogeneous 28-layer decoder is not expected to be equally optimal.  Nevertheless,
its +2.4 gain shows that the localization-and-looping principle survives the
architectural change.  These are mechanism-aligned interpretations, not claims
that layer count alone causally determines accuracy.

\section{Extended Ablations}
\label{app:ablations}

\subsection{Where Recurrence Is Applied}

\begin{center}
\footnotesize
\setlength{\tabcolsep}{2.4pt}
\renewcommand{\arraystretch}{1.00}
\begin{tabular}{@{}lrrrrr@{}}
\toprule
\textbf{Layers} & \textbf{Apps.} & \textbf{Image} & \textbf{Video} & \textbf{VisDoc} & \textbf{All} \\
\midrule
0--16 & 79 & 66.0 & 39.4 & 67.0 & 60.2 \\
0--27 & 112 & 67.1 & 39.8 & 68.8 & 61.3 \\
17--27 & 61 & 68.1 & 40.2 & 70.2 & 62.3 \\
27 & 31 & 65.2 & 39.0 & 66.2 & 59.5 \\
\rowcolor{gray!14}17--26 & 58 & \textbf{68.9} & \textbf{40.5} & \textbf{71.5} & \textbf{63.2} \\
\bottomrule
\end{tabular}
\captionof{table}{Ablation on the recurrent interval with $M=5$ and $T=4$.}
\label{tab:range-ablation}
\end{center}

The complete-stack loop performs 112 decoder applications, nearly twice the 58
applications of layers 17--26, yet remains 1.9 points worse.  Repeating only the
final mapping layer is also insufficient.  The gain is therefore explained by
where additional depth is allocated, not by arbitrary extra computation.

\subsection{Register Count and Recurrent Depth}

\begin{center}
\footnotesize
\setlength{\tabcolsep}{2.3pt}
\renewcommand{\arraystretch}{1.00}
\begin{tabular}{@{}lcrrrr@{}}
\toprule
\textbf{Setting} & \textbf{Value} & \textbf{Image} & \textbf{Video} & \textbf{VisDoc} & \textbf{All} \\
\midrule
\multicolumn{6}{c}{\textit{Retrieval registers}} \\
$M$ & 0 & 67.5 & 39.9 & 70.4 & 61.8 \\
$M$ & 5 & \textbf{68.9} & 40.5 & \textbf{71.5} & \textbf{63.2} \\
$M$ & 8 & 68.5 & 40.6 & 71.2 & 62.9 \\
$M$ & 10 & 68.0 & \textbf{40.8} & 70.8 & 62.7 \\
\midrule
\multicolumn{6}{c}{\textit{Recurrent depth}} \\
$T$ & 1 & 66.5 & 39.7 & 67.5 & 60.6 \\
$T$ & 2 & 68.0 & 40.2 & 69.8 & 62.1 \\
$T$ & 4 & \textbf{68.9} & \textbf{40.5} & \textbf{71.5} & \textbf{63.2} \\
$T$ & 8 & 68.7 & 40.4 & 70.8 & 62.8 \\
\bottomrule
\end{tabular}
\captionof{table}{Register-count and recurrent-depth ablations.  Five registers add only 7,680 trainable parameters.}
\label{tab:register-depth}
\end{center}

\subsection{Complete $T\in\{1,4\}\times M\in\{0,5\}$ Factorial Test}

Table~\ref{tab:factorial} completes the $2\times2$ design on MMEB-V2 All:
$T=1,M=0$ is single-forward encoding, $T=1,M=5$ adds registers only,
$T=4,M=0$ adds recurrence only, and $T=4,M=5$ is the full model.

\begin{center}
\small
\setlength{\tabcolsep}{5.0pt}
\renewcommand{\arraystretch}{1.05}
\begin{tabular}{@{}lrrr@{}}
\toprule
& $\boldsymbol{T=1}$ & $\boldsymbol{T=4}$ & \textbf{Recurrence gain} \\
\midrule
$M=0$ & 60.6 & 61.8 & +1.2 \\
$M=5$ & 60.6 & \textbf{63.2} & +2.6 \\
\midrule
\textbf{Register gain} & +0.0 & +1.4 & \textbf{Interaction: +1.4} \\
\bottomrule
\end{tabular}
\captionof{table}{Complete $2\times2$ factorial experiment on MMEB-V2 All.  ``Interaction'' is the difference-in-differences $(63.2-61.8)-(60.6-60.6)$.}
\label{tab:factorial}
\end{center}

% This command is intentionally placed in the right column.  Minor font-metric
% differences can otherwise push the final two lines past the following \onecolumn.
\enlargethispage{3\baselineskip}
Averaged over $M$, recurrence contributes +1.9 points; averaged over $T$,
registers contribute +0.7. Registers leave a single forward pass unchanged, but
their gain increases to +1.4 when the block is reused, yielding a positive +1.4
interaction. This closes the method story: recurrence supplies additional
retrieval-forming computation, while the persistent workspace becomes more
valuable when evidence can be accumulated across repeated applications.

\clearpage
\onecolumn
\section{Qualitative Comparisons and Failure Cases}
\label{app:qualitative-failures}

This section complements the aggregate tables with eleven diagnostic examples.
The first eight cases compare the characteristic errors made by Explicit CoT and
PLUME with the final retrieval selected by ReLoop-UME.  The final three cases are
failures of ReLoop-UME itself and instantiate the Video and VisDoc limitations
reported in the main paper.  These examples are intended to expose mechanisms
that are difficult to see from a single average score; they are not substitutes
for the controlled quantitative comparisons.

\subsection{Explicit CoT: Salient Narratives Can Replace Retrieval Evidence}

\begin{center}
\setlength{\tabcolsep}{7pt}
\begin{tabular}{@{}cc@{}}
\includegraphics[height=.535\textheight,keepaspectratio]{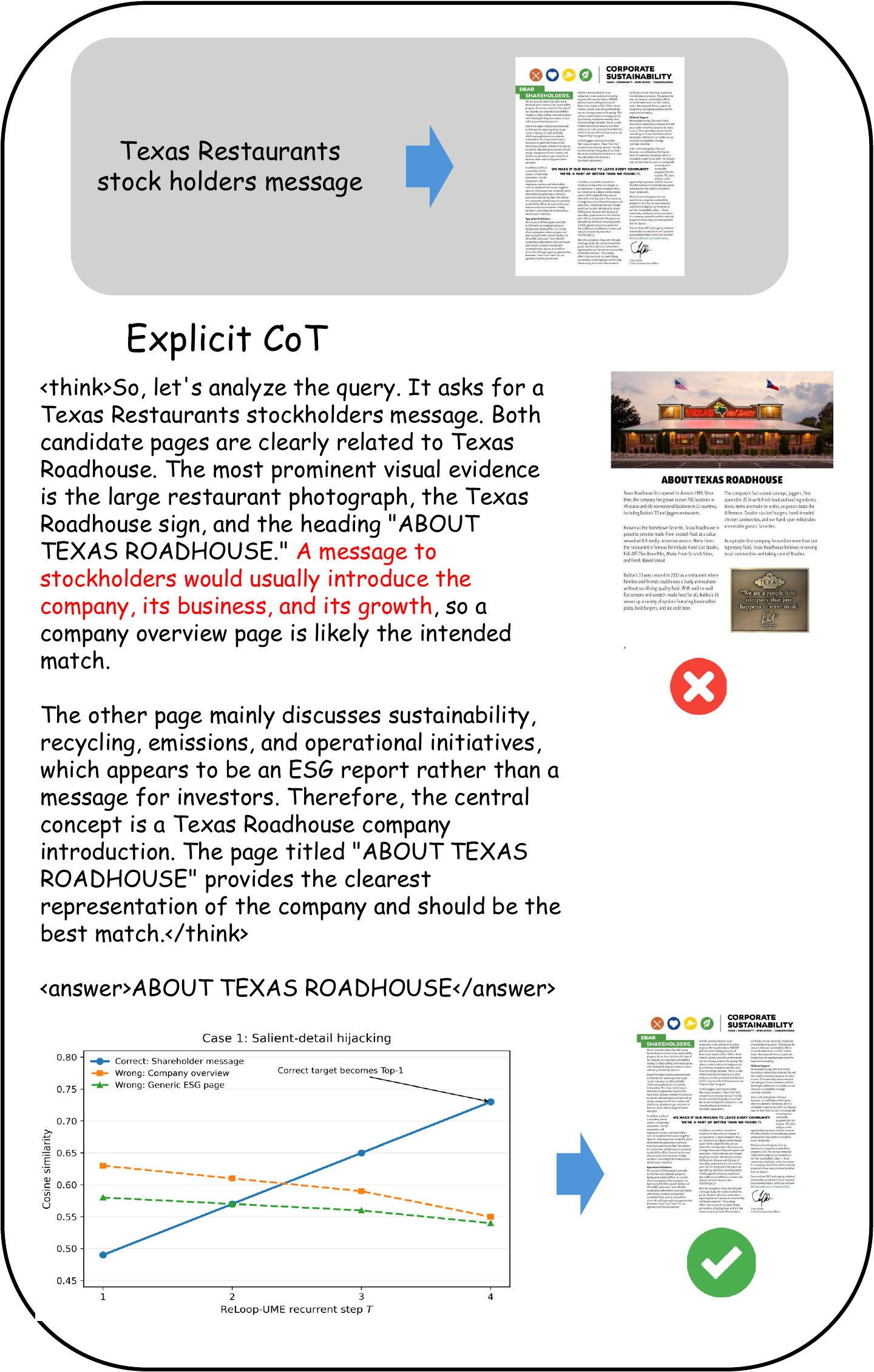} &
\includegraphics[height=.535\textheight,keepaspectratio]{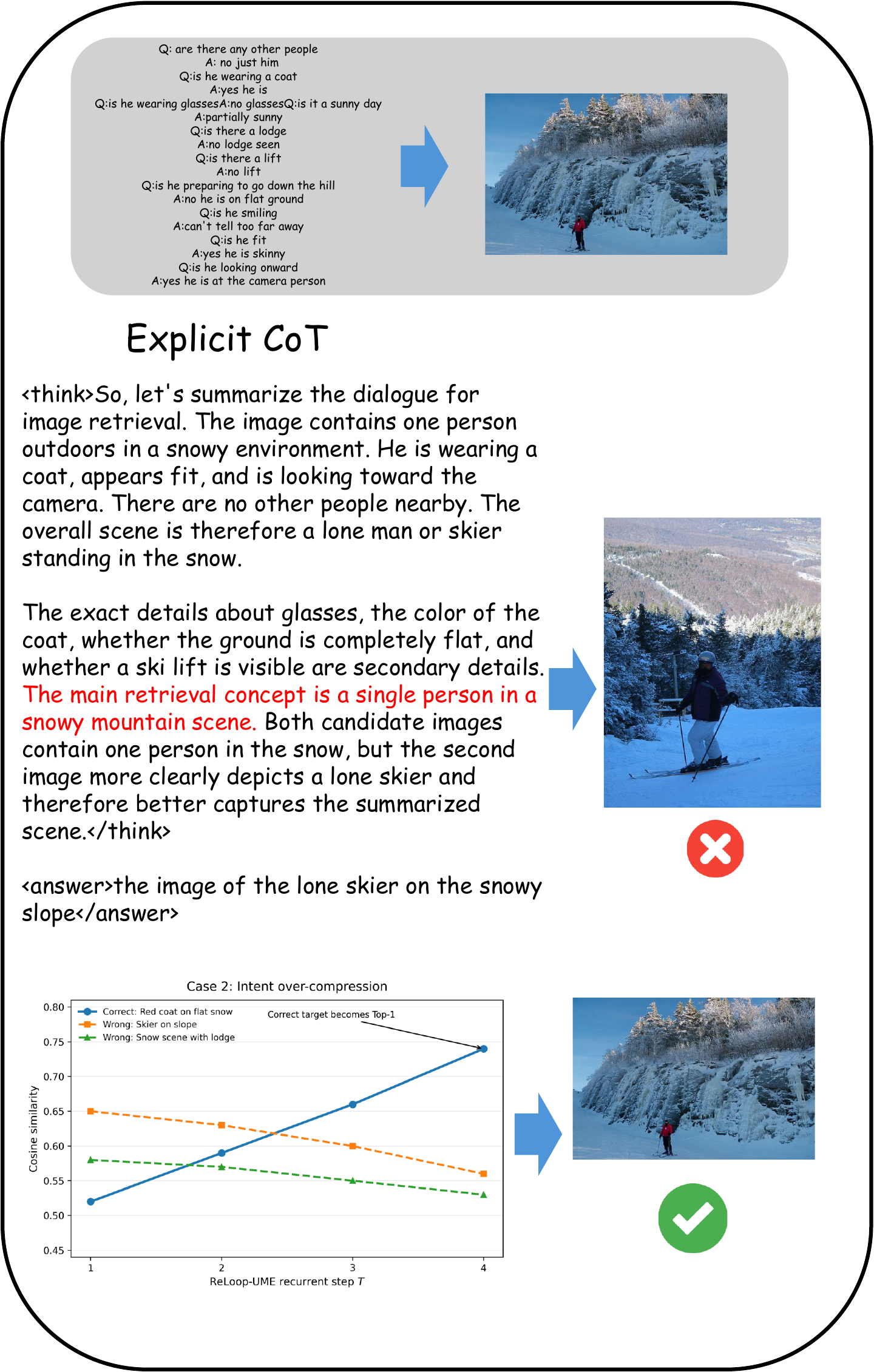} \\
\scriptsize (a) Texas Roadhouse shareholder page &
\scriptsize (b) VisDial snowy-person dialogue
\end{tabular}
\captionof{figure}{Explicit-CoT failures caused by \emph{irrelevant-information hijacking} and \emph{query-intent over-compression}.  In (a), the generated rationale follows visually salient company-overview cues and retrieves an ``About'' page rather than the shareholder message.  In (b), a long dialogue containing multiple positive and negative constraints is reduced to the generic concept ``a lone skier in snow,'' removing the details that separate the two candidates.}
\label{fig:explicit-cot-12}
\end{center}
\vspace{-3pt}

\paragraph{Case 1: irrelevant-information hijacking.}
The Texas Roadhouse query is relational: the target is not merely a page about
Texas Roadhouse, but a shareholder message.  Explicit CoT constructs a fluent
company-introduction narrative from the restaurant photograph, logo, and
``About Texas Roadhouse'' heading.  Once this salient but overly broad hypothesis
is verbalized, the subsequent embedding is dominated by it and the
shareholder-specific relation is lost.  ReLoop-UME instead keeps the original
visual and textual tokens inside the retrieval encoder, allowing the recurrent
block and registers to preserve the narrow query relation while suppressing
visually prominent but non-discriminative content.

\paragraph{Case 2: query-intent over-compression.}
The VisDial query contains a sequence of constraints: one person, a coat, no
visible glasses, no lodge, no lift, flat ground, and a camera-facing pose.
Explicit CoT compresses them into ``a lone skier in a snowy scene.''  That
summary is semantically plausible but insufficient for ranking two highly
similar candidates.  The error illustrates a retrieval-specific bottleneck:
a natural-language summary can be shorter and more readable while being less
faithful to the attributes that determine nearest-neighbor ranking.

\newpage
\begin{center}
\setlength{\tabcolsep}{7pt}
\begin{tabular}{@{}cc@{}}
\includegraphics[height=.535\textheight,keepaspectratio]{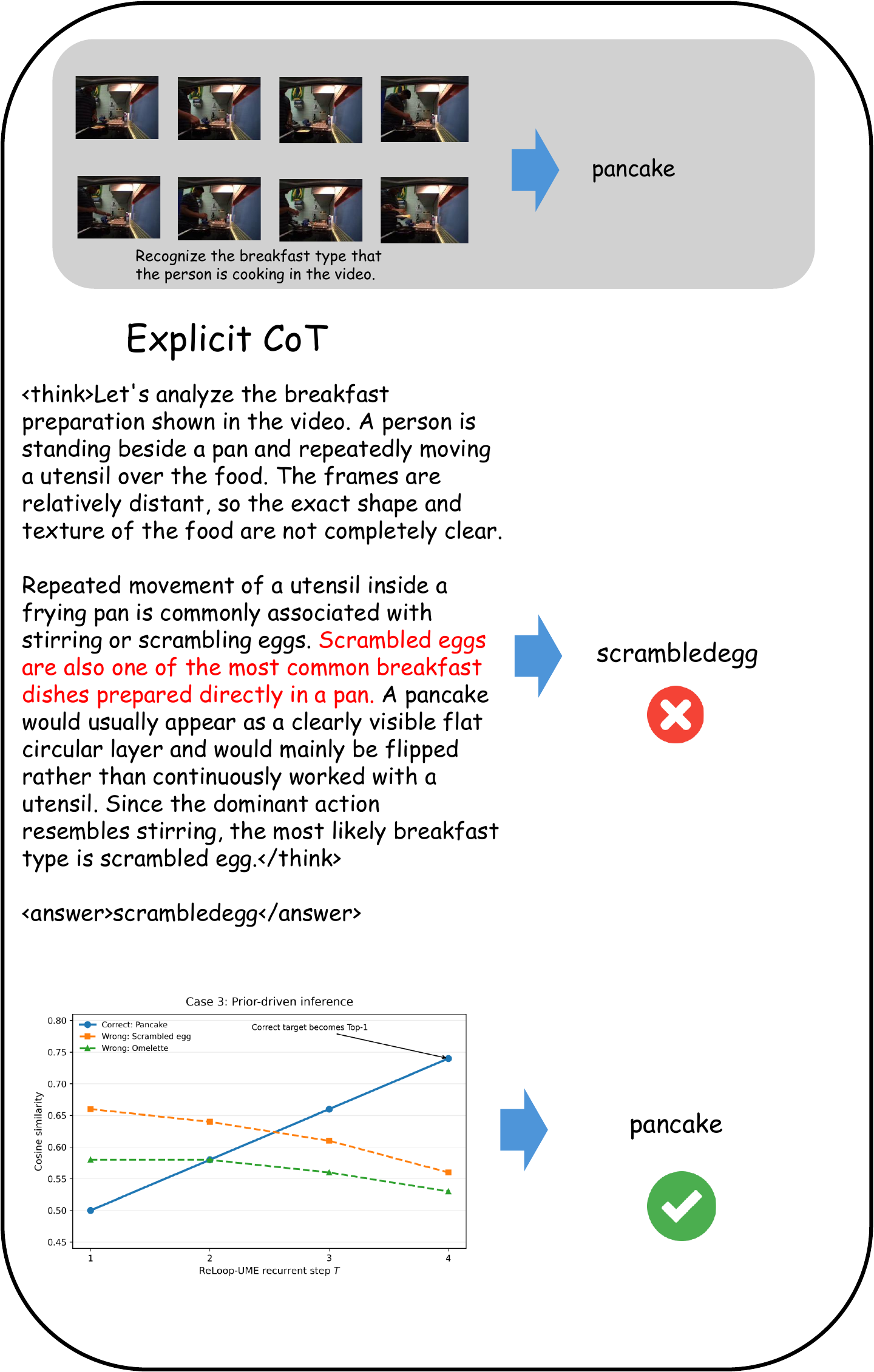} &
\includegraphics[height=.535\textheight,keepaspectratio]{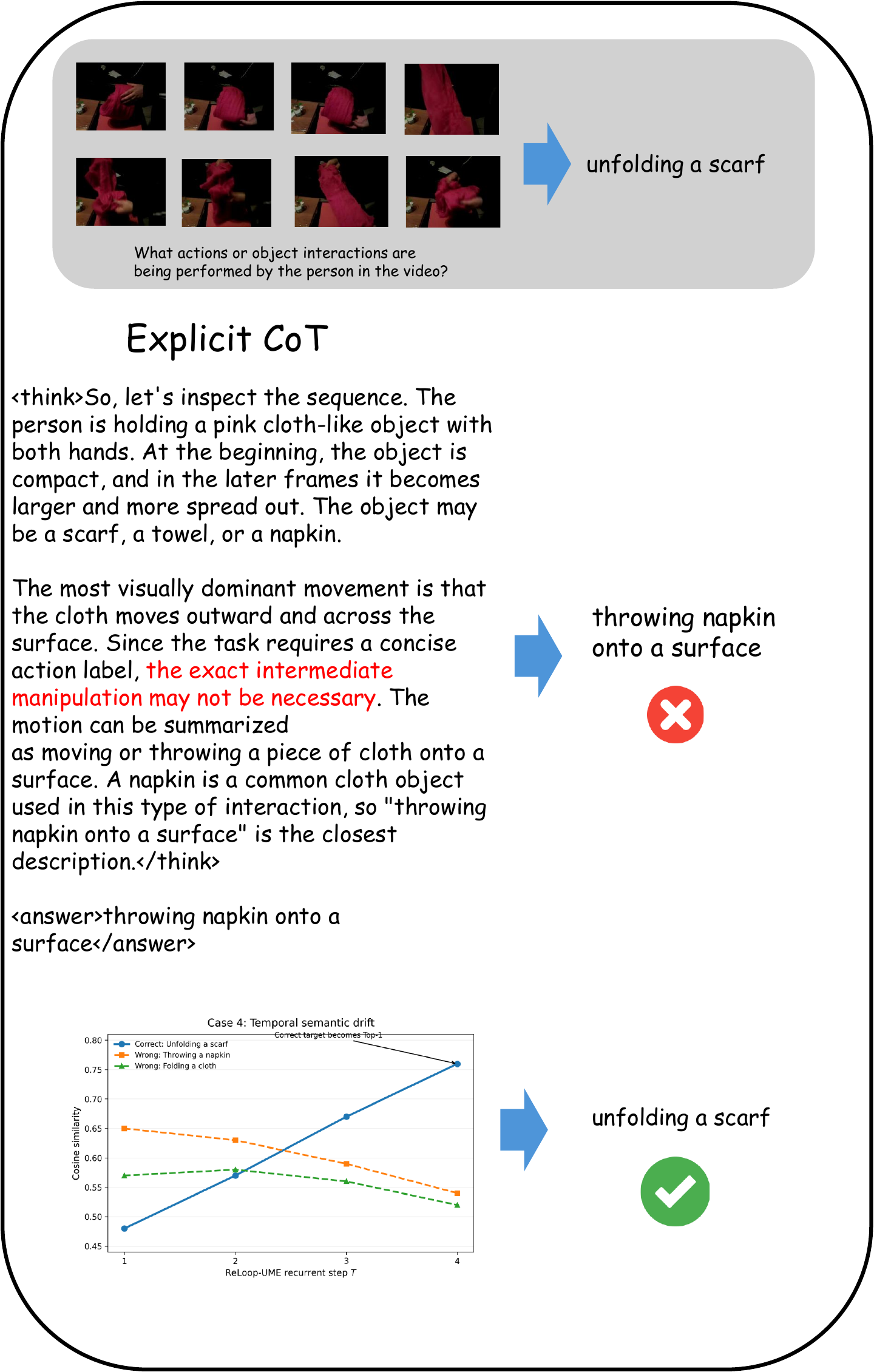} \\
\scriptsize (c) Pancake versus scrambled egg &
\scriptsize (d) Unfolding a scarf versus throwing a napkin
\end{tabular}
\captionof{figure}{Explicit-CoT failures in which a plausible linguistic prior overrides the discriminative visual evidence.  The correct targets are shown in green in the supplied examples.}
\label{fig:explicit-cot-34}
\end{center}
\vspace{-3pt}

\paragraph{Case 3: commonsense priors cover the visual evidence.}
The video frames correspond to preparing a pancake, but the generated rationale
associates repeated utensil motion with scrambling eggs and then uses the
frequency of that everyday action as additional support.  This is not a failure
of linguistic coherence; it is a failure of evidence weighting.  A strong prior
about what usually happens in a frying pan outweighs the weak but decisive
visual pattern in the sampled frames.

\paragraph{Case 4: action-process semantic drift.}
The target action is unfolding a scarf.  Explicit CoT first recognizes a
cloth-like object becoming larger and more spread out, but then replaces that
process with the more generic action ``throwing a napkin onto a surface.''  The
intermediate language therefore drifts from a temporally grounded state change
to a familiar object-action phrase.  Together, Cases 3--4 show why longer
reasoning is not automatically more retrieval-faithful: every generated token is
another opportunity to replace fine visual evidence with a high-probability
verbal interpretation.  ReLoop-UME adds depth without introducing this discrete
textual handoff.

\begin{center}
\footnotesize
\setlength{\tabcolsep}{4pt}
\renewcommand{\arraystretch}{1.04}
\begin{tabular}{@{}p{0.16\textwidth}p{0.25\textwidth}p{0.27\textwidth}p{0.25\textwidth}@{}}
\toprule
\textbf{Explicit-CoT case} & \textbf{Discriminative query signal} & \textbf{What the rationale retains} & \textbf{What is lost before retrieval} \\
\midrule
Shareholder page & The document relation ``message to shareholders'' & Texas Roadhouse identity and company-overview imagery & Investor-facing document function and page type \\
Snowy-person dialogue & Multiple positive and negative visual constraints & One person in a snowy outdoor scene & No lift/lodge, ground state, pose, and other candidate-separating details \\
Pancake video & Shape and evolution of the food across frames & Generic frying-pan motion and a common breakfast prior & The weak but decisive pancake appearance \\
Unfolding scarf & Object state changes from compact to spread out & Generic cloth motion and a familiar object-action phrase & The temporal process of unfolding and the scarf identity \\
\bottomrule
\end{tabular}
\captionof{table}{Evidence-preservation view of the four Explicit-CoT examples.  In every case, the rationale retains a plausible dominant concept while discarding the relation, constraint, appearance, or state change that determines the positive candidate.}
\label{tab:explicit-cot-case-map}
\end{center}

\newpage
\subsection{PLUME: Latent Compression Can Still Lose Fine-Grained Evidence}

\begin{center}
\setlength{\tabcolsep}{5pt}
\renewcommand{\arraystretch}{0.96}
\begin{tabular}{@{}cc@{}}
\includegraphics[height=.225\textheight,keepaspectratio]{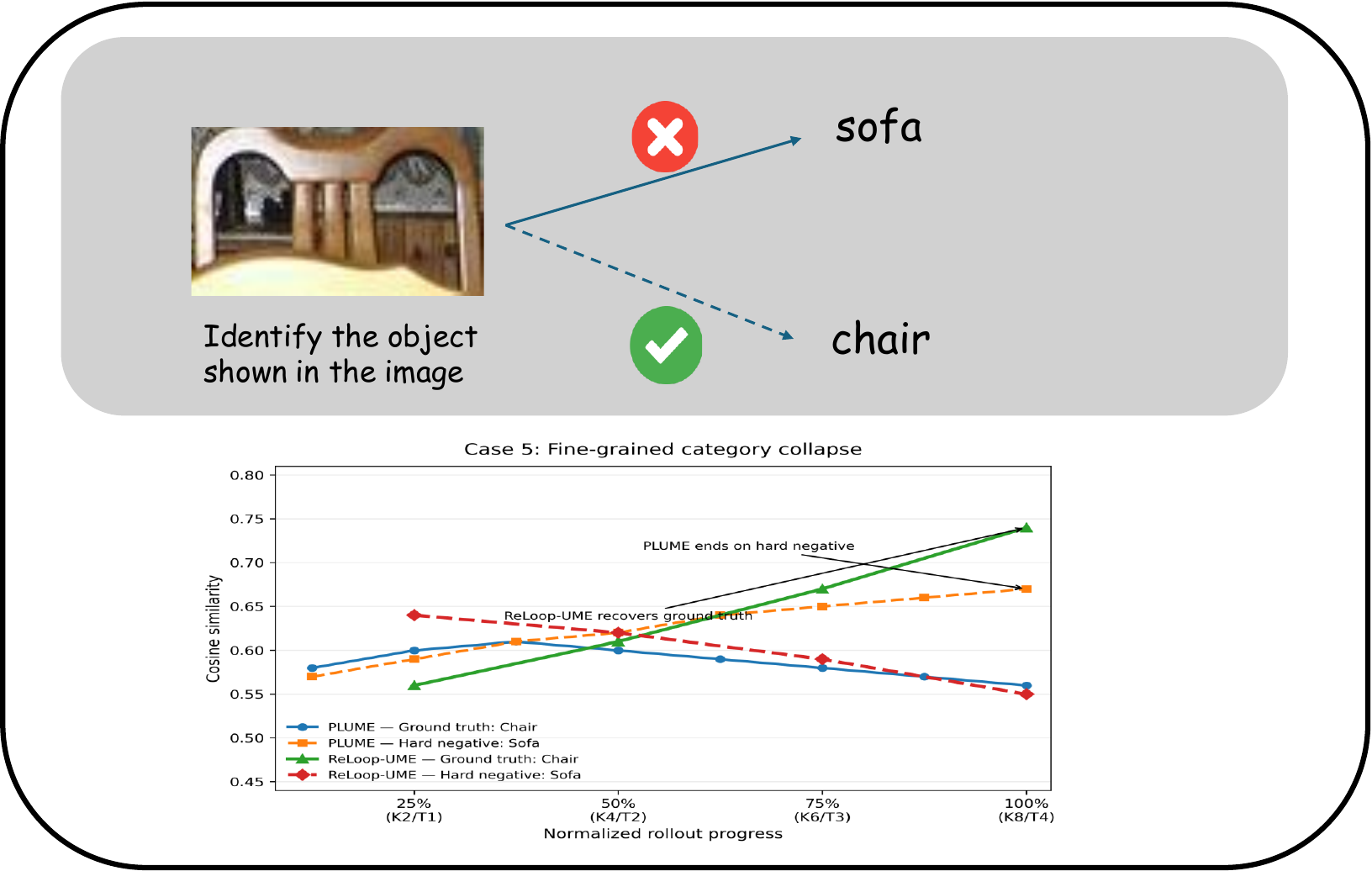} &
\includegraphics[height=.225\textheight,keepaspectratio]{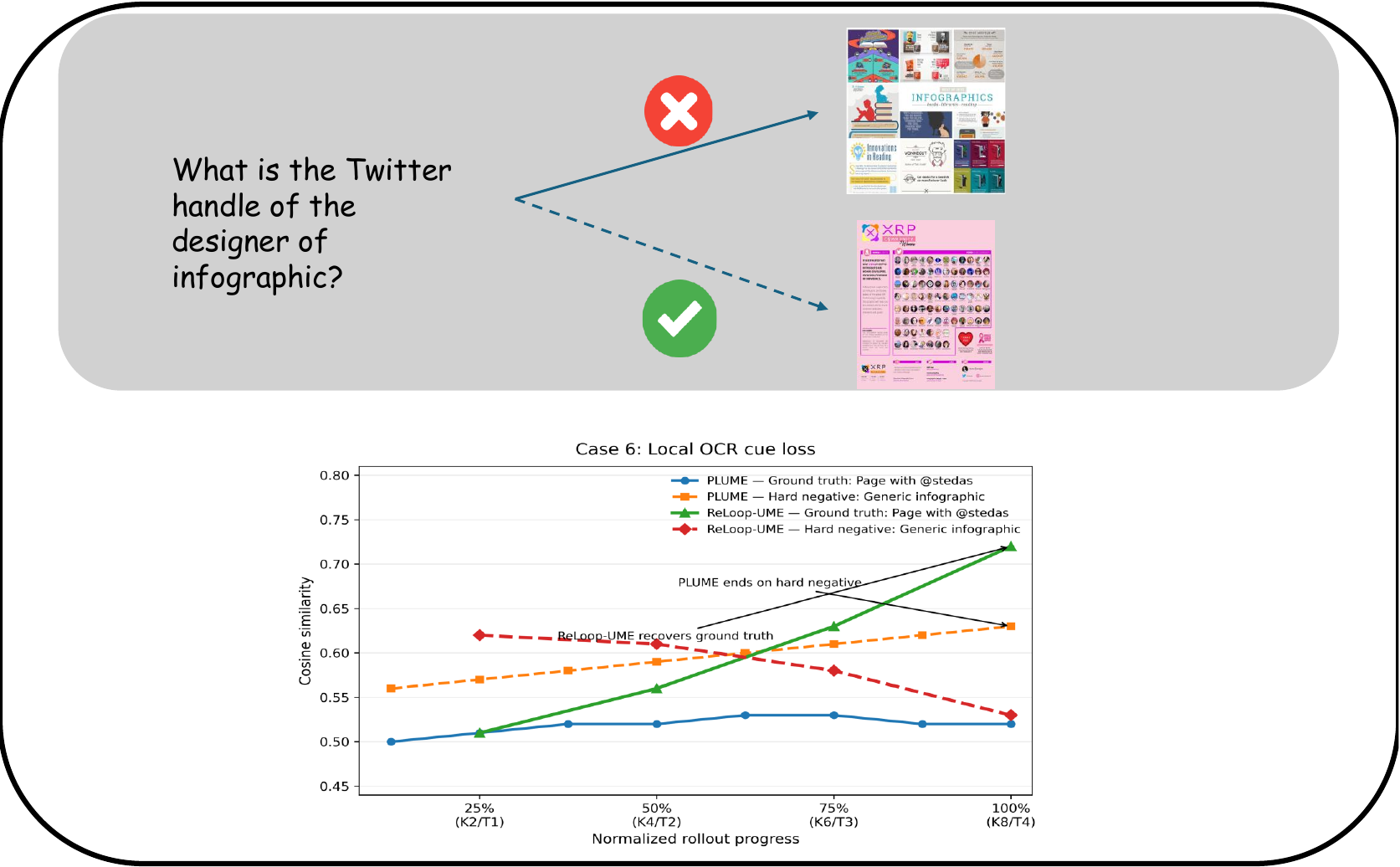} \\
[-2pt]\scriptsize (a) Chair versus sofa & \scriptsize (b) Designer Twitter handle \\
\includegraphics[height=.245\textheight,keepaspectratio]{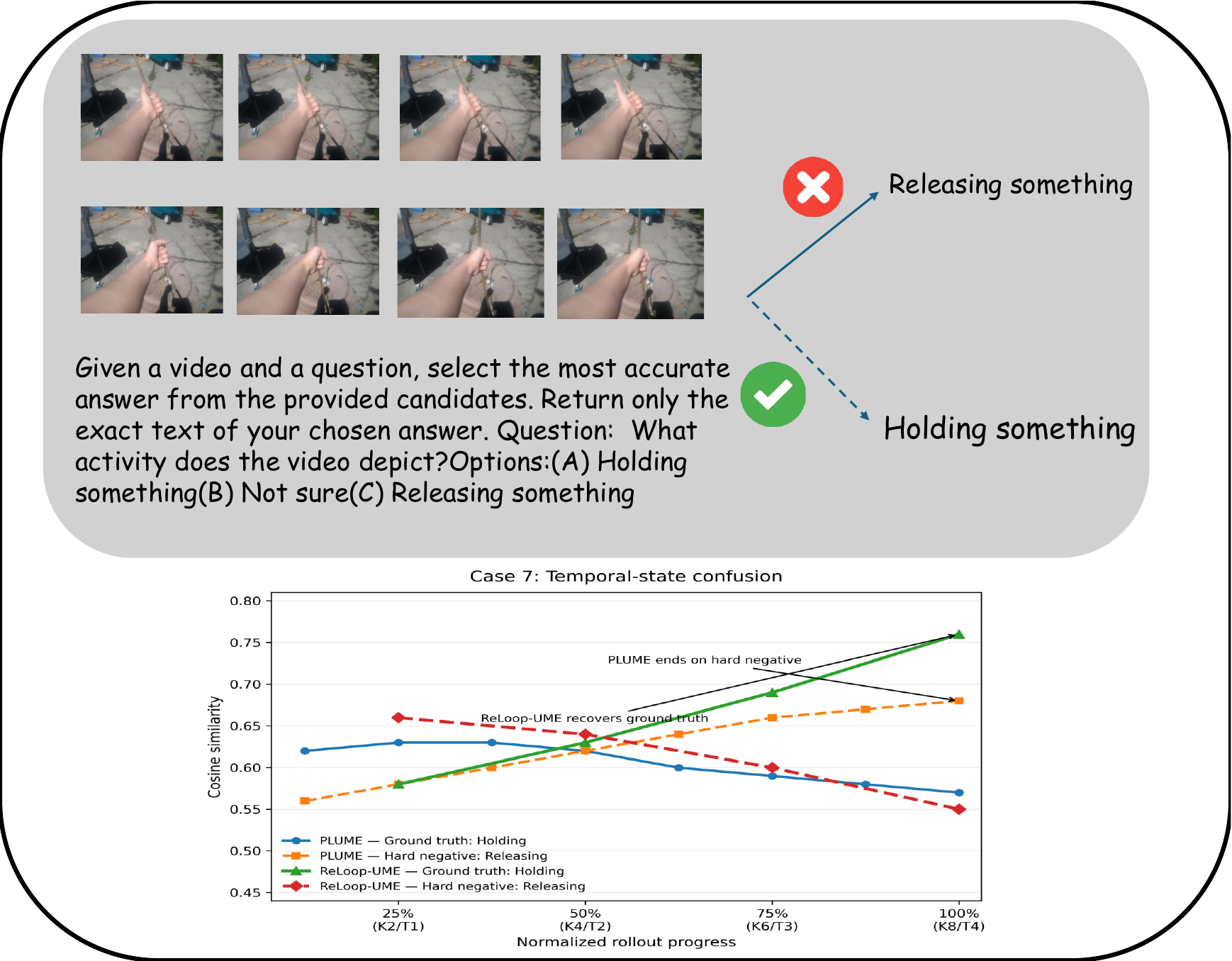} &
\includegraphics[height=.225\textheight,keepaspectratio]{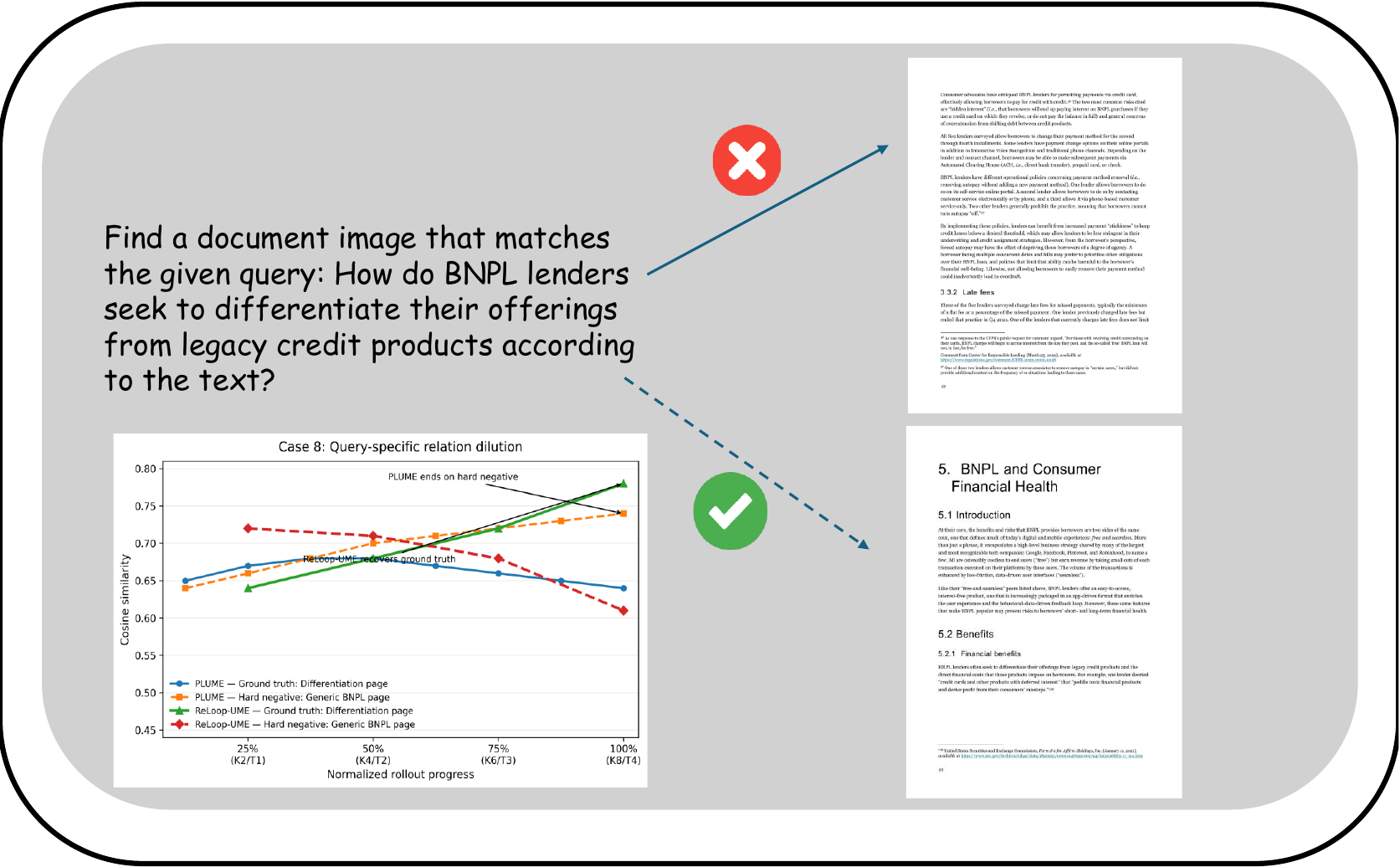} \\
[-2pt]\scriptsize (c) Holding versus releasing & \scriptsize (d) BNPL differentiation
\end{tabular}
\captionof{figure}{Four PLUME cases selected to expose fine-grained category collapse, local OCR loss, temporal-state confusion, and relation masking by a coarse document topic.  Each PDF includes the candidate comparison and the supplied layer-wise diagnostic curve.}
\label{fig:plume-failures}
\end{center}
\vspace{-4pt}

PLUME avoids a long explicit rationale, but its latent trajectory must still
compress the query and multimodal evidence before the final embedding is read
out.  The four cases expose complementary forms of information loss.  The chair
example collapses a fine-grained object boundary into the nearby category
\emph{sofa}.  The infographic query retrieves the correct broad document type
but loses the small OCR span containing the designer's Twitter handle.  In the
video example, the model preserves the manipulated object but confuses the
opposite temporal states \emph{holding} and \emph{releasing}.  Finally, the BNPL
example matches the general financing topic while missing the queried relation:
how BNPL lenders differentiate themselves from legacy credit products.

These errors share a common structure.  The retained representation preserves
the dominant topic or object while discarding the attribute, exact string,
state transition, or relation that actually determines the positive candidate.
ReLoop-UME does not create an auxiliary explicit or latent generation endpoint.
Instead, the same retrieval-forming layers are reapplied under the terminal
contrastive objective, and the registers remain available as persistent
workspace across loops.  The selected examples illustrate why this design can
preserve fine-grained evidence more reliably, although the aggregate results,
rather than the examples alone, establish the overall advantage.

\paragraph{Why these errors are complementary.}
The four PLUME cases span object identity, local text, temporal state, and
cross-sentence relation matching.  They therefore cannot be explained by one
modality-specific preprocessing error.  In each case the hard negative agrees
with most of the query and differs only in the smallest decisive component: the
furniture subclass, the exact handle, the direction of the action, or the
comparison to legacy credit.  A representation optimized around the dominant
semantics can appear well aligned while still placing the wrong item slightly
closer than the positive.

\paragraph{Connection to the recurrent-register design.}
The recurrent block is useful precisely when the relevant evidence is already
present but insufficiently consolidated.  The registers can carry a candidate-
separating attribute across repeated applications, while terminal InfoNCE keeps
the recurrent computation tied to ranking rather than to reconstructing a
verbal or latent explanation.  This does not guarantee immunity to every local
detail failure--the VisDoc example below shows that ReLoop-UME can still dilute
an exact modifier--but it explains why the method improves the average without
requiring longer generated trajectories.

\newpage
\subsection{ReLoop-UME Failures and Their Corresponding Limitations}
\label{app:reloop-failure-cases}

\begin{center}
\setlength{\tabcolsep}{3.5pt}
\begin{tabular}{@{}ccc@{}}
\includegraphics[height=.315\textheight,keepaspectratio]{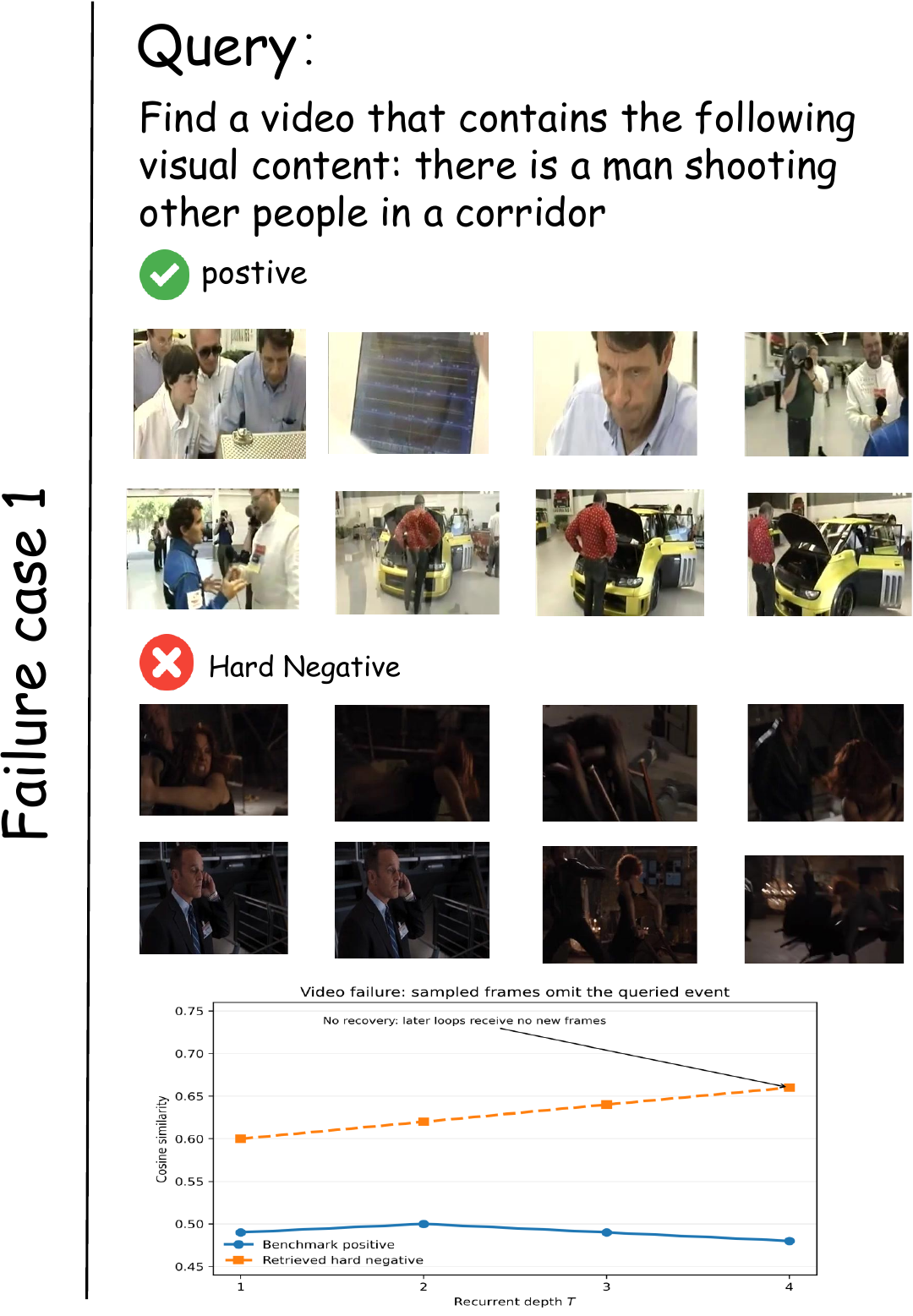} &
\includegraphics[height=.315\textheight,keepaspectratio]{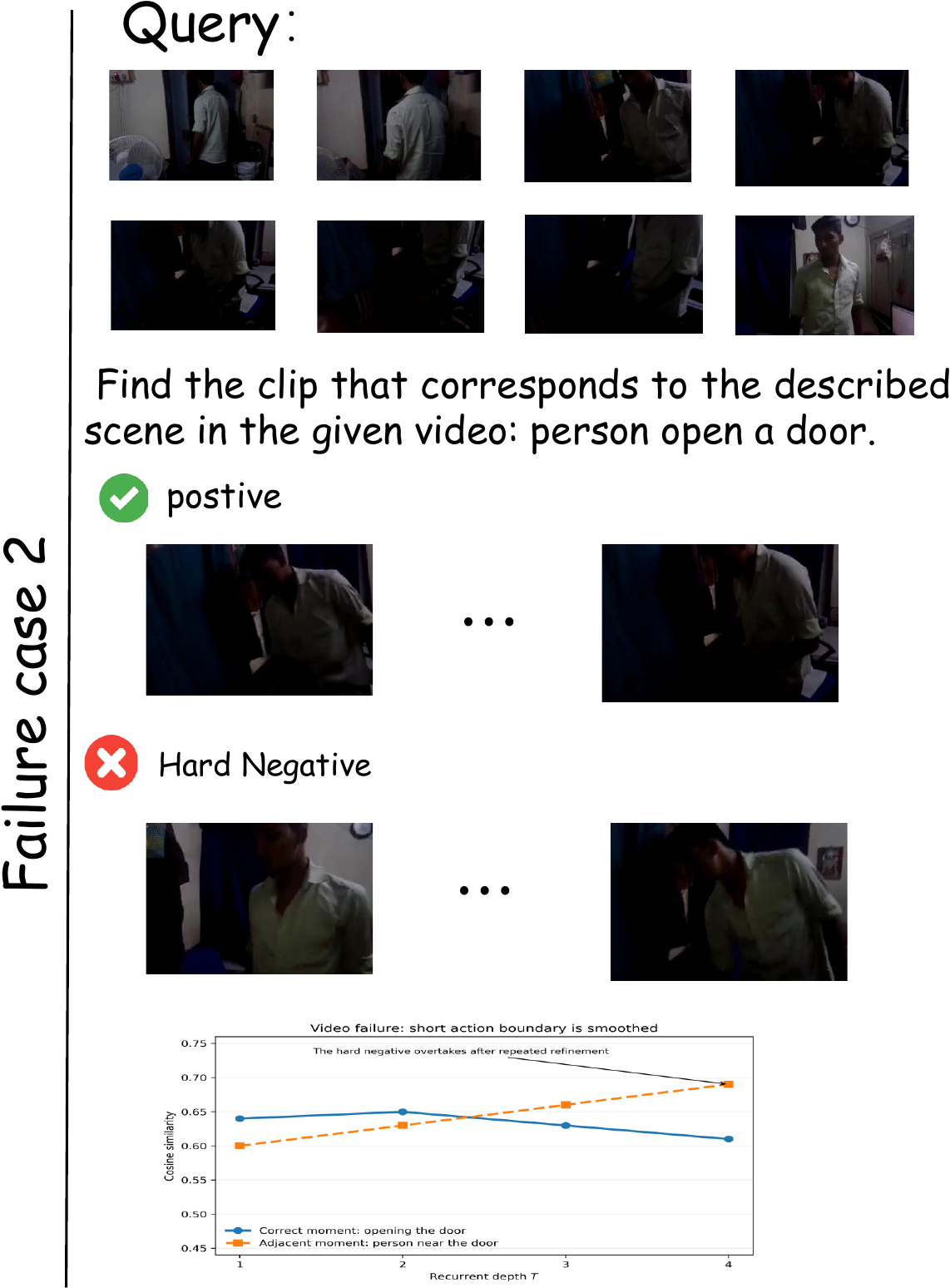} &
\includegraphics[height=.315\textheight,keepaspectratio]{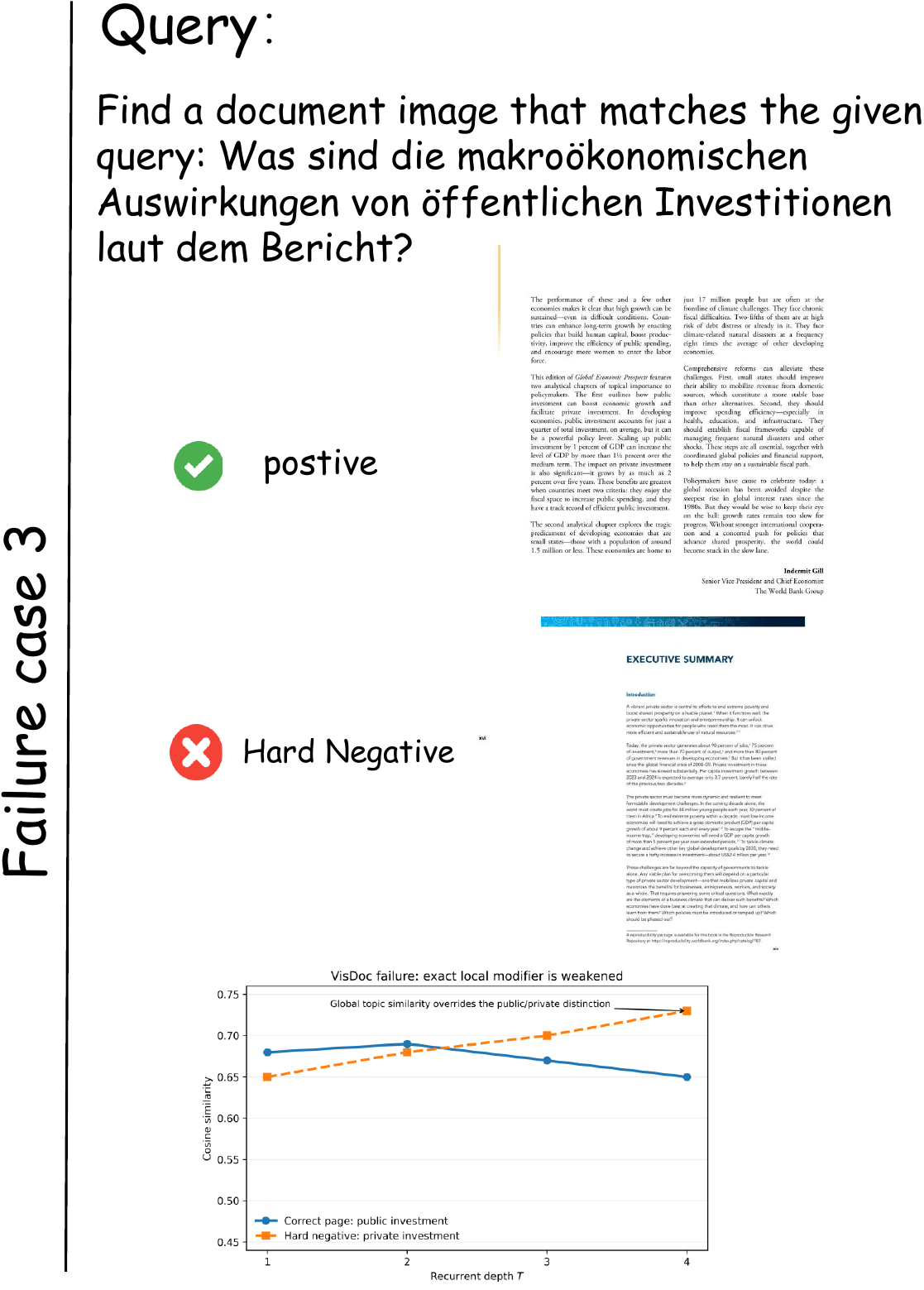} \\
\scriptsize (a) MSR-VTT shooting event &
\scriptsize (b) Charades-STA door opening &
\scriptsize (c) VisDoc public/private investment
\end{tabular}
\captionof{figure}{Three representative ReLoop-UME failures.  The green item is the positive candidate and the red item is the retrieved hard negative.  The examples instantiate temporal evidence omission, temporal-boundary smoothing, and local-detail dilution in visual documents.}
\label{fig:reloop-failures}
\end{center}
\vspace{-3pt}

\begin{center}
\footnotesize
\setlength{\tabcolsep}{4pt}
\renewcommand{\arraystretch}{1.04}
\begin{tabular}{@{}p{0.19\textwidth}p{0.19\textwidth}p{0.34\textwidth}p{0.20\textwidth}@{}}
\toprule
\textbf{Failure case} & \textbf{Limitation} & \textbf{Why recurrence does not recover it} & \textbf{Targeted remedy} \\
\midrule
MSR-VTT: shooting event is absent from the sampled frames & Temporal evidence omission & The input contains eight fixed frames sampled at 2 FPS.  Every recurrent application refines the same hidden sequence, so an action that was not sampled or was weakly encoded by the prefix cannot be reconstructed by increasing $T$. & Adaptive or denser frame sampling; per-loop re-injection of frame features. \\
Charades-STA: an adjacent static segment replaces the door-opening moment & Temporal-boundary smoothing & Positive and hard-negative clips share the person, door, and room; they differ mainly in the brief transition.  Repeated refinement strengthens stable scene semantics and can attenuate the event boundary and motion direction. & Frame-difference registers; transition-aware or moment-ranking losses. \\
VisDoc: a page about public investment is confused with one about private investment & Local-detail dilution & Both pages share the dominant topics of investment, GDP, and economic growth.  A single global embedding can preserve the topic while weakening the exact \emph{public/private} modifier and its local relation. & Page/region-specific registers; OCR residual re-injection and layout-aware contrast. \\
\bottomrule
\end{tabular}
\captionof{table}{Direct correspondence between the three supplied ReLoop-UME failure cases and the limitations discussed in the paper.}
\label{tab:failure-limitation-map}
\end{center}
\vspace{-4pt}

\paragraph{Temporal evidence omission.}
Case (a) is a sampling failure before it is a recurrent-computation failure.  If
the decisive gunshot or interaction does not appear in the eight sampled frames,
the later loops receive no new visual observation from which to infer it.  More
recurrent depth can reorganize available evidence, but it cannot create an
unobserved event.  This distinction explains why the method may obtain negative
or small gains on videos whose label depends on a short, sparsely sampled action.

\paragraph{Temporal-boundary smoothing.}
Case (b) contains the correct person, room, and door in both candidates; only the
brief opening transition separates them.  The recurrent block repeatedly
optimizes a global retrieval state and can therefore emphasize the stable scene
shared across neighboring moments.  The result is an adjacent-clip error rather
than a completely unrelated retrieval.  Motion-sensitive residuals or a
separate local temporal register bank would make the transition itself a
persistent feature rather than a small perturbation of the scene embedding.

\paragraph{Local-detail dilution in visual documents.}
Case (c) is similarly a near-neighbor error.  Both pages discuss macroeconomic
investment effects, but the query asks specifically about \emph{public}
investment while the hard negative centers on \emph{private} investment.  Global
recurrent refinement keeps the shared document topic and can suppress the exact
modifier, OCR span, or sentence-level relation.  Hierarchical page/region
registers and per-loop access to OCR and patch features are natural extensions.

\clearpage
\twocolumn
\section{Remaining Limitations and Future Directions}
\label{app:limitations}

\subsection{Scope of Stage Localization}

Stage localization should be rerun when the backbone, data mixture, readout, or
objective changes.  The interval identified here is reproducible for the tested
checkpoints, but it is not claimed to be a universal layer index for every UME.
Likewise, the pooled piecewise-linear criterion establishes a stable operational
partition; it does not prove that every neuron inside the middle segment performs
only retrieval formation or that the terminal layer performs only readout.
Controlled interventions on internal states remain necessary for a causal account.

\subsection{Adaptive Depth and Training Cost}

A query-dependent stopping rule could improve on fixed $T=4$.  Simple semantic
matching may need only one or two applications, whereas compositional, OCR-heavy,
or knowledge-intensive inputs may benefit from more.  Learning such a controller
under an explicit compute penalty would preserve the non-autoregressive character
of ReLoop-UME while reducing average latency.  Recurrent unrolling also increases
training FLOPs and activation memory even though parameters are shared; memory can
be reduced through selective checkpointing or truncated gradient paths, but those
alternatives require a separate accuracy--efficiency study.

\subsection{Additional Diagnostics}

The current analysis uses terminal retrieval accuracy and layer-wise
positive--negative separation as complementary evidence.  Future work should add
three diagnostics.  First, token-level attribution across loops can test whether
registers repeatedly attend to the same evidence or progressively integrate new
regions, frames, and OCR spans.  Second, representational similarity between
successive loops can quantify when recurrence saturates and can provide a natural
signal for adaptive stopping.  Third, counterfactual ablations that replace,
shuffle, or reset individual registers can separate persistent memory from the
benefit of simply adding extra tokens.

\subsection{Controlled Visual Re-Injection}

The most direct extension is controlled visual re-injection.  Instead of feeding
newly generated text, each recurrent application could receive a low-cost residual
from the original vision features, optionally gated by the current registers.
For video, this residual can preserve frame identity and temporal order; for
visual documents, it can preserve region and page identity.  Such a design would
retain the central advantage of recurrent depth--additional retrieval-forming
computation without autoregressive reasoning tokens--while directly addressing
the three failure mechanisms in Figure~\ref{fig:reloop-failures}.

\subsection{How the Qualitative Cases Should Be Interpreted}

The examples in Section~\ref{app:qualitative-failures} were selected because the
positive candidate, the retrieved hard negative, and the corresponding
layer-wise diagnostic were all available, allowing the error to be traced to a
specific missing distinction.  They are not used to estimate the population
frequency of each failure type and should not be read as a replacement for the
full MMEB-V2 and MRMR tables.  Their role is narrower: to show that the aggregate
patterns have concrete manifestations.  Explicit CoT can replace evidence with
a fluent narrative; PLUME can retain the topic while losing an attribute or
relation; and ReLoop-UME can still fail when the required evidence was not
sampled, lies at a brief temporal boundary, or is confined to a small OCR span.

A systematic follow-up should annotate a larger error set with these mechanism
labels and report conditional accuracy.  For video, useful strata include
whether the decisive event appears in the sampled frames, the duration of the
target action, and the temporal distance between the positive moment and the
hard negative.  For documents, strata should include OCR-span length, page
count, table or chart structure, and whether the positive differs from the hard
negative by a single modifier or number.  Such stratification would turn the
current qualitative hypotheses into directly testable predictions.

\subsection{Evaluation of the Proposed Remedies}

The proposed interventions should be evaluated with matched compute wherever
possible.  Denser or adaptive video sampling should be compared with fixed
sampling at the same number of decoded visual tokens; frame-feature re-injection
should be compared with an equally sized residual that is shuffled across time;
and page- or region-specific registers should be compared with the same number
of unstructured global registers.  In addition to retrieval accuracy, temporal
experiments should report boundary error and recall conditioned on event
coverage, while document experiments should report exact-modifier, exact-number,
and OCR-relation subsets.  These controls are necessary to distinguish genuine
evidence preservation from gains caused only by additional visual computation.

\FloatBarrier
\setlength{\bibsep}{0.5pt}

\clearpage
\begin{thebibliography}{34}
\providecommand{\natexlab}[1]{#1}

\bibitem[{Bai et~al.(2025)}]{bai2025qwen3vl}
Bai, S.; et~al. 2025.
\newblock {Qwen3-VL} Technical Report.
\newblock arXiv:2511.21631.

\bibitem[{Cui et~al.(2025)Cui, Cheng, Chen, Shukla, Awasthi, Pan, Ahuja,
  Mishra, Guo, Lim, Singh, and Fan}]{cui2025tte}
Cui, X.; Cheng, J.; Chen, H.-y.; Shukla, S.~N.; Awasthi, A.; Pan, X.; Ahuja,
  C.; Mishra, S.~K.; Guo, Q.; Lim, S.-N.; Singh, A.; and Fan, X. 2025.
\newblock Think Then Embed: Generative Context Improves Multimodal Embedding.
\newblock arXiv:2510.05014.

\bibitem[{Dehghani et~al.(2019)Dehghani, Gouws, Vinyals, Uszkoreit, and
  Kaiser}]{dehghani2019universal}
Dehghani, M.; Gouws, S.; Vinyals, O.; Uszkoreit, J.; and Kaiser, L. 2019.
\newblock Universal Transformers.
\newblock In \emph{International Conference on Learning Representations}.

\bibitem[{Fan et~al.(2025)Fan, Du, Ramchandran, and Lee}]{fan2025looped}
Fan, Y.; Du, Y.; Ramchandran, K.; and Lee, K. 2025.
\newblock Looped Transformers for Length Generalization.
\newblock In \emph{International Conference on Learning Representations}.

\bibitem[{Faysse et~al.(2025)Faysse, Sibille, Wu, Omrani, Viaud, Hudelot, and
  Colombo}]{faysse2025colpali}
Faysse, M.; Sibille, H.; Wu, T.; Omrani, B.; Viaud, G.; Hudelot, C.; and
  Colombo, P. 2025.
\newblock {ColPali}: Efficient Document Retrieval with Vision Language Models.
\newblock In \emph{International Conference on Learning Representations}.

\bibitem[{Geiping et~al.(2025)Geiping, McLeish, Jain, Kirchenbauer, Singh,
  Bartoldson, Kailkhura, Bhatele, and Goldstein}]{geiping2025recurrentdepth}
Geiping, J.; McLeish, S.; Jain, N.; Kirchenbauer, J.; Singh, S.; Bartoldson,
  B.~R.; Kailkhura, B.; Bhatele, A.; and Goldstein, T. 2025.
\newblock Scaling up Test-Time Compute with Latent Reasoning: A Recurrent Depth
  Approach.
\newblock In \emph{Advances in Neural Information Processing Systems},
  volume~38.

\bibitem[{Gu et~al.(2025)Gu, Yang, Feng, Wang, Zhang, Long, Chen, Cai, and
  Deng}]{gu2025unime}
Gu, T.; Yang, K.; Feng, Z.; Wang, X.; Zhang, Y.; Long, D.; Chen, Y.; Cai, W.;
  and Deng, J. 2025.
\newblock Breaking the Modality Barrier: Universal Embedding Learning with
  Multimodal {LLM}s.
\newblock arXiv:2504.17432.

\bibitem[{Hao et~al.(2026)Hao, Wang, Yang, Wang, Guo, and Wang}]{hao2026trace}
Hao, X.; Wang, S.; Yang, T.; Wang, T.; Guo, H.; and Wang, J. 2026.
\newblock {TRACE}: Task-Adaptive Reasoning and Representation Learning for
  Universal Multimodal Retrieval.
\newblock arXiv:2603.02929.

\bibitem[{He et~al.(2026)He, Hao, Yang, Ma, Jia, Wu, Zhao, Guo, and
  Wang}]{he2026plume}
He, C.; Hao, X.; Yang, T.; Ma, Y.; Jia, Y.; Wu, L.; Zhao, C.; Guo, H.; and
  Wang, J. 2026.
\newblock {PLUME}: Latent Reasoning Based Universal Multimodal Embedding.
\newblock arXiv:2604.02073.

\bibitem[{Jia et~al.(2021)Jia, Yang, Xia, Chen, Parekh, Pham, Le, Sung, Li, and
  Duerig}]{jia2021align}
Jia, C.; Yang, Y.; Xia, Y.; Chen, Y.-T.; Parekh, Z.; Pham, H.; Le, Q.; Sung,
  Y.-H.; Li, Z.; and Duerig, T. 2021.
\newblock Scaling Up Visual and Vision-Language Representation Learning With
  Noisy Text Supervision.
\newblock In \emph{Proceedings of the 38th International Conference on Machine
  Learning}, volume 139 of \emph{Proceedings of Machine Learning Research},
  4904--4916. PMLR.

\bibitem[{Jiang et~al.(2026)Jiang, Wang, Zhu, Lu, Qin, Wang, Wan, and
  Tang}]{jiang2026embedrl}
Jiang, H.; Wang, Y.; Zhu, Y.; Lu, X.; Qin, W.; Wang, M.; Wan, P.; and Tang, Y.
  2026.
\newblock {Embed-RL}: Reinforcement Learning for Reasoning-Driven Multimodal
  Embeddings.
\newblock arXiv:2602.13823.

\bibitem[{Jiang et~al.(2024)Jiang, Song, Zhang, Huang, Deng, Sun, Zhang, Wang,
  and Zhuang}]{jiang2024e5v}
Jiang, T.; Song, M.; Zhang, Z.; Huang, H.; Deng, W.; Sun, F.; Zhang, Q.; Wang,
  D.; and Zhuang, F. 2024.
\newblock {E5-V}: Universal Embeddings with Multimodal Large Language Models.
\newblock arXiv:2407.12580.

\bibitem[{Jiang et~al.(2025)Jiang, Meng, Yang, Yavuz, Zhou, and
  Chen}]{jiang2025vlm2vec}
Jiang, Z.; Meng, R.; Yang, X.; Yavuz, S.; Zhou, Y.; and Chen, W. 2025.
\newblock {VLM2Vec}: Training Vision-Language Models for Massive Multimodal
  Embedding Tasks.
\newblock In \emph{The Thirteenth International Conference on Learning
  Representations}.

\bibitem[{Koishekenov, Lipani, and Cancedda(2025)}]{koishekenov2025etd}
Koishekenov, Y.; Lipani, A.; and Cancedda, N. 2025.
\newblock Encode, Think, Decode: Scaling Test-Time Reasoning with Recursive
  Latent Thoughts.
\newblock \emph{arXiv preprint arXiv:2510.07358}.

\bibitem[{Lan et~al.(2026)Lan, Niu, Meng, Zhou, and Su}]{lan2026umer1}
Lan, Z.; Niu, L.; Meng, F.; Zhou, J.; and Su, J. 2026.
\newblock {UME-R1}: Exploring Reasoning-Driven Generative Multimodal
  Embeddings.
\newblock In \emph{The Fourteenth International Conference on Learning
  Representations}.

\bibitem[{Li et~al.(2026)Li, Zhang, Long, Chen, Song, Bai, Yang, Xie, Yang,
  Liu, Zhou, and Lin}]{li2026qwen3vlembedding}
Li, M.; Zhang, Y.; Long, D.; Chen, K.; Song, S.; Bai, S.; Yang, Z.; Xie, P.;
  Yang, A.; Liu, D.; Zhou, J.; and Lin, J. 2026.
\newblock {Qwen3-VL-Embedding and Qwen3-VL-Reranker}: A Unified Framework for
  State-of-the-Art Multimodal Retrieval and Ranking.
\newblock arXiv:2601.04720.

\bibitem[{Lin et~al.(2024)Lin, Lee, Shoeybi, Lin, Catanzaro, and
  Ping}]{lin2024mmembed}
Lin, S.-C.; Lee, C.; Shoeybi, M.; Lin, J.; Catanzaro, B.; and Ping, W. 2024.
\newblock {MM-Embed}: Universal Multimodal Retrieval with Multimodal {LLM}s.
\newblock arXiv:2411.02571.

\bibitem[{Liu et~al.(2025{\natexlab{a}})Liu, Yang, Gao, Zhu, Zhu, Zhao, and
  Wang}]{liu2025rge}
Liu, C.; Yang, J.; Gao, R.; Zhu, Y.; Zhu, F.; Zhao, R.; and Wang, L.
  2025{\natexlab{a}}.
\newblock Reasoning Guided Embeddings: Leveraging {MLLM} Reasoning for Improved
  Multimodal Retrieval.
\newblock arXiv:2511.16150.

\bibitem[{Liu et~al.(2025{\natexlab{b}})Liu, Zhang, Cai, Jiang, Hu, Yao, Wang,
  and Xie}]{liu2025lamra}
Liu, Y.; Zhang, Y.; Cai, J.; Jiang, X.; Hu, Y.; Yao, J.; Wang, Y.; and Xie, W.
  2025{\natexlab{b}}.
\newblock {LamRA}: Large Multimodal Model as Your Advanced Retrieval Assistant.
\newblock In \emph{Proceedings of the IEEE/CVF Conference on Computer Vision
  and Pattern Recognition}, 4015--4025.

\bibitem[{Meng et~al.(2026)Meng, Jiang, Liu, Su, Yang, Fu, Qin, Thirukovalluru,
  Zhang, Chen, Xu, Xiong, Zhou, Chen, and Yavuz}]{meng2026vlm2vecv2}
Meng, R.; Jiang, Z.; Liu, Y.; Su, M.; Yang, X.; Fu, Y.; Qin, C.;
  Thirukovalluru, R.; Zhang, X.; Chen, Z.; Xu, R.; Xiong, C.; Zhou, Y.; Chen,
  W.; and Yavuz, S. 2026.
\newblock {{VLM}2Vec-V2}: Advancing Multimodal Embedding for Videos, Images,
  and Visual Documents.
\newblock \emph{Transactions on Machine Learning Research}.

\bibitem[{Park et~al.(2026)Park, Lee, Kim, and Bae}]{park2026loopus}
Park, T.; Lee, Y.; Kim, D.; and Bae, H. 2026.
\newblock {LoopUS}: Recasting Pretrained {LLM}s into Looped Latent Refinement
  Models.
\newblock arXiv:2605.11011.

\bibitem[{{Qwen Team}(2026)}]{qwen2026qwen35}
{Qwen Team}. 2026.
\newblock {Qwen3.5}: Towards Native Multimodal Agents.

\bibitem[{Radford et~al.(2021)Radford, Kim, Hallacy, Ramesh, Goh, Agarwal,
  Sastry, Askell, Mishkin, Clark, Krueger, and Sutskever}]{radford2021clip}
Radford, A.; Kim, J.~W.; Hallacy, C.; Ramesh, A.; Goh, G.; Agarwal, S.; Sastry,
  G.; Askell, A.; Mishkin, P.; Clark, J.; Krueger, G.; and Sutskever, I. 2021.
\newblock Learning Transferable Visual Models From Natural Language
  Supervision.
\newblock In \emph{Proceedings of the 38th International Conference on Machine
  Learning}, volume 139 of \emph{Proceedings of Machine Learning Research},
  8748--8763. PMLR.

\bibitem[{Saunshi et~al.(2025)Saunshi, Dikkala, Li, Kumar, and
  Reddi}]{saunshi2025latentthoughts}
Saunshi, N.; Dikkala, N.; Li, Z.; Kumar, S.; and Reddi, S.~J. 2025.
\newblock Reasoning with Latent Thoughts: On the Power of Looped Transformers.
\newblock In \emph{International Conference on Learning Representations}.

\bibitem[{Sun et~al.(2026)Sun, Ren, Liao, Mao, Ren, Zhang, Zhao, Lin, Jiang,
  Zhang, and Zheng}]{sun2026btoks}
Sun, S.; Ren, J.; Liao, Z.; Mao, D.; Ren, X.; Zhang, Y.; Zhao, H.; Lin, W.;
  Jiang, S.; Zhang, L.; and Zheng, Y. 2026.
\newblock Bottleneck Tokens for Unified Multimodal Retrieval.
\newblock arXiv:2604.11095.

\bibitem[{Wang et~al.(2024)Wang, Bai, Tan, Wang, Fan, Bai, Chen, Liu, Wang, Ge,
  Fan, Dang, Du, Ren, Men, Liu, Zhou, Zhou, and Lin}]{wang2024qwen2vl}
Wang, P.; Bai, S.; Tan, S.; Wang, S.; Fan, Z.; Bai, J.; Chen, K.; Liu, X.;
  Wang, J.; Ge, W.; Fan, Y.; Dang, K.; Du, M.; Ren, X.; Men, R.; Liu, D.; Zhou,
  C.; Zhou, J.; and Lin, J. 2024.
\newblock {Qwen2-VL}: Enhancing Vision-Language Model's Perception of the World
  at Any Resolution.
\newblock arXiv:2409.12191.

\bibitem[{Wei et~al.(2024)Wei, Chen, Chen, Hu, Zhang, Fu, Ritter, and
  Chen}]{wei2024uniir}
Wei, C.; Chen, Y.; Chen, H.; Hu, H.; Zhang, G.; Fu, J.; Ritter, A.; and Chen,
  W. 2024.
\newblock {UniIR}: Training and Benchmarking Universal Multimodal Information
  Retrievers.
\newblock In \emph{European Conference on Computer Vision}, 387--404. Springer.

\bibitem[{Xiao et~al.(2026)Xiao, Ma, Gu, cheng Jason~Chen, Chen, Ordonez, and
  Mohan}]{lee2025metaembed}
Xiao, Z.; Ma, Q.; Gu, M.; cheng Jason~Chen, C.; Chen, X.; Ordonez, V.; and
  Mohan, V. 2026.
\newblock MetaEmbed: Scaling Multimodal Retrieval at Test-Time with Flexible
  Late Interaction.
\newblock arXiv:2509.18095.

\bibitem[{Zhai et~al.(2023)Zhai, Mustafa, Kolesnikov, and
  Beyer}]{zhai2023siglip}
Zhai, X.; Mustafa, B.; Kolesnikov, A.; and Beyer, L. 2023.
\newblock Sigmoid Loss for Language Image Pre-Training.
\newblock In \emph{Proceedings of the IEEE/CVF International Conference on
  Computer Vision}, 11975--11986.

\bibitem[{Zhang et~al.(2024)Zhang, Luan, Hu, Lee, Qiao, Chen, Su, and
  Chang}]{zhang2024magiclens}
Zhang, K.; Luan, Y.; Hu, H.; Lee, K.; Qiao, S.; Chen, W.; Su, Y.; and Chang,
  M.-W. 2024.
\newblock {MagicLens}: Self-Supervised Image Retrieval with Open-Ended
  Instructions.
\newblock arXiv:2403.19651.

\bibitem[{Zhang et~al.(2026)Zhang, Gao, Zhou, Zhao, Song, Cohan, Luu, and
  Zhao}]{zhang2026mrmr}
Zhang, S.; Gao, Y.; Zhou, X.; Zhao, Y.; Song, T.; Cohan, A.; Luu, A.~T.; and
  Zhao, C. 2026.
\newblock {MRMR}: A Realistic and Expert-Level Multidisciplinary Benchmark for
  Reasoning-Intensive Multimodal Retrieval.
\newblock In \emph{The Fourteenth International Conference on Learning
  Representations}.

\bibitem[{Zhang et~al.(2025)Zhang, Zhang, Xie, Li, Dai, Long, Xie, Zhang, Li,
  and Zhang}]{zhang2025gme}
Zhang, X.; Zhang, Y.; Xie, W.; Li, M.; Dai, Z.; Long, D.; Xie, P.; Zhang, M.;
  Li, W.; and Zhang, M. 2025.
\newblock Bridging Modalities: Improving Universal Multimodal Retrieval by
  Multimodal Large Language Models.
\newblock In \emph{Proceedings of the IEEE/CVF Conference on Computer Vision
  and Pattern Recognition}, 9274--9285.

\bibitem[{Zhou et~al.(2025)Zhou, Xiong, Liu, Liu, Xiao, Wang, Zhao, Zhang, and
  Lian}]{zhou2025megapairs}
Zhou, J.; Xiong, Y.; Liu, Z.; Liu, Z.; Xiao, S.; Wang, Y.; Zhao, B.; Zhang,
  C.~J.; and Lian, D. 2025.
\newblock {MegaPairs}: Massive Data Synthesis for Universal Multimodal
  Retrieval.
\newblock In \emph{Proceedings of the 63rd Annual Meeting of the Association
  for Computational Linguistics (Volume 1: Long Papers)}, 19076--19095.

\bibitem[{Zhu et~al.(2025)Zhu, Wang, Hua, Zhang, Li, Que, Wei, Wen, Yin, Xing,
  Li, Shi, Ma, Li, Kergan, Smith, Qu, Hui, Wu, Min, Huang, Zhou, Ye, Liu, Yang,
  Shi, Lin, Zhao, Cai, Zhang, Huang, Bengio, and Eshraghian}]{zhu2025ouro}
Zhu, R.-J.; Wang, Z.; Hua, K.; Zhang, T.; Li, Z.; Que, H.; Wei, B.; Wen, Z.;
  Yin, F.; Xing, H.; Li, L.; Shi, J.; Ma, K.; Li, S.; Kergan, T.; Smith, A.;
  Qu, X.; Hui, M.; Wu, B.; Min, Q.; Huang, H.; Zhou, X.; Ye, W.; Liu, J.; Yang,
  J.; Shi, Y.; Lin, C.; Zhao, E.; Cai, T.; Zhang, G.; Huang, W.; Bengio, Y.;
  and Eshraghian, J. 2025.
\newblock Scaling Latent Reasoning via Looped Language Models.
\newblock arXiv:2510.25741.

\end{thebibliography}

{\scriptsize\begin{thebibliography}{8}
\providecommand{\natexlab}[1]{#1}

\bibitem[{Jiang et~al.(2025)Jiang, Meng, Yang, Yavuz, Zhou, and
  Chen}]{supp_jiang2025vlm2vec}
Jiang, Z.; Meng, R.; Yang, X.; Yavuz, S.; Zhou, Y.; and Chen, W. 2025.
\newblock {VLM2Vec}: Training Vision-Language Models for Massive Multimodal
  Embedding Tasks.
\newblock In \emph{The Thirteenth International Conference on Learning
  Representations}.

\bibitem[{Meng et~al.(2026)Meng, Jiang, Liu, Su, Yang, Fu, Qin, Thirukovalluru,
  Zhang, Chen, Xu, Xiong, Zhou, Chen, and Yavuz}]{supp_meng2026vlm2vecv2}
Meng, R.; Jiang, Z.; Liu, Y.; Su, M.; Yang, X.; Fu, Y.; Qin, C.;
  Thirukovalluru, R.; Zhang, X.; Chen, Z.; Xu, R.; Xiong, C.; Zhou, Y.; Chen,
  W.; and Yavuz, S. 2026.
\newblock {{VLM}2Vec-V2}: Advancing Multimodal Embedding for Videos, Images,
  and Visual Documents.
\newblock \emph{Transactions on Machine Learning Research}.

\bibitem[{{Qwen Team}(2024)}]{supp_qwen2vl2bconfig}
{Qwen Team}. 2024.
\newblock {Qwen2-VL-2B-Instruct}: Official Model Configuration.
\newblock Hugging Face.

\bibitem[{{Qwen Team}(2025{\natexlab{a}})}]{supp_qwen2025embeddingblog}
{Qwen Team}. 2025{\natexlab{a}}.
\newblock {Qwen3 Embedding}: Advancing Text Embedding and Reranking Through
  Foundation Models.
\newblock Qwen technical blog.

\bibitem[{{Qwen Team}(2025{\natexlab{b}})}]{supp_qwen3vl2bconfig}
{Qwen Team}. 2025{\natexlab{b}}.
\newblock {Qwen3-VL-2B-Instruct}: Official Model Configuration.
\newblock Hugging Face.

\bibitem[{{Qwen Team}(2026)}]{supp_qwen35_2bconfig}
{Qwen Team}. 2026.
\newblock {Qwen3.5-2B}: Official Model Configuration.
\newblock Hugging Face.

\bibitem[{Zhang et~al.(2026)Zhang, Gao, Zhou, Zhao, Song, Cohan, Luu, and
  Zhao}]{supp_zhang2026mrmr}
Zhang, S.; Gao, Y.; Zhou, X.; Zhao, Y.; Song, T.; Cohan, A.; Luu, A.~T.; and
  Zhao, C. 2026.
\newblock {MRMR}: A Realistic and Expert-Level Multidisciplinary Benchmark for
  Reasoning-Intensive Multimodal Retrieval.
\newblock In \emph{The Fourteenth International Conference on Learning
  Representations}.

\bibitem[{Zhang et~al.(2025)Zhang, Li, Long, Zhang, Lin, Yang, Xie, Yang, Liu,
  Lin, Huang, and Zhou}]{supp_zhang2025qwen3embedding}
Zhang, Y.; Li, M.; Long, D.; Zhang, X.; Lin, H.; Yang, B.; Xie, P.; Yang, A.;
  Liu, D.; Lin, J.; Huang, F.; and Zhou, J. 2025.
\newblock {Qwen3 Embedding}: Advancing Text Embedding and Reranking Through
  Foundation Models.
\newblock arXiv:2506.05176.

\end{thebibliography}
}
\end{document}